\documentclass{article}
\usepackage{iclr2027_conference,times}
\usepackage[medium]{sourcecodepro}

\usepackage{amsmath,amsfonts,bm}

\def\eqref#1{equation~\ref{#1}}

\def\1{\bm{1}}

\DeclareMathAlphabet{\mathsfit}{\encodingdefault}{\sfdefault}{m}{sl}
\SetMathAlphabet{\mathsfit}{bold}{\encodingdefault}{\sfdefault}{bx}{n}

\newcommand{\E}{\mathbb{E}}

\renewcommand{\eqref}[1]{(\ref{#1})}
\usepackage{amssymb}
\usepackage[scr=boondox]{mathalpha}

\usepackage{hyperref}
\usepackage{url}
\usepackage{graphicx}
\usepackage{booktabs}
\usepackage{tikz-cd}
\definecolor{unreviewedblue}{RGB}{0,51,102}
\newtheorem{theorem}{Theoretical Result}
\DeclareMathOperator{\tr}{tr}
\title{Sliding Networks: A Family of Residual
Structures Preserving Feature Learning and Feature Diversity}

\author{Datong Zhou$^\star$ \& Blake Bordelon$^\dagger$  
\\
$^{\star}$Laboratoire Jacques-Louis Lions (LJLL) \& Laboratoire de Probabilités,
\\
Statistique et Modélisation (LPSM)
\\
Sorbonne Université
\\
$^{\dagger}$Oden Institute and Department of Neuroscience
\\
University of Texas at Austin
\\
\texttt{datong.zhou@sorbonne-universite.fr}
\\
\texttt{blake\_bordelon@utexas.edu} 
}

\iclrfinalcopy

\begin{document}
\maketitle
\begin{abstract}
Deep learning models tend to perform best at large widths and depths.
The growing interest in scaling laws, especially in the language modeling
domain, has motivated procedures for scaling up network widths and depths
in order to preserve stability and computational efficiency of training dynamics.
Existing depth-scaling approaches force a choice between retaining locally maximal
\emph{feature learning} and preserving a form of
\emph{feature diversity} in the residual stream arising from
random differences between channels.
We examine a \textit{joint scaling recipe} that preserves both.
With residual width $N_\flat$, hidden width $N_\sharp$, and depth $L$,
the key quantity is, by default, the limiting ratio
$N_\flat/(N_\sharp L)\to\alpha$, while correlated initialization (Half Copy)
can partially relax the resulting shape constraints.
Theory and residual-MLP experiments support convergence at fixed positive
$\alpha$ to an SDE limit with locally maximal feature learning.
Hidden computations can then be redistributed between width and depth
while preserving this limit, like turning melody into harmony and back.
Language-model experiments nevertheless show performance differences
under regrouping at fixed parameter count.
These results distinguish approximating a limit (ODE or SDE) from scaling for better
performance.
\end{abstract}
\section{Introduction}
\subsection{Motivation and prior work}
\label{sec:scaling-motivation}
Deep learning has driven major advances in artificial intelligence.
This approach takes its name from the composition of
multiple layers of trainable transformations \citep[see][for a review]{lecun2015deep}.
Architectural landmarks such as AlexNet \citep{krizhevsky2012imagenet},
ResNet \citep{he2015deep}, and the Transformer \citep{vaswani2017attention}
helped establish the foundations of modern models, including large
language models \citep[see][for a survey]{zhao2026survey}.

As models grow larger, choices of depth, width, and optimization procedure become
increasingly important because they largely govern the performance obtained
from a given computational budget \citep{tay2022scale,petty2024impact,brown2022wide}.
Quantitative studies of this relationship have yielded empirical scaling laws
linking performance to model size, data volume, and training compute
\citep{kaplan2020scaling}.
These laws have been used to guide the allocation of training budgets for
larger models \citep{hoffmann2022training}.
However, parameter count, whether measured at training or inference time,
is only one of many factors shaping training dynamics and final performance
\citep{sutskever2013importance}.
Exhaustively optimizing the other architectural and training choices at every
model size is impractical \citep{bergstra2012random}, so comparisons by parameter
count remain conditional on the configurations explored.

\begin{figure}[!t]
\centering
\includegraphics[scale=.92]{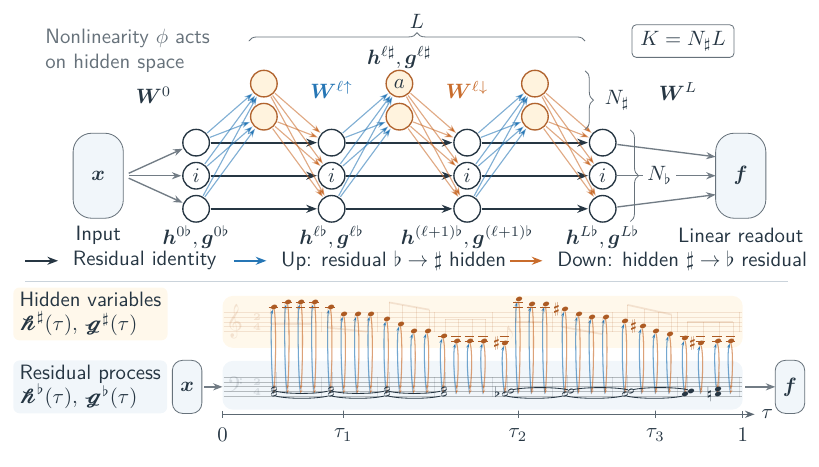}
\caption{Upper: a standard residual MLP structure. Lower: the continuous-depth
picture. Residual coordinates evolve through depth, while hidden units
contribute at specific depths. The symbols $\flat$ and $\sharp$ distinguish
the two parts. $h$ denotes preactivations and $g$ rescaled gradients (Jacobian).}
\label{fig:architecture-limit}
\end{figure}

\begin{figure}[!t]
\centering
\includegraphics{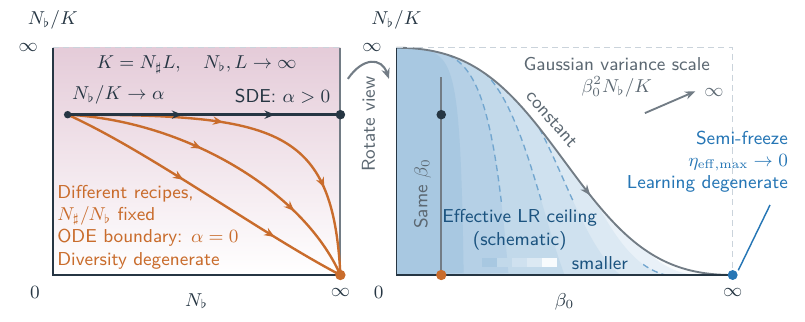}
\caption{Left: growth paths and their limits.
Right: the coupling--learning tradeoff within the limiting family.
Colors indicate the variance of the random depth contribution on the left
and the effective learning-rate ceiling on the right.
The color trends and contours are schematic.}
\label{fig:scaling-phase}
\end{figure}

Theory provides tools for understanding and classifying how parameterization
and training hyperparameters affect learning as networks grow.
For example, as network size grows under NTK scaling, changes in the activations of
individual hidden neurons decay, while
the network's output prediction dynamics approach those described by a fixed kernel
\citep{jacot2018neural,lee2019wide}.
Mean-field / $\mu$P scalings instead retain constant scale updates to these hidden features
in two-layer \citep{mei2018meanfield,chizat2018global,sirignano2020meanfield}
and deeper networks \mbox{\citep{yang2021feature,bordelon2022selfconsistent}}.
Parameterizations that preserve feature learning can help hyperparameter
settings tuned on smaller models remain effective as model size increases,
providing a systematic route to training and understanding larger networks
\citep{yang2021tuning,dey2025completep}.

However, for residual architectures such as Transformers, a large depth scaling limit
that preserves feature learning does not by itself guarantee better performance
and may suppress a form of feature diversity present at finite size. 
We use the MLP residual structure in a Transformer as a simple example
(illustrated in the upper panel of Fig.~\ref{fig:architecture-limit}),
with attention temporarily set aside \footnote{Attention has the same projection
structure when each head is treated as one finite-dimensional nonlinear
channel, corresponding to an orange circle in Fig.~\ref{fig:architecture-limit}; see
Appendix~\ref{app:attention-channel-structure}.}.
Here, the residual stream has width $N_\flat$, each block has hidden width
$N_\sharp$, and the network contains $L$ such blocks.
In practice, one common choice is $N_\sharp = 4N_\flat$
\citep{vaswani2017attention}.
The contributions of hidden neurons accumulate in the residual stream as
hidden width and depth increase.
Controlling this accumulation is already necessary
for stable, nondegenerate signal propagation at initialization
\citep{schoenholz2017deep,hayou2021stable,noci2023shaped}.
Scaling down each block's coupling to the residual stream, together with
corresponding learning-rate adjustments, can stabilize training while
retaining feature learning, and yield a continuous-depth description for
sufficiently large $L$ \citep{bordelon2024infinite,chizat2025hidden}.
Yet the accompanying averaging may diminish the diversity of learned
residual features.

The ratio of residual width to the total hidden population helps explain
this tension between preserving feature learning and feature diversity:
$N_\flat/K$, where $K=N_\sharp L$.
The common choice $N_\sharp\asymp N_\flat\asymp N$ forces
$N_\flat/K\to0$ as $L\to\infty$, regardless of how $N$ grows relative to $L$
(Fig.~\ref{fig:scaling-phase}, left).
In this setting, \citet{yang2023tensorvi} already pointed out the dilemma that
preserving feature learning requires stronger averaging over $K$, which washes out
random depth variation in residual trajectories and yields an ODE-like
limit that they regard as redundant because it loses feature diversity.
Allowing the three sizes to vary independently, \citet{chizat2025hidden}
and \citet{chaintron2026resnets} rigorously establish this ODE regime when
the ratio vanishes, while \citet{jiang2026hyperparameter} develop related
scaling recipes for mixture-of-experts networks, where total hidden
capacity differs from the capacity activated per token.
These works also point to non-ODE dynamics at nonzero
corresponding ratios \citep{chaintron2026resnets,jiang2026hyperparameter}.

\subsection{Our perspective and main findings}

\begin{figure}[!t]
\centering
\includegraphics{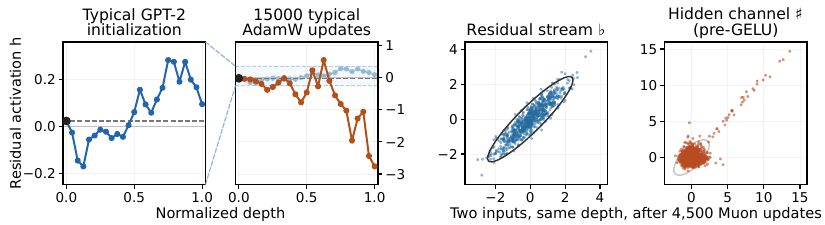}
\caption{Selected GPT-2 activations (Appendix~\ref{app:gpt2-illustration}).
Left: one residual coordinate through depth.
Right: paired standardized activations for two inputs;
ellipses show covariance-matched Gaussian 95\% contours.}
\label{fig:gpt2-theory-illustration}
\end{figure}

Our theory examines network scaling through an SDE description (Theoretical Result~\ref{thm:joint-limit}) that
retains random depth variation in residual trajectories (Fig.~\ref{fig:gpt2-theory-illustration}, left),
with the ODE description
recovered when this variation vanishes.
In this description, residual coordinates follow random paths through depth,
while individual hidden features learn at their respective depths
(Fig.~\ref{fig:architecture-limit}, lower).
For SGD and Muon \citep{jordan2024muon}, with the latter understood as a spectral transform,
the updates do not privilege a residual coordinate frame,
allowing the residual statistics to remain jointly Gaussian,
even though nonlinearities can make hidden
features strongly non-Gaussian (Fig.~\ref{fig:gpt2-theory-illustration}, right).

One natural question is whether the training recipe can compensate for
the effect of $N_\flat/K$ in the limiting dynamics, allowing more freedom on network shapes, such as $N_\flat$ and $N_\sharp$ 
proportional considered in \citet{yang2023tensorvi}.
Increasing the residual coupling strength $\beta_0$ can keep the random
variance scale $\beta_0^2N_\flat/K$ finite and nonzero at initialization
as $N_\flat/K\to0$
(Theoretical Result~\ref{thm:linear-network}, first part).
However, even one GD step in a linear network shows that this also
amplifies learning feedback
(Theoretical Result~\ref{thm:linear-network}, second part).
Controlling the size of residual updates can then
force the effective learning rate for individual hidden features
toward zero, approaching the semi-freeze boundary in the right panel of
Fig.~\ref{fig:scaling-phase}
(Appendix~\ref{app:linear-local-learning}).
More structured recipes, such as correlated initialization, can instead
preserve both feature learning and random depth increments.

Another question is whether it matters how $K=N_\sharp L$ is factored
into $N_\sharp$ and $L$, as our limiting description suggests that only
their product is essential.
In the ODE regime, this width--depth interchange follows naturally from
the framework of \citet{chizat2025hidden}.
Theoretical Result~\ref{thm:finite-errors} supports this picture by
identifying the contributions to the finite-size error.
Within this approximation, the same ordered hidden computations can be
combined into wider blocks or spread across more layers, like turning
melody into harmony and back.
We call this operation \emph{sliding}.

Measurements on practical networks such as GPT-2 show that contributions from randomly initialized residual blocks can dominate the input-embedding at initialization, making their scale important
for describing finite networks. Our fixed $\alpha$ scaling limit balances both of these contributions across scales. Our \emph{Half Copy}
recipe uses correlated initialization within conventional shapes to achieve a similar effect. In the CIFAR-10 and Tiny Shakespeare comparisons,
we conservatively estimate savings of $20\%$ of total parameters for SDE
and $40\%$ for Half Copy when matching or improving the ODE recipe's
mean training loss after fixed training schedules
(Appendix~\ref{app:parameter-savings}).
These gains do not settle the choice of depth: at matched parameter
budgets, using \emph{sliding} to combine computations into one wide block
gives the lowest training loss among the tested Tiny Shakespeare shapes.
 
\section{Joint scaling and limiting dynamics}
\label{sec:mlp-example}

We study how residual structure shapes learning, focusing on the residual
representation width $N_\flat$ and the associated population of
$K=N_\sharp L$ nonlinear channels.
These channels are arranged into $L$ layers, with $N_\sharp$ channels per layer.
The MLP structure in a Transformer provides a concrete example
(Fig.~\ref{fig:architecture-limit}, upper panel).
Initially omitting biases and normalization, we write its computation as
pseudocode on the left and in rescaled mathematical form on the right
(\texttt{W} absorbs the displayed scaling factors before $\boldsymbol W$, while $\boldsymbol W$
is initialized with independent standard Gaussian entries):

\begin{equation}
\label{eq:mlp-forward}
\begin{array}{@{}l@{\qquad}r@{}l@{}}
    \text{\texttt{h\_flat[0] = W\_in @ x}}
    & \bm h^{0\flat}
    & \displaystyle{}= \frac{1}{\sqrt{D_{\textrm{i}}}}\bm W^0\bm x,\\[5pt]
    \text{\texttt{h\_sharp[l] = W\_up[l] @ h\_flat[l]}}
    & \bm h^{\ell\sharp}
    & \displaystyle{}= \frac{1}{\sqrt{N_\flat}}\bm W^{\ell\uparrow}\bm h^{\ell\flat},\\[5pt]
    \begin{aligned}
        &\text{\texttt{h\_flat[l+1] = (h\_flat[l]}}\\[-2pt]
        &\quad\text{\texttt{+ W\_down[l] @ phi(h\_sharp[l]))}}
    \end{aligned}
    & \bm h^{(\ell+1)\flat}
    & \displaystyle{}= \bm h^{\ell\flat}
       + \frac{\beta_0\sqrt{N_\flat}}{L N_\sharp}
         \bm W^{\ell\downarrow}\phi\!\left(\bm h^{\ell\sharp}\right),\\[5pt]
    \text{\texttt{f = W\_out @ h\_flat[L]}}
    & \bm f
    & \displaystyle{}= \frac{1}{\gamma_0 N_\flat}\bm W^L\bm h^{L\flat}.
\end{array}
\end{equation}

We hold the input and output dimensions
$D_{\textrm i}$ and $D_{\textrm o}$, the data, loss, and batch sequence fixed
and compare networks at the same training step $t$.
Token indices and the internal coordinates of each attention head
likewise provide additional axes orthogonal to the network size $N_\flat, N_\sharp, L$ varying.
With a tensor interpretation of the variables and maps, the same residual
structure also accommodates attention and other nonlinear channel
computations (Appendix~\ref{app:finite-transformer}).

The output $\bm f$ is a macroscopic observable with fixed, task-defined
coordinates. Internal coordinates have no such prescribed roles.
Identity connections carry the same residual coordinates through depth,
whereas each layer introduces a new set of hidden coordinates.
A common rotation of the residual basis, accompanied by the corresponding
weight transformations, leaves the computation unchanged; coordinatewise
hidden nonlinearities instead select a preferred basis.
We track backward propagation through the rescaled Jacobians of each
output component:
\begin{equation}
\label{eq:residual-backward}
\bm g_q^{\ell\flat}=\gamma_0N_\flat
\frac{\partial f_q}{\partial\bm h^{\ell\flat}},\qquad
\bm g_q^{\ell\sharp}=\frac{\gamma_0K}{\beta_0}
\frac{\partial f_q}{\partial\bm h^{\ell\sharp}},
\quad q=1,\ldots,D_{\textrm o}.
\end{equation}
These factors compensate for the decreasing size of individual output
derivatives as the network grows.
Table~\ref{tab:optimizer-scaling} gives the initialization and optimizer
scalings for the code weights, likewise compensating for network growth
to retain order-one hidden-feature ($\bm h$ and $\bm g$) updates and their collective effect
on the output.
Optimizer conventions and the conversion to mathematical weights are
given in Appendix~\ref{app:optimizer-conventions}.

\begin{table}[t]
\centering
\setlength{\tabcolsep}{6pt}
\renewcommand{\arraystretch}{1.16}
\begin{tabular}{@{}lcccc@{}}
\hline
Quantity & Input & Up & Down & Output\\
\hline
Initialization std. & $1$ & $1/\sqrt{N_\flat}$ & $\sqrt{N_\flat}/K$ & $1/N_\flat$\\
SGD learning rate & $N_\flat$ & $K/N_\flat$ & $N_\flat/K$ & $1/N_\flat$\\
Adam learning rate & $1$ & $1/N_\flat$ & $1/K$ & $1/N_\flat$\\
Muon learning rate & $\sqrt{N_\flat}$ & $\sqrt{N_\sharp/N_\flat}$ & $\sqrt{N_\flat N_\sharp}/K$ & $1/\sqrt{N_\flat}$\\
Muon clipping threshold & $1/\sqrt{N_\flat}$ & $\sqrt{N_\flat N_\sharp}/K$ & $\sqrt{N_\sharp/N_\flat}$ & $\sqrt{N_\flat}$\\
\hline
\end{tabular}
\caption{Size dependence of initialization and updates
for code weights $\texttt{W}$ in Eq.~\ref{eq:mlp-forward}, with $K=N_\sharp L$. Fixed task dimensions and
multipliers, including $\beta_0,\gamma_0$, are omitted.
Here, Muon is understood as a normalized spectral-clipping map defined in
Appendix~\ref{app:optimizer-conventions}.}
\label{tab:optimizer-scaling}
\end{table}

As networks grow, statistics across the many coordinates of
$\bm h^{\flat}, \bm h^{\sharp}, \bm g^{\flat}, \bm g^{\sharp}$ are
captured by the joint probability law of representative residual
trajectories $(\bm{\mathscr h}^{\flat},\bm{\mathscr g}^{\flat})$
and hidden variables $(\bm{\mathscr h}^{\sharp},\bm{\mathscr g}^{\sharp})$
at each depth.
The residual trajectories remain discrete in depth at fixed $L$ and become
continuous trajectories indexed by $\tau=\ell/L\in[0,1]$ when width and
depth grow together (Appendix~\ref{app:limit-diagram}).

\begin{theorem}[Optimizer-dependent limiting dynamics]
\label{thm:joint-limit}
Consider networks trained with SGD, Muon, or Adam, using the scalings in
Table~\ref{tab:optimizer-scaling} and Appendix~\ref{app:optimizer-conventions},
keeping the remaining initialization and learning-rate constants fixed.
As $N_\flat,L\to\infty$ and $N_\flat/K\to\alpha$, the training dynamics
converge to a continuous-depth limit that generally depends on the optimizer.
For Muon, this limit additionally requires $N_\sharp\to\infty$.
In each case, the variables
$(\bm{\mathscr{h}}^\flat,\bm{\mathscr{g}}_q^\flat,
\bm{\mathscr{h}}^\sharp,\bm{\mathscr{g}}_q^\sharp)$
satisfy optimizer-dependent self-consistent equations
(system~\eqref{eq:sgd-depth-system} or its optimizer- and model-specific
variants in Appendix~\ref{app:sgd-dynamics}),
and the limiting macroscopic statistics (such as the output $\bm f$) are deterministic.

For SGD and Muon, the residual variables
$(\bm{\mathscr{h}}^\flat,\bm{\mathscr{g}}^\flat_q)$ remain jointly Gaussian across
depth, inputs, and training steps.
Adam does not generally preserve residual Gaussianity.
\end{theorem}

The statistical distinctions described above are visible
in standard models.
Figure~\ref{fig:gpt2-theory-illustration} provides an illustration
(detailed measurements follow in the experimental section): its right
panel shows approximately Gaussian residual alongside
non-Gaussian hidden pre-activations.
Full equations can be developed along the lines of the dynamical
mean-field formulation of \citet{bordelon2022selfconsistent}, or
equivalently the tensor-program formulation of \citet{yang2021feature}.
We leave those equations, including their tensor formulation for
Transformer attention, to Appendix~\ref{app:sgd-dynamics} and present
here only the minimal calculations needed to explain our picture.

\begin{theorem}[Zero- and one-step checkpoints]
\label{thm:linear-network}
At initialization, the limiting residual trajectory is a
continuous Gaussian random walk through depth with independent increments.
For the residual MLP structure used in Transformers, let $C_h(\tau;t)$
denote the limiting residual variance for a fixed input.
With $c=\beta_0^2\alpha$, the diffusion intensity is
\begin{equation}
\label{eq:initialization-checkpoint}
\begin{aligned}
\partial_\tau C_h(\tau;0)
&=\begin{cases}
c\,\mathbb E[\phi(\sqrt{C_h(\tau;0)}Z)^2],
    &\text{without RMSNorm},\\
c\,\mathbb E[\phi(Z)^2],
    &\text{with unit pre-RMSNorm},\ \epsilon_{\mathrm{norm}}=0.
\end{cases}
\end{aligned}
\end{equation}
Here $Z\sim\mathcal N(0,1)$, and
$C_h(0;0)=\|\bm x\|^2/D_{\mathrm i}$ is the variance contributed by
the input embedding; random residual increments supply the subsequent increase.
For $\phi = \mathrm{ReLU}$, these give $C_h(\tau;0)=C_h(0;0)e^{c\tau/2}$ and
$C_h(\tau;0)=C_h(0;0)+c\tau/2$, respectively
(Appendix~\ref{app:sgd-initialization}).

For its linear specialization
($\phi = \mathrm{Id}$, without normalization), take
$\|\bm x\|^2=D_{\mathrm i}$,
loss $\frac12(f-1)^2$, and frozen input and readout maps.
Starting from $C_h(\tau;0)=e^{c\tau}$, one simultaneous SGD step on
both residual projections with effective local learning rate
$\eta_{\rm eff}$ (Appendix~\ref{app:effective-step}) gives
\begin{equation}
\label{eq:linear-checkpoint}
\begin{aligned}
C_h(\tau;1)
&=e^{c\tau}+\eta_{\rm eff}^2e^{c(1+\tau)}
 \left[c\tau+2(2\beta_0^2+c^2)\tau^2
                  \left(1+\frac{2c\tau}{3}\right)\right].
\end{aligned}
\end{equation}
\end{theorem}

\begin{figure}[t]
\centering
\includegraphics{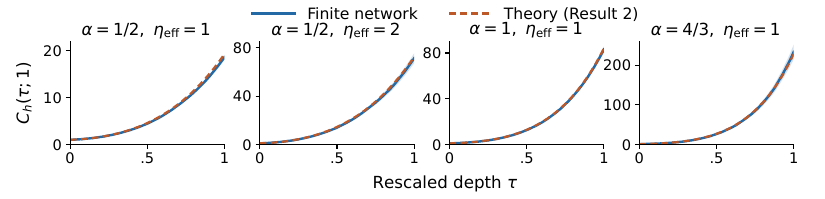}
\caption{One-step variance: finite-network means (solid) versus
Result~\ref{thm:linear-network} (dashed), with $\pm2$ Monte Carlo SE shaded;
$\beta_0=1$, $L=128$ (Appendix~\ref{app:linear-numerics}).}
\label{fig:linear-profiles-main}
\end{figure}

Figure~\ref{fig:linear-profiles-main} compares full finite-network updates
with \eqref{eq:linear-checkpoint}, showing close numerical agreement.

At initialization, comparing the contribution of Gaussian variance provides a practical
distinction between ODE-like behavior, where the input contribution
dominates, and SDE-like behavior, where residual random increments contribute
substantially.
Standard GPT-2 models clearly favor the latter
(Fig.~\ref{fig:gpt2-theory-illustration}, left; \citealp{karpathy2024buildnanogpt}).
Increasing $\beta_0$ as $\alpha$ decreases can preserve
$c=\beta_0^2\alpha>0$ and hence the initialization variance, but not
the training dynamics: even the first-step energy change grows as
$(\beta_0\eta_{\rm eff})^2$.
Controlling it requires $\eta_{\rm eff}=O(\beta_0^{-1})$ (the decreasing learning-rate ceiling illustrated in Fig.~\ref{fig:scaling-phase} (right)), weakening
direct local hidden-feature learning.
The calculation and finite-size checks are in
Appendix~\ref{app:single-linear}; the general nonlinear-channel
description of this degeneration is given in
Appendix~\ref{app:sgd-tensor-dmft}.

At fixed $N_\flat/K$, the same ordered hidden computations can be
regrouped across width and depth (\emph{sliding}), much like changing
the step size of an Euler discretization.
For example, label the projection matrices of two consecutive layers
by $1$ and $2$. Gluing them stacks the up projections and concatenates
the down projections, so both groups read the same residual state:
\begin{equation}
\label{eq:swing-gluing}
\begin{aligned}
\widetilde{\bm h}^{\flat}
&=\bm h^\flat+\frac{\beta_0\sqrt{N_\flat}}{K}
\begin{bmatrix}\bm W_1^\downarrow&\bm W_2^\downarrow\end{bmatrix}
\phi\!\left(\frac{1}{\sqrt{N_\flat}}
\begin{bmatrix}\bm W_1^\uparrow\\ \bm W_2^\uparrow\end{bmatrix}
\bm h^\flat\right).
\end{aligned}
\end{equation}
Pairwise gluing doubles $N_\sharp$ and halves $L$, leaving $K$ and
the coupling factor unchanged.
This regrouping of the same ordered weights is reversible.
Finite-size errors can be understood as
\begin{equation*}
\begin{gathered}
\text{Residual-kernel}\\
\text{fluctuations}
\end{gathered}
\;\;\;+\;\;\;
\begin{gathered}
\text{Euler}\\
\text{discretization}
\end{gathered}
\;\;\;+\;\;\;
\begin{gathered}
\text{Hidden-population}\\
\text{fluctuations}
\end{gathered}
\;\;\;+\;\;\;
\begin{gathered}
\text{Spectral bias}\\
\text{(Muon)}.
\end{gathered}
\end{equation*}
The following result gives their combined scales.

\begin{theorem}[Finite-size error scales]
\label{thm:finite-errors}
At a fixed training step, the root-mean-square error of a macroscopic
observable, such as the output $\bm f$, relative to its optimizer's limit
at $\alpha=N_\flat/K$ is
\begin{equation}
\label{eq:sgd-errors}
\begin{aligned}
&\text{SGD and Adam:}\quad
O\!\left(\frac{1}{\sqrt{N_\flat}}+\frac1L+
                  \frac{1}{\sqrt{N_\sharp L}}\right),\qquad
\text{Muon:}\quad
O\!\left(\frac{1}{\sqrt{N_\flat}}+\frac1L+
                  \frac1{N_\sharp}\right).
\end{aligned}
\end{equation}
Constants depend on the task, training step, and optimizer settings.
Details are in Appendix~\ref{app:macroscopic-errors}.
\end{theorem}

\section{Experiments}
\label{sec:experiments}

\begin{figure}[t]
\centering
\includegraphics{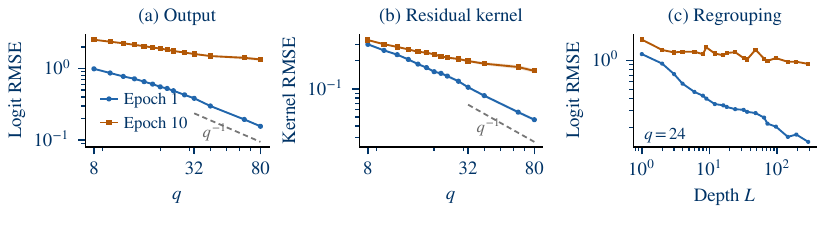}
\caption{CIFAR-10 convergence: output and final-residual cosine-kernel
errors against $q=96$ ensemble means (left/middle; bands: $\pm2$ conditional SE;
dashed: $q^{-1}$), and coupled regrouping against the finest finite partition
at $N_\flat=K=24^2$ (right).}
\label{fig:cifar-convergence}
\end{figure}

\subsection{Do the networks converge to the predicted limit?}

Output and kernel comparisons provide effective empirical checks
of a proposed scaling limit
\citep{bordelon2022selfconsistent,bordelon2023dynamics,bordelon2024depthwise,bordelon2024infinite}.
We use CIFAR-10 residual MLPs because their scalar hidden units permit
flexible regrouping, allowing large depths at modest parameter counts.
Under joint growth with $N_\flat=q^2$ and $N_\sharp=L=q$,
output and final-residual cosine-kernel errors relative to the
largest-network ensemble decrease with $q$, both early and later in training
(Fig.~\ref{fig:cifar-convergence}, left and middle).
Along this family, Theoretical
Result~\ref{thm:finite-errors} predicts $O(q^{-1})$ errors.
At epoch one, slopes fitted over $q=22$--$80$ are $-0.95$ and $-0.90$,
respectively, quantitatively consistent with this rate at short
training times.
Here we test the reverse of the gluing in Eq.~\ref{eq:swing-gluing}:
wider blocks are split into more, narrower
layers, keeping $N_\flat=K$ and the ordered hidden computations fixed.
Successive groups then read updated residual states.
Output discrepancies from the finest finite partition generally
decrease with depth (Fig.~\ref{fig:cifar-convergence}, right).
Here $N_\flat=K$ is fixed, so the sampling terms in our error estimate
do not decrease with depth, and a clean $L^{-1}$ discretization rate
need not be visible.
Further checks and protocols are given in Appendix~\ref{app:l01-cifar}.
\par

\subsection{What should a scaling recipe preserve?}

\begin{figure}[t]
\centering
\includegraphics{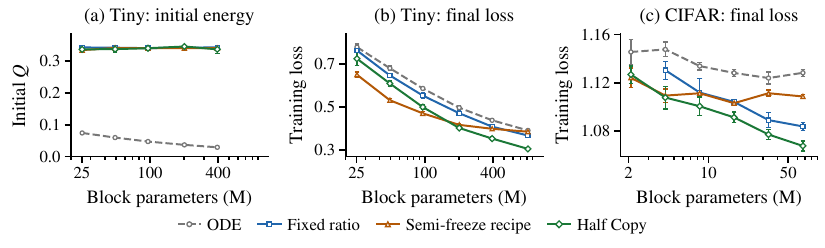}
\caption{Larger-scale comparisons, counting block weights only.
Tiny: initial $Q$ and loss after 5,000 Adam steps; CIFAR-10: loss after five
SGD epochs. Means $\pm$ sample SD over three seeds; full ranges:
Appendices~\ref{app:cifar-growth-recipes},~\ref{app:shakespeare-protocol}.}
\label{fig:scaling-recipes}
\end{figure}

\begin{figure}[!t]
\centering
\includegraphics{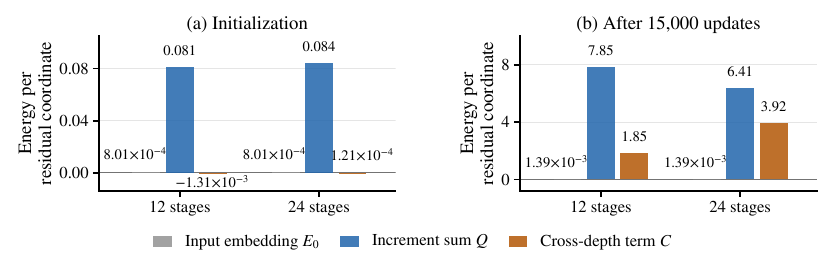}
\caption{Residual energy before and after training.
For complete attention--MLP stage increments $\Delta\bm h_\ell$,
$E_0=\|\bm h_0\|^2/N_\flat$,
$Q=\sum_\ell\|\Delta\bm h_\ell\|^2/N_\flat$, and
$C=\|\bm h_L-\bm h_0\|^2/N_\flat-Q$.
Each quantity is computed per token and averaged over the same four
1024-token sequences. The signed cross-depth term $C$ is retained;
the panels use different vertical scales
(Appendix~\ref{app:gpt2-energy}).}
\label{fig:gpt2-residual-energy}
\end{figure}

A scaling recipe should account for both the magnitude of the Gaussian
residual contribution (feature variance) and the effective local learning
rate (feature learning).
We compare scaling routes for Tiny Shakespeare Transformers and CIFAR-10
residual MLPs, starting from a common small-network anchor within each
task, tuned one hyperparameter at a time.
Full protocols are given in
Appendices~\ref{app:cifar-growth-recipes} and \ref{app:shakespeare-protocol}.
While our computational budget limits large-scale
sweeps, measurements from individual GPT-2 runs using
initialization and learning-rate settings in \citet{karpathy2024buildnanogpt} without re-tuning also support the SDE picture
(Fig.~\ref{fig:gpt2-residual-energy}).
At initialization, the cumulative squared residual increments $Q$ are
about one hundred times the input-embedding energy $E_0$:
random depth variation is a dominant contribution.
After training, the squared total displacement exceeds $Q$ by a
substantial positive cross-depth term $C$, indicating learned
correlations alongside the still-large increment energy.
A useful statistical description must account for both.

The ODE recipe follows Table~\ref{tab:optimizer-scaling} while increasing
both widths and depth by a factor $R$, so $N_\flat/K$ decreases as $R^{-1}$.
The semi-freeze recipe uses the same shapes but increases $\beta_0$ by
$\sqrt R$ to compensate for this loss of initial random variance,
at the cost of reducing the effective learning rate of individual hidden
features, as illustrated by Theoretical Result~\ref{thm:linear-network}.
The SDE recipe instead keeps the coupling and learning-rate constants
fixed and reallocates dimensions to preserve $N_\flat/K$.
For Tiny Shakespeare, growth factors $(R^{3/2},R^{1/2},R)$ for
$(N_\flat,N_\sharp,L)$ match both the ODE depth and its $R^3$
residual-block parameter budget.
Attention heads follow the same allocation principle at fixed head
dimension; task-specific geometries are given in
Appendices~\ref{app:cifar-growth-recipes} and \ref{app:shakespeare-protocol}.

Despite its unusual width--depth allocation, the SDE recipe achieves
lower mean training loss than ODE on both tasks at larger scales
(Fig.~\ref{fig:scaling-recipes}).
Even in the displayed examples, where semi-freeze initially leads,
continued growth reverses its advantage over SDE.
The CIFAR curve additionally shows an increase in loss with model size.

No single budget can match every resource across differently shaped
networks.
We use residual-block parameter count as our common comparison axis,
matching budgets proportional to $R^3$ and counting both MLP and
attention projections in Transformers.
For the Transformer allocation above, ignoring integer rounding gives
the following ratios for $R>1$ at fixed batch size and precision:
\begin{center}
\begin{tabular}{@{}lcc@{}}
\toprule
Resource & SDE / ODE (or semi-freeze) & Higher cost\\
\midrule
Residual-block parameters & $1$ & Equal\\
Input/output parameters and weight storage & $\sqrt R$ & SDE\\
KV cache at fixed sequence length & $1/\sqrt{R}$ & ODE\\
\bottomrule
\end{tabular}
\end{center}
Endpoint parameters become a smaller fraction of the total as scale grows.
In the largest Tiny comparison, rounding even leaves SDE with slightly
fewer total parameters (814.81M versus 815.22M), yet its loss is lower
(0.3680 versus 0.3916).
The KV-cache difference need not be small.
The corresponding result in residual MLPs, which have no attention or
KV cache, shows that the tendency is not confined to token-based models.

We propose the \emph{Half Copy} recipe to retain conventional network shapes while
maintaining both the initial Gaussian residual contribution and the
effective local learning-rate scale.
In both CIFAR-10 and Tiny Shakespeare, it outperforms the fixed-ratio
SDE and ODE recipes (Fig.~\ref{fig:scaling-recipes}).
Against ODE, geometry, coupling, and learning-rate constants are unchanged:
this isolates initialization from shape and opens another direction
for scaling-recipe design.

Consider doubling $N_\sharp$, and hence $K$, at fixed $N_\flat,L,\beta_0$.
Copying each hidden unit's up- and down-projections within its layer
exactly cancels the halved $1/K$ coefficient.
This preserves the random residual variance, but only trivially:
twice as many hidden parameters compute exactly the same function.
Half Copy instead uses larger, partly correlated groups.
For a doubling, four units with identical up-projections and
down-projection correlation $\rho=1/3$ supply the same variance factor
as two identical copies:
\begin{equation}
\label{eq:half-copy-main}
\begin{aligned}
U_a=U_{p(a)},\qquad
V_a=\sqrt{\rho}\,Z_{p(a)}+\sqrt{1-\rho}\,E_a,
\qquad \rho=\frac13.
\end{aligned}
\end{equation}
Here $p(a)$ labels a layer-local group; parent weights and innovations
are independent standard Gaussians.
Each weight keeps its Gaussian marginal, while independent innovations
allow the separately trainable copies to develop distinct features
(Appendix~\ref{app:half-copy-unlocking}).
With both widths and depth growing by a factor $R$ and the scalings of
Table~\ref{tab:optimizer-scaling}, groups of approximately $3R$ units
compensate for the falling $N_\flat/K$; attention heads are grouped
analogously.
Correlations remain local to each layer, while the total number of
independent parent groups grows with $N_\flat$
(Appendix~\ref{app:half-copy-recipe}).
\par\smallskip
\noindent\makebox[\linewidth][c]{%
\includegraphics{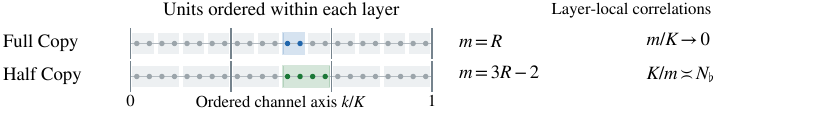}}
\par\smallskip

\subsection{What shape is actually best? Depends on the task}

\begin{figure}[!t]
\centering
\includegraphics{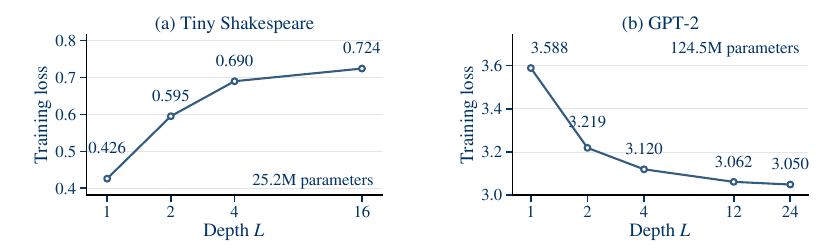}
\caption{Width--depth allocation and finite performance.
Left: Tiny Shakespeare training loss on 512 fixed sequences
at a fixed parameter budget, after 5,000 Adam updates
(Appendix~\ref{app:shakespeare-depth}).
Right: GPT-2 training loss averaged over the final 100 updates
at matched parameter and token budgets
(Appendix~\ref{app:gpt2-shape}).}
\label{fig:width-depth-allocation}
\end{figure}

\citet{chizat2025hidden} derives sharp error estimates for approximating
the ODE limit, offering a way to guide width--depth allocation.
Our estimate (Theoretical Result~\ref{thm:finite-errors})
invites the same exercise: balancing its three contributions under a
compute budget $C\propto N_\flat N_\sharp L$ suggests a scaling recipe,
\begin{equation}
\label{eq:balanced-scaling-recipe}
\begin{aligned}
\text{error proxy:}\quad&
\frac{1}{\sqrt{N_\flat}}+\frac1L+
\frac{1}{\sqrt{N_\sharp L}},
\qquad
\text{balanced recipe:}\quad&
N_\sharp\asymp L\asymp\sqrt{N_\flat}.
\end{aligned}
\end{equation}
But a smaller approximation error need not buy a lower loss, and our
SDE picture gets no exemption.
Tiny Shakespeare supplies the punchline: in the single-seed Half Copy scan,
\emph{sliding} to one block gives the lowest training loss at fixed parameter
budget, with layer-local correlation groups rebuilt at each depth
(Fig.~\ref{fig:width-depth-allocation}, left).
There is no contradiction: shallow, wide networks have their own limits,
obtained at fixed depth along the way to our continuous-depth description
(Appendix~\ref{app:sgd-dynamics}).
Continuous depth has no automatic claim to better performance.
Nor does one block: in our GPT-2 experiment, twelve or twenty-four stages
substantially outperform one at matched parameter and token budgets
(Fig.~\ref{fig:width-depth-allocation}, right).
Although we have some theoretical guidance and intuition for when
random residual variation makes an SDE description useful, we still
lack a theory, or even a reliable intuition, for when depth should beat width.

\section{Conclusion}
\label{sec:conclusion}

We identify joint width--depth scalings that retain local hidden-feature
learning together with random residual variation.
Our measurements show that this variation can be a major contribution
to residual features at initialization.
Half Copy illustrates how initialization correlations can offer an
alternative to unusual width--depth allocations.

Our analysis keeps the data and training horizon fixed as model size
grows and centers on dense MLP residual blocks.
Extending it to sparse mixture-of-experts models may build on the
scaling analysis of \citet{jiang2026hyperparameter}.
Training times growing with model size may require more than limiting
dynamics and leading finite-size corrections
\citep{bordelon2023dynamics,bordelon2024dynamical}.
Our compute budget also limits the experimental scope; substantially
larger-scale comparisons are needed to assess whether these ideas lead
to more compute-efficient Transformer training.

Limits provide a language for comparing network statistics across sizes,
but neither closed equations nor convergence guarantee better
finite-model performance.
Our SDE family faces the same test as ODE limits.
Finite-network behavior should guide which statistics a useful limit
retains, rather than be altered merely to fit a convenient limit.
This perspective calls for initialization and scaling recipes that help
finite models approach useful limits efficiently, with their value assessed through finite-model performance.

\section*{Reproducibility statement}
The appendices give the model, optimizer, initialization, data, and
measurement conventions for each experiment. Numerical data accompany
the displayed curves and tables. Controlled calculations include
independent finite-step checks, and empirical comparisons specify
their seeds, finite references, and uncertainty summaries.
The language-model distribution measurements use fixed probes across
training, with raw residual and output-Jacobian vectors retained
separately from their standardized plots.

\section*{Statement on AI assistance}
AI assistants were used for literature checks, drafting and revising
proofs, mathematical and numerical verification, experiment design,
implementation and analysis, figure generation, translation, and
manuscript preparation. They also supported long-running experiments
through periodic automated check-ins (``heartbeats'') and unattended
workflows, including progress monitoring, fault diagnosis and recovery,
and checkpoint preservation, within author-defined protocols and
resource budgets.
The authors are responsible for the claims, interpretation,
and submitted manuscript.

\section*{Acknowledgements} DZ has received funding from the European Union’s Horizon 2020 research and innovation programme under the Marie Skłodowska-Curie grant agreement No. 101034255. BB is supported by the University of Texas at Austin Oden Institute for Computational Engineering and Sciences and Department of Neuroscience.  We would like to thank Louis Pierre Chaintron, Lenaic Chizat, Javier Maass, Mufan (Bill) Li, Boris Hanin, Zhen Yang, and Samet Demir for inspiring and insightful discussions on depth limits of residual networks. We thank the organizers of the ProbAI Theory of Scaling Laws Workshop 2026 where some of this paper's results were worked out.
\bibliography{sliding_networks}

@inproceedings{schoenholz2017deep,
  title     = {Deep Information Propagation},
  author    = {Schoenholz, Samuel S. and Gilmer, Justin and Ganguli, Surya and Sohl-Dickstein, Jascha},
  booktitle = {International Conference on Learning Representations},
  year      = {2017},
  url       = {https://arxiv.org/abs/1611.01232}
}

@inproceedings{hayou2021stable,
  title     = {Stable {ResNet}},
  author    = {Hayou, Soufiane and Clerico, Eugenio and He, Bobby and Deligiannidis, George and Doucet, Arnaud and Rousseau, Judith},
  booktitle = {Proceedings of The 24th International Conference on Artificial Intelligence and Statistics},
  series    = {Proceedings of Machine Learning Research},
  volume    = {130},
  pages     = {1324--1332},
  year      = {2021},
  url       = {https://proceedings.mlr.press/v130/hayou21a.html}
}

@inproceedings{noci2023shaped,
  title     = {The Shaped Transformer: Attention Models in the Infinite Depth-and-Width Limit},
  author    = {Noci, Lorenzo and Li, Chuning and Li, Mufan Bill and He, Bobby and Hofmann, Thomas and Maddison, Chris J. and Roy, Daniel M.},
  booktitle = {Advances in Neural Information Processing Systems},
  volume    = {36},
  year      = {2023},
  url       = {https://proceedings.neurips.cc/paper_files/paper/2023/hash/aa31dc84098add7dd2ffdd20646f2043-Abstract-Conference.html}
}

@inproceedings{bordelon2024depthwise,
  title     = {Depthwise Hyperparameter Transfer in Residual Networks: Dynamics and Scaling Limit},
  author    = {Bordelon, Blake and Noci, Lorenzo and Li, Mufan and Hanin, Boris and Pehlevan, Cengiz},
  booktitle = {International Conference on Learning Representations},
  year      = {2024},
  url       = {https://proceedings.iclr.cc/paper_files/paper/2024/hash/5f5f7b6080dcadced61cf5d96f7c6dde-Abstract-Conference.html}
}

@inproceedings{bordelon2023dynamics,
  title     = {Dynamics of Finite Width Kernel and Prediction Fluctuations in Mean Field Neural Networks},
  author    = {Bordelon, Blake and Pehlevan, Cengiz},
  booktitle = {Advances in Neural Information Processing Systems},
  volume    = {36},
  year      = {2023},
  url       = {https://proceedings.neurips.cc/paper_files/paper/2023/hash/1ec69275e9f002ee068f5d68380f3290-Abstract-Conference.html}
}

@inproceedings{bordelon2024dynamical,
  title     = {A Dynamical Model of Neural Scaling Laws},
  author    = {Bordelon, Blake and Atanasov, Alexander and Pehlevan, Cengiz},
  booktitle = {Proceedings of the 41st International Conference on Machine Learning},
  series    = {Proceedings of Machine Learning Research},
  volume    = {235},
  pages     = {4345--4382},
  year      = {2024},
  url       = {https://proceedings.mlr.press/v235/bordelon24a.html}
}

@misc{chaintron2026resnets,
  title         = {{ResNets} of All Shapes and Sizes: Convergence of Training Dynamics in the Large-scale Limit},
  author        = {Chaintron, Louis-Pierre and Chizat, L\'{e}na\"{i}c and Maass, Javier},
  year          = {2026},
  eprint        = {2603.18168},
  archivePrefix = {arXiv},
  primaryClass  = {stat.ML},
  url           = {https://arxiv.org/abs/2603.18168}
}

@misc{jiang2026hyperparameter,
  title         = {Hyperparameter Transfer with Mixture-of-Expert Layers},
  author        = {Jiang, Tianze and Bordelon, Blake and Pehlevan, Cengiz and Hanin, Boris},
  year          = {2026},
  eprint        = {2601.20205},
  archivePrefix = {arXiv},
  primaryClass  = {cs.LG},
  note          = {ICML 2026},
  url           = {https://arxiv.org/abs/2601.20205v3}
}

@inproceedings{tay2022scale,
  title     = {Scale Efficiently: Insights from Pre-training and Fine-tuning Transformers},
  author    = {Tay, Yi and Dehghani, Mostafa and Rao, Jinfeng and Fedus, William and
               Abnar, Samira and Chung, Hyung Won and Narang, Sharan and Yogatama, Dani and
               Vaswani, Ashish and Metzler, Donald},
  booktitle = {International Conference on Learning Representations},
  year      = {2022},
  url       = {https://arxiv.org/abs/2109.10686}
}

@inproceedings{petty2024impact,
  title     = {The Impact of Depth on Compositional Generalization in Transformer Language Models},
  author    = {Petty, Jackson and van Steenkiste, Sjoerd and Dasgupta, Ishita and
               Sha, Fei and Garrette, Dan and Linzen, Tal},
  booktitle = {Proceedings of the 2024 Conference of the North American Chapter of the
               Association for Computational Linguistics: Human Language Technologies
               (Volume 1: Long Papers)},
  publisher = {Association for Computational Linguistics},
  pages     = {7239--7252},
  year      = {2024},
  doi       = {10.18653/v1/2024.naacl-long.402},
  url       = {https://aclanthology.org/2024.naacl-long.402/}
}

@misc{brown2022wide,
  title         = {Wide Attention Is The Way Forward For Transformers?},
  author        = {Brown, Jason Ross and Zhao, Yiren and Shumailov, Ilia and Mullins, Robert D.},
  year          = {2022},
  eprint        = {2210.00640},
  archivePrefix = {arXiv},
  primaryClass  = {cs.LG},
  note          = {Presented at the NeurIPS 2022 Workshop on All Things Attention: Bridging Different Perspectives on Attention},
  url           = {https://arxiv.org/abs/2210.00640}
}

@inproceedings{sutskever2013importance,
  title     = {On the Importance of Initialization and Momentum in Deep Learning},
  author    = {Sutskever, Ilya and Martens, James and Dahl, George and Hinton, Geoffrey},
  booktitle = {Proceedings of the 30th International Conference on Machine Learning},
  series    = {Proceedings of Machine Learning Research},
  volume    = {28},
  pages     = {1139--1147},
  year      = {2013},
  url       = {https://proceedings.mlr.press/v28/sutskever13.html}
}

@article{bergstra2012random,
  title   = {Random Search for Hyper-Parameter Optimization},
  author  = {Bergstra, James and Bengio, Yoshua},
  journal = {Journal of Machine Learning Research},
  volume  = {13},
  number  = {10},
  pages   = {281--305},
  year    = {2012},
  url     = {https://jmlr.org/papers/v13/bergstra12a.html}
}

@article{zhao2026survey,
  title   = {A Survey of Large Language Models},
  author  = {Zhao, Wayne Xin and Zhou, Kun and Li, Junyi and Tang, Tianyi and
             Dong, Zican and Hou, Yupeng and Zhang, Beichen and Min, Yingqian and
             Zhang, Junjie and Liu, Peiyu and Wang, Xiaolei and Du, Yifan and
             Yang, Chen and Chen, Yushuo and Chen, Zhipeng and Jiang, Jinhao and
             Ren, Ruiyang and Li, Yifan and Tang, Xinyu and Liu, Zikang and
             Hu, Yiwen and Nie, Jian-Yun and Wen, Ji-Rong},
  journal = {Frontiers of Computer Science},
  volume  = {20},
  pages   = {2012627},
  year    = {2026},
  doi     = {10.1007/s11704-026-60308-3},
  url     = {https://link.springer.com/article/10.1007/s11704-026-60308-3}
}

@article{lecun2015deep,
  title   = {Deep learning},
  author  = {LeCun, Yann and Bengio, Yoshua and Hinton, Geoffrey},
  journal = {Nature},
  volume  = {521},
  number  = {7553},
  pages   = {436--444},
  year    = {2015},
  doi     = {10.1038/nature14539},
  url     = {https://www.nature.com/articles/nature14539}
}

@inproceedings{krizhevsky2012imagenet,
  title     = {{ImageNet} Classification with Deep Convolutional Neural Networks},
  author    = {Krizhevsky, Alex and Sutskever, Ilya and Hinton, Geoffrey E.},
  booktitle = {Advances in Neural Information Processing Systems},
  volume    = {25},
  year      = {2012},
  url       = {https://papers.nips.cc/paper_files/paper/2012/hash/c399862d3b9d6b76c8436e924a68c45b-Abstract.html}
}

@misc{kaplan2020scaling,
  title         = {Scaling Laws for Neural Language Models},
  author        = {Kaplan, Jared and McCandlish, Sam and Henighan, Tom and
                   Brown, Tom B. and Chess, Benjamin and Child, Rewon and
                   Gray, Scott and Radford, Alec and Wu, Jeffrey and Amodei, Dario},
  year          = {2020},
  eprint        = {2001.08361},
  archivePrefix = {arXiv},
  primaryClass  = {cs.LG},
  url           = {https://arxiv.org/abs/2001.08361}
}

@inproceedings{hoffmann2022training,
  title     = {An empirical analysis of compute-optimal large language model training},
  author    = {Hoffmann, Jordan and Borgeaud, Sebastian and Mensch, Arthur and
               Buchatskaya, Elena and Cai, Trevor and Rutherford, Eliza and
               de Las Casas, Diego and Hendricks, Lisa Anne and Welbl, Johannes and
               Clark, Aidan and Hennigan, Thomas and Noland, Eric and Millican, Katherine and
               van den Driessche, George and Damoc, Bogdan and Guy, Aurelia and
               Osindero, Simon and Simonyan, Kar{\'e}n and Elsen, Erich and
               Vinyals, Oriol and Rae, Jack and Sifre, Laurent},
  booktitle = {Advances in Neural Information Processing Systems},
  volume    = {35},
  pages     = {30016--30030},
  year      = {2022},
  doi       = {10.52202/068431-2176},
  url       = {https://proceedings.neurips.cc/paper_files/paper/2022/hash/c1e2faff6f588870935f114ebe04a3e5-Abstract.html}
}

@inproceedings{jacot2018neural,
  title     = {Neural Tangent Kernel: Convergence and Generalization in Neural Networks},
  author    = {Jacot, Arthur and Gabriel, Franck and Hongler, Cl\'{e}ment},
  booktitle = {Advances in Neural Information Processing Systems},
  volume    = {31},
  pages     = {8571--8580},
  year      = {2018},
  url       = {https://arxiv.org/abs/1806.07572}
}

@inproceedings{lee2019wide,
  title     = {Wide Neural Networks of Any Depth Evolve as Linear Models Under Gradient Descent},
  author    = {Lee, Jaehoon and Xiao, Lechao and Schoenholz, Samuel S. and
               Bahri, Yasaman and Novak, Roman and Sohl-Dickstein, Jascha and
               Pennington, Jeffrey},
  booktitle = {Advances in Neural Information Processing Systems},
  volume    = {32},
  year      = {2019},
  url       = {https://arxiv.org/abs/1902.06720}
}

@article{mei2018meanfield,
  title   = {A mean field view of the landscape of two-layer neural networks},
  author  = {Mei, Song and Montanari, Andrea and Nguyen, Phan-Minh},
  journal = {Proceedings of the National Academy of Sciences},
  volume  = {115},
  number  = {33},
  pages   = {E7665--E7671},
  year    = {2018},
  doi     = {10.1073/pnas.1806579115},
  url     = {https://arxiv.org/abs/1804.06561}
}

@inproceedings{chizat2018global,
  title     = {On the Global Convergence of Gradient Descent for Over-parameterized Models using Optimal Transport},
  author    = {Chizat, L\'{e}na\"{i}c and Bach, Francis},
  booktitle = {Advances in Neural Information Processing Systems},
  volume    = {31},
  year      = {2018},
  url       = {https://arxiv.org/abs/1805.09545}
}

@article{sirignano2020meanfield,
  title   = {Mean Field Analysis of Neural Networks: A Law of Large Numbers},
  author  = {Sirignano, Justin and Spiliopoulos, Konstantinos},
  journal = {SIAM Journal on Applied Mathematics},
  volume  = {80},
  number  = {2},
  pages   = {725--752},
  year    = {2020},
  doi     = {10.1137/18M1192184},
  url     = {https://arxiv.org/abs/1805.01053}
}

@inproceedings{yang2021feature,
  title     = {Tensor Programs {IV}: Feature Learning in Infinite-Width Neural Networks},
  author    = {Yang, Greg and Hu, Edward J.},
  booktitle = {Proceedings of the 38th International Conference on Machine Learning},
  volume    = {139},
  pages     = {11727--11737},
  publisher = {PMLR},
  year      = {2021},
  url       = {https://proceedings.mlr.press/v139/yang21c.html}
}

@inproceedings{bordelon2022selfconsistent,
  title     = {Self-Consistent Dynamical Field Theory of Kernel Evolution in Wide Neural Networks},
  author    = {Bordelon, Blake and Pehlevan, Cengiz},
  booktitle = {Advances in Neural Information Processing Systems},
  volume    = {35},
  pages     = {32240--32256},
  year      = {2022},
  url       = {https://arxiv.org/abs/2205.09653}
}

@inproceedings{bordelon2024infinite,
  title     = {Infinite Limits of Multi-head Transformer Dynamics},
  author    = {Bordelon, Blake and Chaudhry, Hamza Tahir and Pehlevan, Cengiz},
  booktitle = {Advances in Neural Information Processing Systems},
  volume    = {37},
  year      = {2024},
  doi       = {10.52202/079017-1130},
  url       = {https://papers.neurips.cc/paper_files/paper/2024/hash/3eff068e195daace49955348de9f8398-Abstract-Conference.html}
}

@misc{chizat2025hidden,
  title         = {The Hidden Width of Deep {ResNets}: Tight Error Bounds and Phase Diagram},
  author        = {Chizat, L\'ena\"{i}c},
  year          = {2025},
  eprint        = {2509.10167},
  archivePrefix = {arXiv},
  primaryClass  = {cs.LG},
  url           = {https://arxiv.org/abs/2509.10167}
}

@inproceedings{yang2021tuning,
  title     = {Tuning Large Neural Networks via Zero-Shot Hyperparameter Transfer},
  author    = {Yang, Greg and Hu, Edward J. and Babuschkin, Igor and Sidor, Szymon and
               Liu, Xiaodong and Farhi, David and Ryder, Nick and Pachocki, Jakub and
               Chen, Weizhu and Gao, Jianfeng},
  booktitle = {Advances in Neural Information Processing Systems},
  volume    = {34},
  year      = {2021},
  url       = {https://papers.nips.cc/paper_files/paper/2021/hash/8df7c2e3c3c3be098ef7b382bd2c37ba-Abstract.html}
}

@inproceedings{dey2025completep,
  title     = {Don't be lazy: {CompleteP} enables compute-efficient deep transformers},
  author    = {Dey, Nolan and Zhang, Bin Claire and Noci, Lorenzo and Li, Mufan and
               Bordelon, Blake and Bergsma, Shane and Pehlevan, Cengiz and
               Hanin, Boris and Hestness, Joel},
  booktitle = {Advances in Neural Information Processing Systems},
  year      = {2025},
  url       = {https://papers.nips.cc/paper_files/paper/2025/hash/c917d8b9e01427f3184d80ade22f4d1f-Abstract-Conference.html}
}

@inproceedings{he2015deep,
  title     = {Deep Residual Learning for Image Recognition},
  author    = {He, Kaiming and Zhang, Xiangyu and Ren, Shaoqing and Sun, Jian},
  booktitle = {Proceedings of the IEEE Conference on Computer Vision and Pattern Recognition (CVPR)},
  pages     = {770--778},
  year      = {2016},
  url       = {https://openaccess.thecvf.com/content_cvpr_2016/html/He_Deep_Residual_Learning_CVPR_2016_paper.html}
}

@inproceedings{vaswani2017attention,
  title     = {Attention Is All You Need},
  author    = {Vaswani, Ashish and Shazeer, Noam and Parmar, Niki and
               Uszkoreit, Jakob and Jones, Llion and Gomez, Aidan N. and
               Kaiser, {\L}ukasz and Polosukhin, Illia},
  booktitle = {Advances in Neural Information Processing Systems},
  volume    = {30},
  year      = {2017},
  url       = {https://papers.nips.cc/paper/2017/hash/3f5ee243547dee91fbd053c1c4a845aa-Abstract.html}
}

@inproceedings{yang2023tensorvi,
  title     = {Tensor Programs {VI}: Feature Learning in Infinite Depth Neural Networks},
  author    = {Yang, Greg and Yu, Dingli and Zhu, Chen and Hayou, Soufiane},
  booktitle = {International Conference on Learning Representations},
  year      = {2024},
  url       = {https://proceedings.iclr.cc/paper_files/paper/2024/hash/f2ce452a4a7352feb7a9fb885267ff24-Abstract-Conference.html}
}

@inproceedings{kingma2015adam,
  author = {Kingma, Diederik P. and Ba, Jimmy},
  title = {Adam: A Method for Stochastic Optimization},
  booktitle = {International Conference on Learning Representations},
  year = {2015},
  url = {https://arxiv.org/abs/1412.6980}
}

@misc{jordan2024muon,
  author = {Jordan, Keller and Jin, Yuchen and Boza, Vlado and You, Jiacheng and
            Cesista, Franz and Newhouse, Laker and Bernstein, Jeremy},
  title = {Muon: An Optimizer for Hidden Layers in Neural Networks},
  year = {2024},
  url = {https://kellerjordan.github.io/posts/muon/}
}

@misc{su2025higherordermup,
  author = {Su, Jianlin},
  title = {Higher-Order {MuP}: Simpler yet More Powerful Spectral-Condition Scaling},
  year = {2025},
  month = mar,
  howpublished = {Scientific Spaces},
  note = {Blog post in Chinese; title translated into English},
  url = {https://spaces.ac.cn/archives/10795}
}

@misc{liu2025muon,
  author = {Liu, Jingyuan and Su, Jianlin and Yao, Xingcheng and Jiang, Zhejun and
            Lai, Guokun and Du, Yulun and Qin, Yidao and Xu, Weixin and Lu, Enzhe and
            Yan, Junjie and Chen, Yanru and Zheng, Huabin and Liu, Yibo and
            Liu, Shaowei and Yin, Bohong and He, Weiran and Zhu, Han and
            Wang, Yuzhi and Wang, Jianzhou and Dong, Mengnan and Zhang, Zheng and
            Kang, Yongsheng and Zhang, Hao and Xu, Xinran and Zhang, Yutao and
            Wu, Yuxin and Zhou, Xinyu and Yang, Zhilin},
  title = {{Muon} is Scalable for {LLM} Training},
  year = {2025},
  eprint = {2502.16982},
  archivePrefix = {arXiv},
  primaryClass = {cs.LG},
  url = {https://arxiv.org/abs/2502.16982}
}

@misc{karpathy2024buildnanogpt,
  author = {Karpathy, Andrej},
  title = {{build-nanogpt}: Building {GPT-2} from Scratch},
  year = {2024},
  howpublished = {\url{https://github.com/karpathy/build-nanogpt}},
  note = {Commit 6104ab1b53920f6e2159749676073ff7d815c1fa}
}

@misc{lozhkov2024finewebedu,
  author = {Lozhkov, Anton and Ben Allal, Loubna and von Werra, Leandro and Wolf, Thomas},
  title = {{FineWeb-Edu}: The Finest Collection of Educational Content},
  year = {2024},
  publisher = {Hugging Face},
  doi = {10.57967/hf/2497},
  url = {https://huggingface.co/datasets/HuggingFaceFW/fineweb-edu}
}

@techreport{krizhevsky2009learning,
  title={Learning Multiple Layers of Features from Tiny Images},
  author={Krizhevsky, Alex},
  institution={University of Toronto},
  year={2009},
  url={https://www.cs.toronto.edu/~kriz/learning-features-2009-TR.pdf}
}

@inproceedings{zellers2019hellaswag,
  author = {Zellers, Rowan and Holtzman, Ari and Bisk, Yonatan and
            Farhadi, Ali and Choi, Yejin},
  title = {{HellaSwag}: Can a Machine Really Finish Your Sentence?},
  booktitle = {Proceedings of the 57th Annual Meeting of the Association for Computational Linguistics},
  year = {2019},
  pages = {4791--4800},
  publisher = {Association for Computational Linguistics},
  doi = {10.18653/v1/P19-1472},
  url = {https://aclanthology.org/P19-1472/}
}

@misc{karpathy2015charrnn,
  author = {Karpathy, Andrej},
  title = {{char-rnn}: Multi-layer Recurrent Neural Networks for Character-level Language Models},
  year = {2015},
  howpublished = {GitHub repository},
  note = {Tiny Shakespeare dataset},
  url = {https://github.com/karpathy/char-rnn}
}

@misc{jordan2024moddednanogpt,
  author = {Jordan, Keller and others},
  title = {{modded-nanogpt}},
  year = {2024},
  howpublished = {GitHub repository},
  note = {2024-10-10 Muon training record},
  url = {https://github.com/KellerJordan/modded-nanogpt}
}

@techreport{radford2019language,
  author = {Radford, Alec and Wu, Jeffrey and Child, Rewon and Luan, David and
            Amodei, Dario and Sutskever, Ilya},
  title = {Language Models are Unsupervised Multitask Learners},
  institution = {OpenAI},
  year = {2019},
  url = {https://cdn.openai.com/better-language-models/language_models_are_unsupervised_multitask_learners.pdf}
}

@inproceedings{zhang2019rmsnorm,
  author = {Zhang, Biao and Sennrich, Rico},
  title = {Root Mean Square Layer Normalization},
  booktitle = {Advances in Neural Information Processing Systems},
  volume = {32},
  year = {2019},
  url = {https://arxiv.org/abs/1910.07467}
}

@article{su2021roformer,
  author = {Su, Jianlin and Ahmed, Murtadha and Lu, Yu and Pan, Shengfeng and
            Bo, Wen and Liu, Yunfeng},
  title = {{RoFormer}: Enhanced Transformer with Rotary Position Embedding},
  journal = {Neurocomputing},
  volume = {568},
  pages = {127063},
  year = {2024},
  doi = {10.1016/j.neucom.2023.127063},
  url = {https://doi.org/10.1016/j.neucom.2023.127063}
}

@misc{hendrycks2016gelu,
  author = {Hendrycks, Dan and Gimpel, Kevin},
  title = {Gaussian Error Linear Units ({GELUs})},
  year = {2016},
  eprint = {1606.08415},
  archivePrefix = {arXiv},
  url = {https://arxiv.org/abs/1606.08415}
}

@misc{ba2016layernorm,
  author = {Ba, Jimmy Lei and Kiros, Jamie Ryan and Hinton, Geoffrey E.},
  title = {Layer Normalization},
  year = {2016},
  eprint = {1607.06450},
  archivePrefix = {arXiv},
  url = {https://arxiv.org/abs/1607.06450}
}

@inproceedings{loshchilov2019decoupled,
  author = {Loshchilov, Ilya and Hutter, Frank},
  title = {Decoupled Weight Decay Regularization},
  booktitle = {International Conference on Learning Representations},
  year = {2019},
  url = {https://openreview.net/forum?id=Bkg6RiCqY7}
}

@inproceedings{paszke2019pytorch,
  author = {Paszke, Adam and Gross, Sam and Massa, Francisco and Lerer, Adam and
            Bradbury, James and Chanan, Gregory and Killeen, Trevor and Lin, Zeming and
            Gimelshein, Natalia and Antiga, Luca and Desmaison, Alban and Kopf, Andreas and
            Yang, Edward and DeVito, Zachary and Raison, Martin and Tejani, Alykhan and
            Chilamkurthy, Sasank and Steiner, Benoit and Fang, Lu and Bai, Junjie and
            Chintala, Soumith},
  title = {{PyTorch}: An Imperative Style, High-Performance Deep Learning Library},
  booktitle = {Advances in Neural Information Processing Systems},
  volume = {32},
  year = {2019},
  url = {https://arxiv.org/abs/1912.01703}
}

@misc{liu2026inversedepth,
  author = {Liu, Yizhou and Kangaslahti, Sara and Liu, Ziming and Gore, Jeff},
  title = {Inverse Depth Scaling From Most Layers Being Similar},
  year = {2026},
  eprint = {2602.05970},
  archivePrefix = {arXiv},
  note = {ICML 2026},
  url = {https://arxiv.org/abs/2602.05970}
}
\bibliographystyle{iclr2027_conference}
\clearpage
\appendix
\begingroup
\color{black}
\color{black}
\begingroup
\color{black}
\section{Finite-network and optimizer conventions}
\label{app:optimizer-conventions}
\label{sec:basic-optimizers}

\subsection{A Transformer as two populations of nonlinear channels}
\label{app:finite-transformer}
A Transformer alternates two kinds of nonlinear channel within the
same residual stream: attention heads and MLP units.
Consider a causal decoder with $L$ attention--MLP stages, residual width
$N_\flat$, $N_{\sharp\mathrm A}$ heads per stage, and $N_{\sharp\mathrm M}$
MLP units per stage.
Each head has fixed even internal dimension $d$, whereas the number of heads
can vary independently of the residual width.
There are $L$ updates of each kind, hence
$K_{\rm A}=LN_{\sharp\mathrm A}$ and $K_{\rm M}=LN_{\sharp\mathrm M}$;
the intermediate state $\bm h^{(\ell+1/2)\flat}$ records the attention
update before the MLP update.

The concrete model uses separate pre-RMSNorm before each branch,
rotary positional encoding in attention, GELU, and a final RMSNorm.
Its projections have no biases, the input and readout weights are
untied, and dropout is omitted.
For $S$ tokens, residual states are the columns of
$\bm h^{\ell\flat}\in\mathbb R^{N_\flat\times S}$.
Normalization acts on each column:
\begin{equation}
\label{eq:finite-transformer-norm}
\mathcal N(\bm h_{\mathfrak{s}})
=\frac{\bm h_{\mathfrak{s}}}{\sqrt{\|\bm h_{\mathfrak{s}}\|^2/N_\flat+\epsilon_{\rm norm}}},
\qquad \epsilon_{\rm norm}>0.
\end{equation}
The input columns $\bm x_{\mathfrak{s}}\in\mathbb R^{D_{\rm i}}$ encode tokens;
$\bm x_{\mathfrak{s}}=\sqrt{D_{\rm i}}\,\bm e_{\mathrm{token}(\mathfrak{s})}$ makes the input
map below an embedding lookup.
Position information is supplied by the rotary maps.

The full computation retains the code/mathematical-weight distinction
of Eq.~\ref{eq:mlp-forward}: each code matrix absorbs its displayed
coefficient, and the mathematical matrices have independent standard
Gaussian entries at initialization.
The first and last lines run once; the intervening lines repeat for
$\ell=0,\ldots,L-1$, using that stage's code weights.
The attention operations run over all heads $a=1,\ldots,N_{\sharp\mathrm A}$:
\begin{equation}
\label{eq:finite-transformer}
\begin{array}{@{}l@{\quad}l@{}}
\text{\texttt{h = Win @ x}}
&\displaystyle
\bm h^{0\flat}=\frac{1}{\sqrt{D_{\rm i}}}\bm W^0\bm x
\\[6pt]
\begin{aligned}
&\text{\texttt{q[a],k[a],v[a] =}}\\[-2pt]
&\quad\text{\texttt{split(Wqkv[a] @ rms(h))}}
\end{aligned}
&\displaystyle
\begin{pmatrix}\bm q_a\\ \bm k_a\\ \bm v_a\end{pmatrix}
=\frac{1}{\sqrt{N_\flat}}
\begin{pmatrix}\bm W_a^{\ell Q}\\ \bm W_a^{\ell K}\\ \bm W_a^{\ell V}\end{pmatrix}
\mathcal N(\bm h^{\ell\flat})
\\[6pt]
\begin{aligned}
&\text{\texttt{o[a] = attend(}}\\[-2pt]
&\quad\text{\texttt{q[a],k[a],v[a])}}
\end{aligned}
&\displaystyle \bm o_a=\bm v_aP_a^{\mathsf T}
\\[6pt]
\begin{aligned}
&\text{\texttt{hhalf = h}}\\[-2pt]
&\quad\text{\texttt{+ sum(WO[a] @ o[a])}}
\end{aligned}
&\displaystyle
\bm h^{(\ell+1/2)\flat}=\bm h^{\ell\flat}
+\frac{\beta_{0,\rm A}\sqrt{N_\flat}}{K_{\rm A}\sqrt d}
\sum_{a=1}^{N_{\sharp\mathrm A}}\bm W_a^{\ell O}\bm o_a
\\[6pt]
\text{\texttt{z = Wup @ rms(hhalf)}}
&\displaystyle
\bm h^{\ell\sharp\mathrm M}
=\frac{1}{\sqrt{N_\flat}}\bm W_{\rm M}^{\ell\uparrow}
\mathcal N(\bm h^{(\ell+1/2)\flat})
\\[6pt]
\text{\texttt{h = hhalf + Wdown @ gelu(z)}}
&\displaystyle
\bm h^{(\ell+1)\flat}=\bm h^{(\ell+1/2)\flat}
+\frac{\beta_{0,\rm M}\sqrt{N_\flat}}{K_{\rm M}}
\bm W_{\rm M}^{\ell\downarrow}
\operatorname{GELU}(\bm h^{\ell\sharp\mathrm M})
\\[6pt]
\text{\texttt{f = Wout @ rms(h)}}
&\displaystyle
\bm f=\frac{1}{\gamma_0N_\flat}
\bm W^L\mathcal N(\bm h^{L\flat}).
\end{array}
\end{equation}
Here $\bm q_a,\bm k_a,\bm v_a\in\mathbb R^{d\times S}$,
$\bm W_a^{\ell Q},\bm W_a^{\ell K},\bm W_a^{\ell V}
\in\mathbb R^{d\times N_\flat}$, and
$\bm W_a^{\ell O}\in\mathbb R^{N_\flat\times d}$.
The MLP up/down matrices have dimensions
$N_{\sharp\mathrm M}\times N_\flat$ and
$N_\flat\times N_{\sharp\mathrm M}$, respectively;
$\bm f\in\mathbb R^{D_{\rm o}\times S}$ contains the output logits.
The branch constants $\beta_{0,\rm A},\beta_{0,\rm M}$ play the role of
$\beta_0$ separately for attention and the MLP.

The attention operation mixes tokens within each head.
Let $R_{\mathfrak{s}}$ be the block-diagonal rotary matrix at token position $\mathfrak{s}$,
with two-dimensional rotations by $(\mathfrak{s}-1)\omega_j$ for fixed frequencies
$\omega_j$, $j=0,\ldots,d/2-1$.
In the line \texttt{attend}, the causal attention weights are
\begin{equation}
\label{eq:finite-transformer-attention}
[P_a]_{\mathfrak{s}\mathfrak{r}}=
\begin{cases}
\displaystyle
\frac{\exp((R_{\mathfrak{s}}\bm q_{a,\mathfrak{s}})^{\mathsf T}
                 (R_{\mathfrak{r}}\bm k_{a,\mathfrak{r}})/\sqrt d)}
{\sum_{\mathfrak{r}'=1}^{\mathfrak{s}}\exp((R_{\mathfrak{s}}\bm q_{a,\mathfrak{s}})^{\mathsf T}
                 (R_{\mathfrak{r}'}\bm k_{a,\mathfrak{r}'})/\sqrt d)},&\mathfrak{r}\le\mathfrak{s},\\[6pt]
0,&\mathfrak{r}>\mathfrak{s}.
\end{cases}
\end{equation}
The $1/\sqrt d$ in these logits and the $1/\sqrt d$ in the residual
output coefficient are separate normalizations.
The latter keeps the output scale per head explicit, as in our
Transformer experiments.

Packing the Q/K/V projections makes attention another
up--nonlinearity--down computation.
For one head, define
$\bm W_{{\rm A},a}^{\ell\uparrow}
=[(\bm W_a^{\ell Q})^{\mathsf T},
  (\bm W_a^{\ell K})^{\mathsf T},
  (\bm W_a^{\ell V})^{\mathsf T}]^{\mathsf T}$
and $\bm W_{{\rm A},a}^{\ell\downarrow}=\bm W_a^{\ell O}$.
For one MLP unit, take the corresponding row of
$\bm W_{\rm M}^{\ell\uparrow}$ and column of
$\bm W_{\rm M}^{\ell\downarrow}$.
The two channel maps are
\begin{equation}
\label{eq:transformer-two-channel-maps}
\begin{aligned}
\phi_{\rm A}:(\bm q,\bm k,\bm v)
&\longmapsto \frac{1}{\sqrt d}\bm vP(\bm q,\bm k)^{\mathsf T},
&\phi_{\rm A}&:\mathbb R^{3d\times S}\to\mathbb R^{d\times S},\\
\phi_{\rm M}:\bm z
&\longmapsto\operatorname{GELU}(\bm z),
&\phi_{\rm M}&:\mathbb R^{1\times S}\to\mathbb R^{1\times S},
\end{aligned}
\end{equation}
where $P$ includes the rotary maps and causal mask above.
Applying $\phi_{\rm M}$ is tokenwise; applying $\phi_{\rm A}$ uses the
whole head's token array.

Either residual update now has the same form:
\begin{equation}
\label{eq:transformer-channel-form}
\begin{aligned}
\bm h_a^\sharp
&=\frac{1}{\sqrt{N_\flat}}\bm W_a^\uparrow
                   \mathcal N(\bm h^\flat),\\
\bm h^{+\flat}
&=\bm h^\flat+\frac{\beta_0\sqrt{N_\flat}}{LN_\sharp}
       \sum_{a=1}^{N_\sharp}\bm W_a^\downarrow\phi(\bm h_a^\sharp).
\end{aligned}
\end{equation}
For the first update use $(\phi,N_\sharp,\beta_0)
=(\phi_{\rm A},N_{\sharp\mathrm A},\beta_{0,\rm A})$; for the second use
$(\phi_{\rm M},N_{\sharp\mathrm M},\beta_{0,\rm M})$ and the residual
state already updated by attention.
The attention output factor $1/\sqrt d$ has moved into $\phi_{\rm A}$;
the mathematical weights and the computed function are unchanged.
Each population occurs $L$ times, although a stage contains two
residual updates.
With $\mathcal N=\mathrm{Id}$, this is Eq.~\ref{eq:mlp-forward},
written as a sum over channels.

Tokens and coordinates within a head specify what a channel does;
they are not the channel population $N_\sharp$.
Pre-normalization is shared residual preprocessing and remains explicit
in Eq.~\ref{eq:transformer-channel-form}.
Neither the MLP width $N_{\sharp\mathrm M}$ nor the concatenated attention
width $dN_{\sharp\mathrm A}$ is required to equal $N_\flat$.
The structural variables are therefore still the residual width,
the two populations of $\sharp$ channels, and their allocation over depth.
Biases and affine normalization parameters can be added at their usual
locations. Their parameter and symmetry conventions must then also be
specified; the displayed example uses non-affine RMSNorm throughout.
\par

\subsection{Adam and Muon}
\label{app:adam-muon}
\paragraph{Adam.}
Adam \citep{kingma2015adam} rescales each gradient coordinate using
its own recent history. It maintains exponential moving averages
$m$ and $v$ of the gradient and its entrywise square, and updates
\begin{equation}
W^+=W-\eta\,\frac{\widehat m}{\sqrt{\widehat v}+\epsilon},
\end{equation}
where the hats denote bias correction for zero-initialized moments,
and the square root and division are entrywise.

\paragraph{Muon.}
Muon \citep{jordan2024muon} transforms the singular values of a
momentum update matrix while retaining its singular vectors.
For $D=U\operatorname{diag}(s_j)V^{\mathsf T}$, the ideal operation
maps every nonzero singular value to one and leaves zero at zero.
In practice, the input matrix is first normalized, then a few
Newton--Schulz iterations apply an odd quintic polynomial
$p(s)=as+bs^3+cs^5$ to its singular values
\citep{jordan2024muon,liu2025muon}.

Following the spectral-clipping viewpoint of \citet{su2025higherordermup},
we model the finite-iteration spectral action by normalized matrix clipping:
\begin{equation}
\label{eq:muon-spectral-clip}
\begin{aligned}
W^+&=W-\eta_M F_\lambda(D_W),\qquad D_W=\nabla_W\mathcal L,\\
F_\lambda(D)&=\operatorname{mclip}(D/\lambda)
=U\operatorname{diag}\bigl(\min(s_j/\lambda,1)\bigr)V^{\mathsf T},
\qquad\lambda>0.
\end{aligned}
\end{equation}
With momentum, the same map acts on the accumulated update matrix.
This model retains a linear response near zero and saturation at
larger singular values, with $F_\lambda(0)=0$.
\par

\subsection{Scalings for feature learning}
\nopagebreak[4]
Feature-learning scaling keeps individual hidden activations random and
$\Theta(1)$ at initialization, while the macroscopic output $\bm f$ may
tend to zero as the network grows.
Learning rates are chosen to retain $\Theta(1)$ changes in intermediate
features during training.
The SGD and Adam choices in Table~\ref{tab:optimizer-scaling-full}
follow these standard principles
\citep{bordelon2024infinite,chizat2025hidden}.
The table gives the scales in both code and standard-Gaussian coordinates.

For Adam, write a parameter group's gradient as $D_t^j=d_jR_t^j$,
where $d_j$ is its entry scale, and let $M_t^j,V_t^j$ be the moments
of the normalized gradients $R_t^j$. Then
\begin{equation}
\begin{aligned}
\widehat m_t^j=d_j\widehat M_t^j,\qquad
\widehat v_t^j=d_j^2\widehat V_t^j,\qquad
\frac{\widehat m_t^j}{\sqrt{\widehat v_t^j}+\epsilon_j}
=\frac{\widehat M_t^j}{\sqrt{\widehat V_t^j}+\epsilon_j/d_j}.
\end{aligned}
\end{equation}
Thus $\epsilon_j=d_j\epsilon_0$ removes $d_j$ from the normalized update.
For example, the code up-projection has $d_\uparrow\asymp1/K$,
so $\epsilon_\uparrow\asymp\epsilon_0/K$.

For Muon, consider the code up-projection for one example and a scalar
output. Suppressing fixed multipliers, its rank-one gradient and sole
nonzero singular value have scales
\begin{equation}
\begin{aligned}
D=\nabla_{\texttt{W\_up}}\mathcal L
\propto\frac1K\bm g_1^{\ell\sharp}(\bm h^{\ell\flat})^{\mathsf T},
\qquad
\sigma(D)\asymp\frac{\sqrt{N_\flat N_\sharp}}{K}.
\end{aligned}
\end{equation}
Choosing $\lambda\asymp\sqrt{N_\flat N_\sharp}/K$ compensates for this
gradient dilution, including both matrix dimensions.
The spectral map preserves the gradient's right singular vector,
which is aligned with $\bm h^{\ell\flat}$, so the direct feature change is
\begin{equation}
\begin{aligned}
\frac{\|(\Delta\texttt{W\_up})\bm h^{\ell\flat}\|}{\sqrt{N_\sharp}}
&\asymp\eta_M\sqrt{\frac{N_\flat}{N_\sharp}}.
\end{aligned}
\end{equation}
Hence $\eta_M\asymp\sqrt{N_\sharp/N_\flat}$ gives a $\Theta(1)$
feature update, as in the table.
\par

\begin{table}[!htbp]
    \centering
    \setlength{\tabcolsep}{4pt}
    \renewcommand{\arraystretch}{1.35}
    \begin{tabular}{@{}llcccc@{}}
        \hline
        Optimizer & Quantity
        & \texttt{W\_in} & \texttt{W\_up}
        & \texttt{W\_down} & \texttt{W\_out}\\
        \hline
        \multicolumn{2}{@{}l}{Initialization std.}
        & $1$ & $1/\sqrt{N_\flat}$ & $\sqrt{N_\flat}/K$ & $1/N_\flat$\\
        \hline
        SGD & LR (code)
        & $N_\flat$ & $K/N_\flat$ & $N_\flat/K$ & $1/N_\flat$\\
        & LR (standard Gaussian)
        & $N_\flat$ & $K$ & $K$ & $N_\flat$\\
        \hline
        Muon & LR (code)
        & $\sqrt{N_\flat}$ & $\sqrt{N_\sharp/N_\flat}$
        & $\sqrt{N_\flat N_\sharp}/K$ & $1/\sqrt{N_\flat}$\\
        & Clip $\lambda$ (code)
        & $1/\sqrt{N_\flat}$ & $\sqrt{N_\flat N_\sharp}/K$
        & $\sqrt{N_\sharp/N_\flat}$ & $\sqrt{N_\flat}$\\
        & LR (standard Gaussian)
        & $\sqrt{N_\flat}$ & $\sqrt{N_\sharp}$
        & $\sqrt{N_\sharp}$ & $\sqrt{N_\flat}$\\
        & Clip $\lambda$ (standard Gaussian)
        & $1/\sqrt{N_\flat}$ & $\sqrt{N_\sharp}/K$
        & $\sqrt{N_\sharp}/K$ & $1/\sqrt{N_\flat}$\\
        \hline
        Adam & LR (code)
        & $1$ & $1/N_\flat$ & $1/K$ & $1/N_\flat$\\
        & $\epsilon$ (code)
        & $1/N_\flat$
        & $1/K$
        & $1/N_\flat$
        & $1$\\
        & LR (standard Gaussian)
        & $1$ & $1/\sqrt{N_\flat}$ & $1/\sqrt{N_\flat}$ & $1$\\
        & $\epsilon$ (standard Gaussian)
        & $1/N_\flat$
        & $1/(K\sqrt{N_\flat})$
        & $1/(K\sqrt{N_\flat})$
        & $1/N_\flat$\\
        \hline
    \end{tabular}
    \caption{Complete size scalings in code and standard-Gaussian coordinates. Fixed task dimensions and hyperparameter multipliers, including $\epsilon_0$, are suppressed.}
    \label{tab:optimizer-scaling-full}
\end{table}
\endgroup

\begingroup
\color{black}
\section{Optimizer-dependent dynamics and population descriptions}
\label{app:sgd-dynamics}

The SGD equations follow by accumulating the parameter updates before
taking a population limit. Throughout this appendix, $K=N_\sharp L$.
All four matrices in \eqref{eq:mlp-forward} are trained, and
$t$ counts training steps while $\ell$ counts residual blocks.
We first give the finite identities and their population response
equations, then construct the corresponding spectral and moment
histories for Muon and Adam. Their implications for residual
Gaussianity follow in the same section.

At step $s$, let $\mathcal B_s$ be a batch of $B_s$ examples and set
\begin{equation}
\label{eq:sgd-loss-feedback}
    \Delta_p(\bm x',y',s)
    =-\frac{\partial\mathcal L_{\rm ex}(\bm f(\bm x',s),y')}
            {\partial f_p},
    \qquad p=1,\ldots,D_{\mathrm o}.
\end{equation}
The loss is averaged over the batch, and all parameter groups are
updated simultaneously. Any output-component averaging belongs to
$\mathcal L_{\rm ex}$.
For fixed positive constants
$\eta_{\rm in},\eta_\uparrow,\eta_\downarrow,\eta_{\rm out}$,
the standard-coordinate SGD learning rates are
\begin{equation}
\label{eq:sgd-four-rates}
\begin{aligned}
    \eta^{\rm W}_{\rm in}&=\eta_{\rm in}\gamma_0^2{N_\flat},
    &\eta^{\rm W}_{\uparrow}&=\eta_\uparrow\gamma_0^2K,\\
    \eta^{\rm W}_{\downarrow}&=\eta_\downarrow\gamma_0^2K,
    &\eta^{\rm W}_{\rm out}&=\eta_{\rm out}\gamma_0^2{N_\flat}.
\end{aligned}
\end{equation}
The four constants may differ; Table~\ref{tab:optimizer-scaling}
suppresses them and the fixed factor $\gamma_0^2$.
\par

\subsection{Backward recursion and parameter gradients}
\label{app:sgd-exact-gradients}

The normalized output Jacobians defined in
\eqref{eq:residual-backward} obey the exact recursion
\begin{equation}
\label{eq:sgd-backward-exact}
\begin{aligned}
    \bm g_q^{L\flat}
    &=(\bm W^L)^{\mathsf T}\bm e_q,\\
    \bm g_q^{\ell\sharp}
    &=\phi'(\bm h^{\ell\sharp})\odot
       \frac{(\bm W^{\ell\downarrow})^{\mathsf T}
                   \bm g_q^{(\ell+1)\flat}}{\sqrt {N_\flat}},\\
    \bm g_q^{\ell\flat}
    &=\bm g_q^{(\ell+1)\flat}
       +\frac{\beta_0\sqrt {N_\flat}}{K}
          (\bm W^{\ell\uparrow})^{\mathsf T}\bm g_q^{\ell\sharp}.
\end{aligned}
\end{equation}
The input and training step are the same throughout each line and have
been suppressed. In particular, there is no factor
$\phi'(\bm h^{\ell\flat})$ in the residual recursion.
The four parameter derivatives are
\begin{equation}
\label{eq:sgd-parameter-jacobians}
\begin{aligned}
 \nabla_{\bm W^0}f_q
 &=\frac{\bm g_q^{0\flat}\bm x^{\mathsf T}}
          {\gamma_0 {N_\flat}\sqrt{D_{\mathrm i}}},&
 \nabla_{\bm W^L}f_q
 &=\frac{\bm e_q(\bm h^{L\flat})^{\mathsf T}}{\gamma_0{N_\flat}},\\
 \nabla_{\bm W^{\ell\uparrow}}f_q
 &=\frac{\beta_0}{\gamma_0K\sqrt {N_\flat}}
      \bm g_q^{\ell\sharp}(\bm h^{\ell\flat})^{\mathsf T},&
 \nabla_{\bm W^{\ell\downarrow}}f_q
 &=\frac{\beta_0}{\gamma_0K\sqrt {N_\flat}}
      \bm g_q^{(\ell+1)\flat}
           \phi(\bm h^{\ell\sharp})^{\mathsf T}.
\end{aligned}
\end{equation}
Multiplication by $\Delta_q$, summation over $q$, and batch averaging
produce the descent direction. Thus the loss feedback and the output
Jacobian remain separate.
\par

\subsection{Exact discrete learning histories}
\label{app:sgd-history}

For a current query $z=(\bm x,t)$ and a past example
$z'=(\bm x',s)$, define the empirical kernels
\begin{equation}
\label{eq:sgd-empirical-kernels}
\begin{aligned}
 \Phi^{\ell\flat}(z,z')
 &=\frac{(\bm h^{\ell\flat}(z))^{\mathsf T}
                    \bm h^{\ell\flat}(z')}{{N_\flat}},&
 \Phi^{\ell\sharp}(z,z')
 &=\frac{\phi(\bm h^{\ell\sharp}(z))^{\mathsf T}
                    \phi(\bm h^{\ell\sharp}(z'))}{{N_\sharp}},\\
 G_{qp}^{\ell\flat}(z,z')
 &=\frac{(\bm g_q^{\ell\flat}(z))^{\mathsf T}
                    \bm g_p^{\ell\flat}(z')}{{N_\flat}},&
 G_{qp}^{\ell\sharp}(z,z')
 &=\frac{(\bm g_q^{\ell\sharp}(z))^{\mathsf T}
                    \bm g_p^{\ell\sharp}(z')}{{N_\sharp}}.
\end{aligned}
\end{equation}
The loss-weighted training history will be written as
\begin{equation}
\label{eq:sgd-history-sum}
 \mathcal S_t[F_p(z,z')]
 =
 \sum_{s=0}^{t-1}\frac1{B_s}
 \sum_{(\bm x',y')\in\mathcal B_s}
 \sum_{p=1}^{D_{\mathrm o}}
 \Delta_p(\bm x',y',s)\,F_p(z,z').
\end{equation}
Accumulating the up-projection updates exactly gives
\begin{equation}
\label{eq:sgd-up-history}
 \bm W^{\ell\uparrow}(t)
 =\bm W^{\ell\uparrow}(0)
   +\frac{\eta_\uparrow\gamma_0\beta_0}{\sqrt {N_\flat}}
     \mathcal S_t[
       \bm g_p^{\ell\sharp}(z')
                  (\bm h^{\ell\flat}(z'))^{\mathsf T}].
\end{equation}
The down-projection has the corresponding outer product of
$\bm g_p^{(\ell+1)\flat}$ and $\phi(\bm h^{\ell\sharp})$.
Substitution into the network equations gives the forward histories
\begin{equation}
\label{eq:sgd-finite-history-forward}
\begin{aligned}
 \bm h^{\ell\sharp}(z)
 &=\bm\chi^{\ell\sharp}(z)
   +\eta_\uparrow\gamma_0\beta_0\,
     \mathcal S_t[
       \bm g_p^{\ell\sharp}(z')\Phi^{\ell\flat}(z,z')],\\
 \bm h^{(\ell+1)\flat}(z)
 &=\bm h^{\ell\flat}(z)+\bm\chi^{(\ell+1)\flat}(z)
   +\frac{\eta_\downarrow\gamma_0\beta_0^2}{L}\,
     \mathcal S_t[
       \bm g_p^{(\ell+1)\flat}(z')\Phi^{\ell\sharp}(z,z')].
\end{aligned}
\end{equation}
Here the parts containing the initial matrices are
\begin{equation}
\label{eq:sgd-initial-projections}
\begin{aligned}
 \bm\chi^{\ell\sharp}(z)
 &=\frac{\bm W^{\ell\uparrow}(0)\bm h^{\ell\flat}(z)}{\sqrt {N_\flat}},\\
 \bm\chi^{(\ell+1)\flat}(z)
 &=\frac{\beta_0\sqrt {N_\flat}}{K}
       \bm W^{\ell\downarrow}(0)\phi(\bm h^{\ell\sharp}(z)),\\
 \bm\xi_q^{\ell\sharp}(z)
 &=\frac{(\bm W^{\ell\downarrow}(0))^{\mathsf T}
                   \bm g_q^{(\ell+1)\flat}(z)}{\sqrt {N_\flat}},\\
 \bm\xi_q^{\ell\flat}(z)
 &=\frac{\beta_0\sqrt {N_\flat}}{K}
       (\bm W^{\ell\uparrow}(0))^{\mathsf T}
                   \bm g_q^{\ell\sharp}(z).
\end{aligned}
\end{equation}
The backward histories are
\begin{equation}
\label{eq:sgd-finite-history-backward}
\begin{aligned}
 \bm g_q^{\ell\sharp}(z)
 &=\phi'(\bm h^{\ell\sharp}(z))\odot
   \left\{\bm\xi_q^{\ell\sharp}(z)
   +\eta_\downarrow\gamma_0\beta_0\,
       \mathcal S_t[
          \phi(\bm h^{\ell\sharp}(z'))
                    G_{qp}^{(\ell+1)\flat}(z,z')]\right\},\\
 \bm g_q^{\ell\flat}(z)
 &=\bm g_q^{(\ell+1)\flat}(z)+\bm\xi_q^{\ell\flat}(z)
   +\frac{\eta_\uparrow\gamma_0\beta_0^2}{L}\,
      \mathcal S_t[
          \bm h^{\ell\flat}(z')G_{qp}^{\ell\sharp}(z,z')].
\end{aligned}
\end{equation}
Hidden inner products supply ${N_\sharp}/K=1/L$ in the residual equations.
The hidden equations themselves retain order-one learning-history
coefficients. This separates the update of an individual hidden
feature from its contribution to a residual increment.

For the endpoints, write
$\kappa_{\rm in}(\bm x,\bm x')
=\bm x^{\mathsf T}\bm x'/D_{\mathrm i}$. Their histories are
\begin{equation}
\label{eq:sgd-endpoint-histories}
\begin{aligned}
 \bm h^{0\flat}(z)
 &=\frac{\bm W^0(0)\bm x}{\sqrt{D_{\mathrm i}}}
   +\eta_{\rm in}\gamma_0\,
       \mathcal S_t[
          \bm g_p^{0\flat}(z')\kappa_{\rm in}(\bm x,\bm x')],\\
 \bm g_q^{L\flat}(z)
 &=(\bm W^L(0))^{\mathsf T}\bm e_q
   +\eta_{\rm out}\gamma_0\,
       \mathcal S_t[\delta_{pq}\bm h^{L\flat}(z')],\\
 f_q(z)
 &=\frac{\bm e_q^{\mathsf T}\bm W^L(0)\bm h^{L\flat}(z)}
             {\gamma_0{N_\flat}}
   +\eta_{\rm out}\,
       \mathcal S_t[\delta_{pq}\Phi^{L\flat}(z,z')].
\end{aligned}
\end{equation}
All fields in \eqref{eq:sgd-finite-history-forward}--%
\eqref{eq:sgd-endpoint-histories} are evaluated in the trained network.
The identities therefore include the full finite update, including
interactions between parameter groups updated at the same step.
\par

\subsection{Dynamical mean-field equations from Gaussian weight integration}
\label{app:sgd-closure-scope}

Weight reuse makes the initial projections in
\eqref{eq:sgd-initial-projections} correlated with the fields they
multiply. Integrating those Gaussian weights and taking the population
saddle separates the projections into Gaussian sources and responses
to weight reuse. The derivation below retains the exact discrete
training histories.
Take a finite input set containing the training examples and probes,
and finitely many training steps; denote their pairs
$z=(\bm x,t)$ by $\mathcal Z$.

The representative fields retain the main-text notation
$(\bm{\mathscr h}^{\ell\flat},\bm{\mathscr h}^{\ell\sharp},
  \bm{\mathscr g}_q^{\ell\flat},\bm{\mathscr g}_q^{\ell\sharp})$.
The second-moment kernels are defined self-consistently by
\begin{equation}
\label{eq:sgd-population-kernels}
\begin{aligned}
 \Phi^{\ell\flat}(z,z')&=\mathbb E[\bm{\mathscr h}^{\ell\flat}(z)\bm{\mathscr h}^{\ell\flat}(z')],&
 \Phi^{\ell\sharp}(z,z')&=\mathbb E[\phi(\bm{\mathscr h}^{\ell\sharp}(z))\phi(\bm{\mathscr h}^{\ell\sharp}(z'))],\\
 G_{qp}^{\ell\flat}(z,z')&=\mathbb E[\bm{\mathscr g}_{q}^{\ell\flat}(z)\bm{\mathscr g}_{p}^{\ell\flat}(z')],&
 G_{qp}^{\ell\sharp}(z,z')&=\mathbb E[\bm{\mathscr g}_{q}^{\ell\sharp}(z)\bm{\mathscr g}_{p}^{\ell\sharp}(z')].
\end{aligned}
\end{equation}
The hidden kernels are second moments and may include nonzero means.

\subsubsection{Gaussian weight integration and the population saddle}
\label{app:sgd-path-integral}

The response terms can be derived by integrating the initial Gaussian
matrices, while retaining the exact discrete training histories.
Fix $L$ and a finite set $\mathcal Z$ containing every training and
probe input at steps $0,\ldots,T$. All sums over $z,z'\in\mathcal Z$
below are ordinary finite sums; batch weights and loss derivatives
occur only in $\mathcal S_t$ from \eqref{eq:sgd-history-sum}.
For one block, use the same $\ell$ throughout.

Thus $\bm W^{\ell\uparrow}(0)$ is ${N_\sharp}\times {N_\flat}$, $\bm W^{\ell\downarrow}(0)$ is ${N_\flat}\times {N_\sharp}$, and their entries are
independent standard Gaussians. The four projections in
\eqref{eq:sgd-initial-projections}, with the depth index suppressed,
are
\begin{equation}
\label{eq:sgd-pi-projections}
\begin{aligned}
\bm\chi^\sharp&=\frac{\bm W^{\ell\uparrow}(0)\bm h^{\ell\flat}}{\sqrt {N_\flat}},&
\bm\xi_q^\flat&=\frac{\beta_0\sqrt {N_\flat}}{K}(\bm W^{\ell\uparrow}(0))^{\mathsf T}\bm g_q^{\ell\sharp},\\
\bm\chi^\flat&=\frac{\beta_0\sqrt {N_\flat}}{K}\bm W^{\ell\downarrow}(0)\phi(\bm h^{\ell\sharp}),&
\bm\xi_q^\sharp&=\frac{(\bm W^{\ell\downarrow}(0))^{\mathsf T}\bm g_q^{(\ell+1)\flat}}{\sqrt {N_\flat}}.
\end{aligned}
\end{equation}
In particular, the residual projections already contain ${\beta_0/L}$.
There is no activation applied to $\bm h^{\ell\flat}$.

\paragraph{The training constraints are retained.}
Let $\Gamma$ collect the forward, backward, projection, and output fields.
The generating functional integrates them against the dynamical measure
\begin{equation}
\label{eq:sgd-pi-dynamical-measure}
 \mathcal D_{\rm dyn}
 =\mathcal D_{\rm bdry}
   \prod_{\ell,z}\left[
   \delta[\bm E_{h,\ell}^\sharp(z)]
   \delta[\bm E_{h,\ell}^\flat(z)]
   \prod_q
   \delta[\bm E_{g,\ell q}^\sharp(z)]
   \delta[\bm E_{g,\ell q}^\flat(z)]\right].
\end{equation}
Here the four $\bm E$ are the left sides minus the right sides of
\eqref{eq:sgd-finite-history-forward} and
\eqref{eq:sgd-finite-history-backward}; $\mathcal D_{\rm bdry}$
enforces \eqref{eq:sgd-endpoint-histories}, including the output used
in the loss feedback. These constraints encode simultaneous SGD,
with every update depending on steps $s<t$.
Ordering the computation by training step, then forward and backward
passes, gives a triangular constraint Jacobian with unit diagonal.
Consequently the generating functional $\mathscr Z[J]$ satisfies
$\mathscr Z[0]=1$, also after arbitrary additive perturbations of
the four projection equations.
The endpoint Gaussian averages can be carried out separately; the
calculation below treats the bulk matrices.

\par

Insert Fourier representations of the four projection constraints.
Their field-only exponent is
\begin{equation}
\label{eq:sgd-pi-fourier}
 iS_{\rm F}
 =i\sum_{\ell,z}\left[
 \widehat{\bm\chi}^{\sharp}\cdot\bm\chi^\sharp
 +\widehat{\bm\chi}^{\flat}\cdot\bm\chi^\flat
 +\sum_q\left(
 \widehat{\bm\xi}_q^\sharp\cdot\bm\xi_q^\sharp
 +\widehat{\bm\xi}_q^\flat\cdot\bm\xi_q^\flat\right)\right].
\end{equation}
The remaining initial-weight dependence is
$\exp[-i\sum_{ai}\Lambda^\uparrow_{ai}W_{ai}^{\ell\uparrow}(0)
      -i\sum_{ia}\Lambda^\downarrow_{ia}W_{ia}^{\ell\downarrow}(0)]$, where, for each layer,
\begin{equation}
\label{eq:sgd-pi-weight-sources}
\begin{aligned}
 \Lambda^\uparrow_{ai}
 &=\sum_z\left[
     \frac{\widehat\chi^\sharp_a(z)h_i^{\ell\flat}(z)}{\sqrt {N_\flat}}
     +\frac{\beta_0\sqrt {N_\flat}}{K}
        \sum_q\widehat\xi^\flat_{qi}(z)g_{q,a}^{\ell\sharp}(z)\right],\\
 \Lambda^\downarrow_{ia}
 &=\sum_z\left[
     \frac{\beta_0\sqrt {N_\flat}}{K}\widehat\chi^\flat_i(z)\phi(h_a^{\ell\sharp}(z))
     +\frac1{\sqrt {N_\flat}}
        \sum_q\widehat\xi^\sharp_{qa}(z)g_{q,i}^{(\ell+1)\flat}(z)\right].
\end{aligned}
\end{equation}
Thus the full generating functional before the bulk Gaussian average is
\begin{equation}
\label{eq:sgd-pi-generating-functional}
 \mathscr Z[J]
 =\mathbb E_{\rm bdry}
   \int D\Gamma\,D\widehat\chi\,D\widehat\xi\,
   \mathcal D_{\rm dyn}\,
   e^{J\cdot\Gamma+iS_{\rm F}}
   \prod_\ell
   \mathbb E_{\bm W^{\ell\uparrow}(0),\bm W^{\ell\downarrow}(0)}
   e^{-i\Lambda_\ell^\uparrow:\bm W^{\ell\uparrow}(0)-i\Lambda_\ell^\downarrow:\bm W^{\ell\downarrow}(0)}.
\end{equation}
The outer expectation averages the initial input and readout matrices.
\par
At this point all fields are integration variables, so the Gaussian
average is exact despite their dependence on the trained weights:
\begin{equation}
\label{eq:sgd-pi-weight-average}
 \mathbb E_{\bm W^{\ell\uparrow}(0),\bm W^{\ell\downarrow}(0)}
 \!e^{-i\Lambda^\uparrow:\bm W^{\ell\uparrow}(0)-i\Lambda^\downarrow:\bm W^{\ell\downarrow}(0)}
 =\exp\!\left[-\frac12\sum_{ai}(\Lambda^\uparrow_{ai})^2
              -\frac12\sum_{ia}(\Lambda^\downarrow_{ia})^2\right].
\end{equation}

\paragraph{Macroscopic kernels separate the coordinates.}
Insert the four empirical second moments in
\eqref{eq:sgd-empirical-kernels} and two mixed kernels
\begin{equation}
\label{eq:sgd-pi-mixed-kernels}
\begin{aligned}
 A_q^\flat(z,z')
   &=-\frac{i}{{N_\flat}}\bm h^{\ell\flat}(z)\cdot
                         \widehat{\bm\xi}_q^\flat(z'),\\
 A_q^\sharp(z,z')
   &=-\frac{i}{{N_\sharp}}\phi(\bm h^{\ell\sharp}(z))\cdot
                         \widehat{\bm\xi}_q^\sharp(z').
\end{aligned}
\end{equation}
Their response interpretation will follow from the Fourier integral.
For a generic second moment
$C(z,z')=\bm v(z)\cdot\bm v(z')/d$, its constraint contributes
\begin{equation}
\label{eq:sgd-pi-kernel-constraint}
 1\ \propto\ \int DC\,D\widehat C\,
 \exp\!\left\{\sum_{z,z'}\widehat C(z,z')
       [dC(z,z')-\bm v(z)\cdot\bm v(z')]\right\}.
\end{equation}
Apply this identity to $(\Phi^\flat,\Phi^\sharp,
G^{(\ell+1)\flat},G^\sharp)$ with respective widths $({N_\flat},{N_\sharp},{N_\flat},{N_\sharp})$;
for $G$, the pair $(q,z)$ is a single covariance index.
Conjugate integration contours and field-independent normalizing
constants are suppressed. The two mixed-kernel constraints contribute
\begin{equation}
\label{eq:sgd-pi-response-constraint}
\begin{aligned}
 S_A=-\sum_{q,z,z'}\big[
 &{N_\flat}\widehat A_q^\flat A_q^\flat
  +i\widehat A_q^\flat\,
            \bm h^{\ell\flat}(z)\cdot\widehat{\bm\xi}_q^\flat(z')\\
 &+{N_\sharp}\widehat A_q^\sharp A_q^\sharp
  +i\widehat A_q^\sharp\,
            \phi(\bm h^{\ell\sharp}(z))\cdot\widehat{\bm\xi}_q^\sharp(z')\big].
\end{aligned}
\end{equation}
Undisplayed arguments on the conjugate and mixed kernels are $(z,z')$.
In the dynamical measure, empirical kernels are simultaneously replaced
by these constrained macroscopic variables.

The readout is another macroscopic variable to be closed. Since
$\bm g_q^{L\flat}(z)=(\bm W^L(t))^{\mathsf T}\bm e_q$, it is exactly
$f_q(z)=(\gamma_0{N_\flat})^{-1}
\bm h^{L\flat}(z)\cdot\bm g_q^{L\flat}(z)$.
Represent its constraint in $\mathcal D_{\rm bdry}$ using a conjugate
$\widehat f_q(z)$ with exponent
\begin{equation}
\label{eq:sgd-pi-readout-constraint}
 S_f=\sum_{q,z}\widehat f_q(z)
 \left[\gamma_0{N_\flat} f_q(z)
       -\sum_{i=1}^{N_\flat} h_i^{L\flat}(z)g_{q,i}^{L\flat}(z)\right].
\end{equation}
This replaces the readout delta; the input and terminal-Jacobian
constraints remain in $\mathcal D_{\rm bdry}$.
At fixed $f$, the loss feedback $\Delta_p$ is shared by all coordinates,
and the second term of \eqref{eq:sgd-pi-readout-constraint} is a sum
of single-coordinate terms. Averaging the independent initial endpoint
matrices also preserves this factorization.
\par

Expanding the two squares in \eqref{eq:sgd-pi-weight-average} now gives
the following averaged action for one block:
\begin{equation}
\label{eq:sgd-pi-averaged-action}
\begin{aligned}
 S_{\rm av}=\sum_{z,z'}\Big[
 &-\frac12\Phi^\flat(z,z')\,
       \widehat{\bm\chi}^\sharp(z)\cdot
       \widehat{\bm\chi}^\sharp(z')\\
 &-\frac{\beta_0^2{N_\flat}}{2KL}\sum_{q,p}G^\sharp_{qp}(z,z')\,
       \widehat{\bm\xi}_q^\flat(z)\cdot
       \widehat{\bm\xi}_p^\flat(z')\\
 &-i\frac{\beta_0{N_\flat}}{K}\sum_q A_q^\flat(z,z')\,
       \widehat{\bm\chi}^\sharp(z)\cdot\bm g_q^{\ell\sharp}(z')\\
 &-\frac{\beta_0^2{N_\flat}}{2KL}\Phi^\sharp(z,z')\,
       \widehat{\bm\chi}^\flat(z)\cdot
       \widehat{\bm\chi}^\flat(z')\\
 &-\frac12\sum_{q,p}G_{qp}^{(\ell+1)\flat}(z,z')\,
       \widehat{\bm\xi}_q^\sharp(z)\cdot
       \widehat{\bm\xi}_p^\sharp(z')\\
 &-i\frac{\beta_0}{L}\sum_q A_q^\sharp(z,z')\,
       \widehat{\bm\chi}^\flat(z)\cdot\bm g_q^{(\ell+1)\flat}(z')\Big].
\end{aligned}
\end{equation}
For example, the up-matrix cross term is
$-(\beta_0/K)\sum_{q,z,z'}
(\widehat{\bm\chi}^\sharp(z)\cdot\bm g_q^{\ell\sharp}(z'))
(\bm h^{\ell\flat}(z)\cdot\widehat{\bm\xi}_q^\flat(z'))$.
Substitution of \eqref{eq:sgd-pi-mixed-kernels} gives the factor
$-i\beta_0{N_\flat}/K$ in the third line. The down-matrix cross term similarly
gives $-i\frac{\beta_0}{L}$ in the last line.
The action, together with the constrained histories, factors over
${N_\flat}$ residual coordinates and ${N_\sharp}$ hidden coordinates at each block.
Each residual coordinate retains its entire depth history; it is
not factored separately at different depths.

\paragraph{The physical saddle is normalized and causal.}
Varying the conjugate kernels gives the population second moments
\eqref{eq:sgd-population-kernels} and the single-coordinate versions
of \eqref{eq:sgd-pi-mixed-kernels}.
Varying the mixed kernels gives
\begin{equation}
\label{eq:sgd-pi-conjugate-responses}
\begin{aligned}
 \widehat A_q^\flat(z,z')
   &=-i\frac{\beta_0}{L}\,\mathbb E[
            \widehat\chi^\sharp(z)\bm{\mathscr g}_q^{\ell\sharp}(z')],\\
 \widehat A_q^\sharp(z,z')
   &=-i\frac{\beta_0{N_\flat}}{K}\,\mathbb E[
            \widehat\chi^\flat(z)\bm{\mathscr g}_q^{(\ell+1)\flat}(z')].
\end{aligned}
\end{equation}
Expectations here refer to one representative coordinate of the
corresponding population. The remaining conjugate kernels vanish
at zero physical source:
\begin{equation}
\label{eq:sgd-pi-zero-conjugates}
 \widehat\Phi^\flat=\widehat\Phi^\sharp
 =\widehat G^\flat=\widehat G^\sharp=0.
\end{equation}
Two ingredients establish this statement. First, differentiating
$\mathscr Z[0]=1$ twice with respect to additive projection
perturbations makes every pure Fourier-field second moment vanish.
Without the history terms, the four kernel saddle equations would
read
\begin{equation}
\label{eq:sgd-pi-second-moment-saddle}
\begin{aligned}
 \widehat\Phi^\flat(z,z')
   &=\frac{1}{2{N_\flat}}\,
       \mathbb E[\widehat{\bm\chi}^\sharp(z)\cdot
                 \widehat{\bm\chi}^\sharp(z')],\\
 \widehat\Phi^\sharp(z,z')
   &=\frac{\beta_0^2{N_\flat}}{2KL{N_\sharp}}\,
       \mathbb E[\widehat{\bm\chi}^\flat(z)\cdot
                 \widehat{\bm\chi}^\flat(z')],\\
 \widehat G_{qp}^{(\ell+1)\flat}(z,z')
   &=\frac{1}{2{N_\flat}}\,
       \mathbb E[\widehat{\bm\xi}_q^\sharp(z)\cdot
                 \widehat{\bm\xi}_p^\sharp(z')],\\
 \widehat G^\sharp_{qp}(z,z')
   &=\frac{\beta_0^2{N_\flat}}{2KL{N_\sharp}}\,
       \mathbb E[\widehat{\bm\xi}_q^\flat(z)\cdot
                 \widehat{\bm\xi}_p^\flat(z')].
\end{aligned}
\end{equation}
The vector expectations in this display include the coordinate sums.

Second, the explicit histories in $\mathcal D_{\rm dyn}$ also depend
on $\Phi$ and $G$. Their variations vanish at the causal saddle,
rather than providing extra terms in
\eqref{eq:sgd-pi-second-moment-saddle}.
To see this, temporarily exponentiate the dynamical constraints
with Fourier fields $\widehat E_h^\sharp,\widehat E_h^\flat,
\widehat E_{g,q}^\sharp,\widehat E_{g,q}^\flat$.
Ignoring their displayed learning-rate coefficients and the common
past loss feedback, the four variations give
\begin{equation}
\label{eq:sgd-pi-causal-history-terms}
\begin{aligned}
 \delta\Phi^\flat:\quad&
 \mathbb E[-i\widehat E_h^\sharp(z)\bm{\mathscr g}_p^{\ell\sharp}(z')]=0,\\
 \delta\Phi^\sharp:\quad&
 \mathbb E[-i\widehat E_h^\flat(z)\bm{\mathscr g}_p^{(\ell+1)\flat}(z')]=0,\\
 \delta G_{qp}^{(\ell+1)\flat}:\quad&
 \mathbb E[-i\widehat E_{g,q}^\sharp(z)
              \phi'(\bm{\mathscr h}^{\ell\sharp}(z))\phi(\bm{\mathscr h}^{\ell\sharp}(z'))]=0,\\
 \delta G^\sharp_{qp}:\quad&
 \mathbb E[-i\widehat E_{g,q}^\flat(z)\bm{\mathscr h}^{\ell\flat}(z')]=0,
 \qquad z=(\bm x,t),\quad z'=(\bm x',s),\quad s<t.
\end{aligned}
\end{equation}
Each expectation is the response of its accompanying fields to a
current additive perturbation of the indicated equation.
All primed fields precede that perturbation. In the third line,
$\bm{\mathscr h}^{\ell\sharp}(z)$ also precedes the current backward pass, so its activation
derivative has zero response to that perturbation.
This argument applies to every output pair $(q,p)$ and preserves
the strict discrete-step ordering. It justifies
\eqref{eq:sgd-pi-zero-conjugates} for the full training functional,
not only for its Gaussian-matrix factor.

The readout closes by the same causal saddle. Variation with respect to
$\widehat f$ gives
\begin{equation}
\label{eq:sgd-pi-readout-saddle}
 f_q(z)=\gamma_0^{-1}\mathbb E[\bm{\mathscr h}^{L\flat}(z)\bm{\mathscr g}_{q}^{L\flat}(z)].
\end{equation}
If $S_{\rm dyn}$ denotes the Fourier exponent of the dynamical
constraints other than the readout, variation with respect to $f$ gives
\begin{equation}
\label{eq:sgd-pi-output-conjugate}
 0=\gamma_0{N_\flat}\widehat f_q(z')
   +\mathbb E\!\left[
        \frac{\partial S_{\rm dyn}}{\partial f_q(z')}\right],
 \qquad \widehat f_q(z')=0.
\end{equation}
Indeed, $f_q(\bm x',s)$ enters these histories only through loss feedback
used at later steps $t>s$. Its derivative multiplies one of the four
zero responses in \eqref{eq:sgd-pi-causal-history-terms} or an endpoint
counterpart:
\begin{equation}
\label{eq:sgd-pi-endpoint-causality}
\begin{aligned}
 \mathbb E[-i\widehat E_{h,0}(z)\bm{\mathscr g}_{p}^{0\flat}(z')]&=0,\\
 \mathbb E[-i\widehat E_{g,L,q}(z)\bm{\mathscr h}^{L\flat}(z')]&=0,
 \qquad z=(\bm x,t),\quad z'=(\bm x',s),\quad s<t.
\end{aligned}
\end{equation}
The additional loss derivatives are also past quantities.
Thus no readout-conjugate term survives in the representative equations,
while \eqref{eq:sgd-pi-readout-saddle} determines their feedback.

\par

\paragraph{Gaussian sources and Fourier responses close the equations.}
At this saddle the quadratic Fourier terms are decoupled by
\begin{equation}
\label{eq:sgd-pi-hs-identity}
 \exp\!\left(-\frac12\widehat v^{\mathsf T}C\widehat v\right)
 =\mathbb E_{v\sim\mathcal N(0,C)}
       \exp(-i\widehat v^{\mathsf T}v).
\end{equation}
This characteristic-function identity also applies to singular
positive-semidefinite kernels. The four Gaussian covariances are
\begin{equation}
\label{eq:sgd-pi-hs-kernels}
 C_{u^\sharp}=\Phi^\flat,\qquad
 C_{r^\sharp}=G^{(\ell+1)\flat},\qquad
 C_{u^\flat}=\frac{\beta_0^2{N_\flat}}{KL}\Phi^\sharp,\qquad
 C_{r^\flat}=\frac{\beta_0^2{N_\flat}}{KL}G^\sharp.
\end{equation}
For $r$, covariance indices include $q,p$.
There are no mixed quadratic terms between these four Fourier
families. Their Gaussian sources can therefore be taken independent
across families and blocks, with their displayed within-family
input, training-step, and output correlations.

Integrating the remaining linear Fourier fields gives
\begin{equation}
\label{eq:sgd-pi-source-equations}
\begin{aligned}
 \chi^\sharp(z)
 &=u^\sharp(z)+\frac{\beta_0{N_\flat}}{K}
             \sum_{q,z'}A_q^\flat(z,z')\bm{\mathscr g}_q^{\ell\sharp}(z'),\\
 \chi^\flat(z)
 &=u^\flat(z)+{\beta_0/L}\sum_{q,z'}A_q^\sharp(z,z')\bm{\mathscr g}_q^{(\ell+1)\flat}(z'),\\
 \xi_q^\flat(z)
 &=r_q^\flat(z)+\sum_{z'}
                    \widehat A_q^\flat(z',z)\bm{\mathscr h}^{\ell\flat}(z'),\\
 \xi_q^\sharp(z)
 &=r_q^\sharp(z)+\sum_{z'}
                    \widehat A_q^\sharp(z',z)\phi(\bm{\mathscr h}^{\ell\sharp}(z')).
\end{aligned}
\end{equation}
The transposed arguments in the last two lines follow from
\eqref{eq:sgd-pi-response-constraint}.
Because a source perturbation enters the Fourier exponent with
coefficient $-i\widehat v$, its averaged response is
\begin{equation}
\label{eq:sgd-pi-fourier-response}
 \mathbb E\frac{\partial \mathcal O}{\partial v(z')}
   =-i\,\mathbb E[\mathcal O\,\widehat v(z')].
\end{equation}
The normalization derivative is zero. This differentiates the full
single-coordinate computation while holding the macroscopic kernels
and feedback fixed; it does not treat $\mathcal O$ as independent of
the source. Consequently
\begin{equation}
\label{eq:sgd-pi-response-identification}
\begin{aligned}
 A_q^\flat(z,z')
   &=\mathbb E\frac{\partial \bm{\mathscr h}^{\ell\flat}(z)}
                    {\partial r_q^\flat(z')},&
 A_q^\sharp(z,z')
   &=\mathbb E\frac{\partial\phi(\bm{\mathscr h}^{\ell\sharp}(z))}
                    {\partial r_q^\sharp(z')},\\
 \widehat A_q^\flat(z,z')
   &={\beta_0/L}\,B_q^\sharp(z',z),&
 \widehat A_q^\sharp(z,z')
   &=\frac{\beta_0{N_\flat}}{K}B_q^\flat(z',z).
\end{aligned}
\end{equation}
Here $B_q^\sharp(z,z')=\mathbb E[
\partial \bm{\mathscr g}_q^{\ell\sharp}(z)/\partial u^\sharp(z')]$ and
$B_q^\flat(z,z')=\mathbb E[
\partial \bm{\mathscr g}_q^{(\ell+1)\flat}(z)/\partial u^\flat(z')]$, as in
\eqref{eq:sgd-population-responses}. In particular, the signs of all
four reaction terms are positive.

The population-counting coefficients can equivalently be written as
\begin{equation}
\label{eq:sgd-pi-current-coefficients}
 \frac{\beta_0}{L}\frac{N_\flat}{N_\sharp}
   =\frac{\beta_0N_\flat}{K},\qquad
 \left(\frac{\beta_0}{L}\right)^2\frac{N_\flat}{N_\sharp}
   =\frac{\beta_0^2N_\flat}{KL}.
\end{equation}
Substituting \eqref{eq:sgd-pi-source-equations} and
\eqref{eq:sgd-pi-response-identification} into the exact histories
gives precisely \eqref{eq:sgd-dmft-hidden} and
\eqref{eq:sgd-dmft-residual}, with source covariances
\eqref{eq:sgd-population-sources}.
The Gaussian integration itself is finite-dimensional and exact;
the population approximation is the saddle-point replacement of the
empirical kernels and responses by their self-consistent values.
Training time has remained discrete throughout.
\par

\subsubsection{Self-consistent fields and boundary conditions}

For each layer introduce centered Gaussian source families
$u^{\ell\sharp},r_{q}^{\ell\sharp},u^{(\ell+1)\flat},r_{q}^{\ell\flat}$.
The residual sources already include their branch coefficients.
At the ratio $N_\flat/K$ in this fixed-depth width limit, their
nonzero covariances are
\begin{equation}
\label{eq:sgd-population-sources}
\begin{aligned}
 \mathbb E[u^{\ell\sharp}(z)u^{\ell\sharp}(z')]
   &=\Phi^{\ell\flat}(z,z'),\\
 \mathbb E[r_{q}^{\ell\sharp}(z)r_{p}^{\ell\sharp}(z')]
   &=G_{qp}^{(\ell+1)\flat}(z,z'),\\
 \mathbb E[u^{(\ell+1)\flat}(z)u^{(\ell+1)\flat}(z')]
   &=\frac{\beta_0^2N_\flat}{KL}\Phi^{\ell\sharp}(z,z'),\\
 \mathbb E[r_{q}^{\ell\flat}(z)r_{p}^{\ell\flat}(z')]
   &=\frac{\beta_0^2N_\flat}{KL}G_{qp}^{\ell\sharp}(z,z').
\end{aligned}
\end{equation}
Source families and layers are independent in this Gaussian-source
representation. Their input, training-step, and output-component
correlations are the kernels displayed above; the effects of using
each matrix and its transpose enter through the responses below.

Define the averaged responses
\begin{equation}
\label{eq:sgd-population-responses}
\begin{aligned}
 A_{p}^{\ell\flat}(z,z')
   &=\mathbb E\frac{\partial \bm{\mathscr h}^{\ell\flat}(z)}
                    {\partial r_{p}^{\ell\flat}(z')},&
 A_{p}^{\ell\sharp}(z,z')
   &=\mathbb E\frac{\partial\phi(\bm{\mathscr h}^{\ell\sharp}(z))}
                    {\partial r_{p}^{\ell\sharp}(z')},\\
 B_{q}^{\ell\flat}(z,z')
   &=\mathbb E\frac{\partial \bm{\mathscr g}_{q}^{(\ell+1)\flat}(z)}
                    {\partial u^{(\ell+1)\flat}(z')},&
 B_{q}^{\ell\sharp}(z,z')
   &=\mathbb E\frac{\partial \bm{\mathscr g}_{q}^{\ell\sharp}(z)}
                    {\partial u^{\ell\sharp}(z')}.
\end{aligned}
\end{equation}
These derivatives perturb a source argument while holding the
population kernels and macroscopic feedback fixed.
They follow the full perturbed computation, including the activation
derivative within $\bm{\mathscr g}_{q}^{\ell\sharp}$. Nonsmooth activations require the
corresponding derivatives in expectation.

The Gaussian integral and response identities above give hidden
reaction coefficients $\beta_0\frac{N_\flat}{K}$ and residual reaction coefficients
$\beta_0/L$. Substituting them into the exact histories yields
\begin{equation}
\label{eq:sgd-dmft-hidden}
\begin{aligned}
 \bm{\mathscr h}^{\ell\sharp}(z)
 &=u^{\ell\sharp}(z)
   +\beta_0\frac{N_\flat}{K}\sum_{p,z'}A_{p}^{\ell\flat}(z,z')\bm{\mathscr g}_{p}^{\ell\sharp}(z')\\
 &\quad+\eta_\uparrow\gamma_0\beta_0
       \mathcal S_t[\bm{\mathscr g}_{p}^{\ell\sharp}(z')\Phi^{\ell\flat}(z,z')],\\
 \bm{\mathscr g}_{q}^{\ell\sharp}(z)
 &=\phi'(\bm{\mathscr h}^{\ell\sharp}(z))
   \left\{\begin{aligned}
   &r_{q}^{\ell\sharp}(z)
   +\beta_0\frac{N_\flat}{K}\sum_{z'}B_{q}^{\ell\flat}(z,z')\phi(\bm{\mathscr h}^{\ell\sharp}(z'))\\
   &+\eta_\downarrow\gamma_0\beta_0
       \mathcal S_t[\phi(\bm{\mathscr h}^{\ell\sharp}(z'))G_{qp}^{(\ell+1)\flat}(z,z')]
   \end{aligned}\right\},
\end{aligned}
\end{equation}
\begin{equation}
\label{eq:sgd-dmft-residual}
\begin{aligned}
 \bm{\mathscr h}^{(\ell+1)\flat}(z)-\bm{\mathscr h}^{\ell\flat}(z)
 &=u^{(\ell+1)\flat}(z)
   +\frac{\beta_0}{L}\sum_{p,z'}
          A_{p}^{\ell\sharp}(z,z')\bm{\mathscr g}_{p}^{(\ell+1)\flat}(z')\\
 &\quad+\frac{\eta_\downarrow\gamma_0\beta_0^2}{L}
       \mathcal S_t[\bm{\mathscr g}_{p}^{(\ell+1)\flat}(z')\Phi^{\ell\sharp}(z,z')],\\
 \bm{\mathscr g}_{q}^{\ell\flat}(z)-\bm{\mathscr g}_{q}^{(\ell+1)\flat}(z)
 &=r_{q}^{\ell\flat}(z)
   +\frac{\beta_0}{L}\sum_{z'}
          B_{q}^{\ell\sharp}(z,z')\bm{\mathscr h}^{\ell\flat}(z')\\
 &\quad+\frac{\eta_\uparrow\gamma_0\beta_0^2}{L}
       \mathcal S_t[\bm{\mathscr h}^{\ell\flat}(z')G_{qp}^{\ell\sharp}(z,z')].
\end{aligned}
\end{equation}
Here $\mathcal S_t$ retains the batches and earlier training steps of
\eqref{eq:sgd-history-sum}, with loss derivatives evaluated at the
limiting output. Response sums run over $\mathcal Z$; the dependencies
of the training computation determine which responses vanish.

For the boundaries, let $u_{\rm in}(\bm x)$ have covariance
$\kappa_{\rm in}(\bm x,\bm x')$ and let
$r_{{\rm out},q}$ be independent standard Gaussians.
They are independent of the bulk sources and are reused at all
training steps. The boundary equations and readout close the
feedback loop:
\begin{equation}
\label{eq:sgd-dmft-boundaries}
\begin{aligned}
 \bm{\mathscr h}^{0\flat}(z)
 &=u_{\rm in}(\bm x)
   +\eta_{\rm in}\gamma_0
        \mathcal S_t[\bm{\mathscr g}_{p}^{0\flat}(z')\kappa_{\rm in}(\bm x,\bm x')],\\
 \bm{\mathscr g}_{q}^{L\flat}(z)
 &=r_{{\rm out},q}
   +\eta_{\rm out}\gamma_0
        \mathcal S_t[\delta_{pq}\bm{\mathscr h}^{L\flat}(z')],\\
 f_q(z)&=\gamma_0^{-1}\mathbb E[\bm{\mathscr h}^{L\flat}(z)\bm{\mathscr g}_{q}^{L\flat}(z)].
\end{aligned}
\end{equation}
Together these equations determine kernels, responses, covariances,
and loss feedback from one population law. The final identity uses
the linear readout at depth $L$.

\newpage
{\color{black}
\subsubsection{Finite and limiting objects}
\label{app:limit-diagram}}
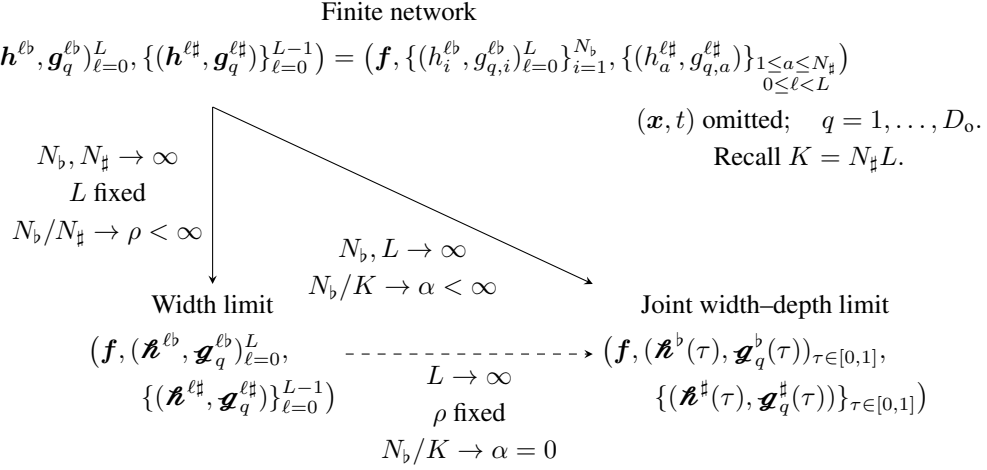
\begin{figure}[!ht]
\color{black}
\centering
\normalsize
    \begin{tikzpicture}[x=1pt,y=1pt,>=stealth]
        \node[anchor=north west,inner sep=0pt] (finite) at (0,0) {$
            \begin{gathered}
                \text{Finite network}\\
                \bigl(\bm f,
                  (\bm h^{\ell\flat},\bm g_q^{\ell\flat})_{\ell=0}^{L},
                  \{(\bm h^{\ell\sharp},\bm g_q^{\ell\sharp})\}_{\ell=0}^{L-1}
                \bigr)
                =
                \bigl(\bm f,
                  \{(h_i^{\ell\flat},g_{q,i}^{\ell\flat})_{\ell=0}^{L}\}_{i=1}^{N_\flat},
                  \{(h_a^{\ell\sharp},g_{q,a}^{\ell\sharp})\}_{
                    \substack{1\leq a\leq N_\sharp\\0\leq\ell<L}}
                \bigr)
            \end{gathered}
        $};
        \coordinate (split) at ([xshift=100pt,yshift=-5pt]finite.south west);
        \node[inner sep=3pt] (widthlimit) at ([yshift=-93pt]split) {$
            \begin{gathered}
                \text{Width limit}\\
                \begin{aligned}
                    &\bigl(\bm f,
                      (\bm{\mathscr{h}}^{\ell\flat},
                       \bm{\mathscr{g}}_q^{\ell\flat})_{\ell=0}^{L},\\
                    &\qquad
                      \{(\bm{\mathscr{h}}^{\ell\sharp},
                         \bm{\mathscr{g}}_q^{\ell\sharp})\}_{\ell=0}^{L-1}
                      \bigr)
                \end{aligned}
            \end{gathered}
        $};
        \node[inner sep=3pt] (jointlimit) at ([xshift=208pt]widthlimit.center) {$
            \begin{gathered}
                \text{Joint width--depth limit}\\
                \begin{aligned}
                    &\bigl(\bm f,
                      (\bm{\mathscr{h}}^\flat(\tau),
                       \bm{\mathscr{g}}_q^\flat(\tau))_{\tau\in[0,1]},\\
                    &\qquad
                      \{(\bm{\mathscr{h}}^\sharp(\tau),
                         \bm{\mathscr{g}}_q^\sharp(\tau))\}_{\tau\in[0,1]}
                      \bigr)
                \end{aligned}
            \end{gathered}
        $};
        \draw[->] (split) -- (widthlimit.north)
            node[midway,left] {$
                \begin{gathered}
                    N_\flat,N_\sharp\to\infty\\
                    L\ \text{fixed}\\
                    N_\flat/N_\sharp\to\rho<\infty
                \end{gathered}
            $};
        \draw[->] (split) -- (jointlimit.north west)
            node[midway,below,yshift=-12pt] {$
                \begin{gathered}
                    N_\flat,L\to\infty\\
                    N_\flat/K\to\alpha<\infty
                \end{gathered}
            $};
        \draw[->,dashed] (widthlimit.east) -- (jointlimit.west)
            node[midway,below] {$
                \begin{gathered}
                    L\to\infty\\
                    \rho\ \text{fixed}\\
                    N_\flat/K\to\alpha=0
                \end{gathered}
            $};
        \node[anchor=north east,inner sep=0pt]
            at ([xshift=390pt,yshift=-5pt]finite.south west) {$
                \begin{gathered}
                    (\bm x,t)\text{ omitted};\quad q=1,\ldots,D_{\textrm{o}}.\\
                    \text{Recall }K=N_\sharp L.
                \end{gathered}
            $};
    \end{tikzpicture}
    \caption{\textcolor{black}{Finite random histories and two scaling paths.
The dashed path keeps the residual-to-hidden ratio fixed before increasing depth,
and therefore has $\alpha=0$. The diagonal permits a positive limiting
$N_\flat/K$. These are different scaling families, not an assertion that
two identical limiting operations can be interchanged.}}
    \label{fig:network-limits}
\end{figure}

\subsection{Complete continuous-depth SGD system}
\label{app:sgd-dmft-depth}

The continuous-depth system couples residual trajectories to local
hidden variables and trainable input and readout maps.
Take $L\to\infty$ with $N_\flat/K\to\alpha$ and write
$\tau\in[0,1]$, $z=(\bm x,t)$, and $z'=(\bm x',s)$.
Training steps remain discrete, with the four learning-rate constants
in \eqref{eq:sgd-four-rates}.
Below, $\mathbb E_{\mathcal B_s}$ denotes the average over the examples
$(\bm x',y')$ in batch $\mathcal B_s$, and
$\Delta_p(\bm x',y',s)=-\partial_{f_p}\mathcal L_{\rm ex}(\bm f(\bm x',s),y')$.
Response sums run over the input--step pairs $z'\in\mathcal Z$;
learning sums run only over $s<t$.

The kernels are determined by the same population law:
\begin{subequations}
\label{eq:sgd-depth-system}
\begin{equation}
\label{eq:sgd-depth-kernels}
\begin{aligned}
 \Phi^\flat(\tau;z,z')
 &=\mathbb E[\bm{\mathscr h}^\flat(\tau,z)\bm{\mathscr h}^\flat(\tau,z')],\\
 \Phi^\sharp(\tau;z,z')
 &=\mathbb E[\phi(\bm{\mathscr h}^\sharp(\tau,z))
             \phi(\bm{\mathscr h}^\sharp(\tau,z'))],\\
 G_{qp}^\flat(\tau;z,z')
 &=\mathbb E[\bm{\mathscr g}_q^\flat(\tau,z)\bm{\mathscr g}_p^\flat(\tau,z')],\\
 G_{qp}^\sharp(\tau;z,z')
 &=\mathbb E[\bm{\mathscr g}_q^\sharp(\tau,z)\bm{\mathscr g}_p^\sharp(\tau,z')].
\end{aligned}
\end{equation}
The centered Gaussian sources have covariances
\begin{equation}
\label{eq:sgd-dmft-noise-depth}
\begin{aligned}
 \mathbb E[du(\tau,z)du(\tau,z')]
    &=\beta_0^2\alpha\,\Phi^\sharp(\tau;z,z')\,d\tau,\\
 \mathbb E[dr_q(\tau,z)dr_p(\tau,z')]
    &=\beta_0^2\alpha\,G_{qp}^\sharp(\tau;z,z')\,d\tau,\\
 \mathbb E[u^\sharp(\tau,z)u^\sharp(\tau,z')]
    &=\Phi^\flat(\tau;z,z'),\\
 \mathbb E[r_q^\sharp(\tau,z)r_p^\sharp(\tau,z')]
    &=G_{qp}^\flat(\tau;z,z').
\end{aligned}
\end{equation}
Different source families and depth slices are independent; correlations
across inputs, training steps, and output components are retained above.
In the four equations below, all bulk fields and kernels on the
right-hand side are evaluated at the displayed depth $\tau$.

The residual variables satisfy the depth-increment equations
\begin{equation}
\label{eq:sgd-dmft-continuous}
\begin{aligned}
 d\bm{\mathscr h}^\flat(\tau,z)
 &=du(\tau,z)
   +\beta_0\sum_{p,z'}
      \mathbb E\!\left[
        \frac{\delta\phi(\bm{\mathscr h}^\sharp(z))}
             {\delta r_p^\sharp(z')}\right]
      \bm{\mathscr g}_p^\flat(z')\,d\tau\\
 &\quad+\eta_\downarrow\gamma_0\beta_0^2
      \sum_{s=0}^{t-1}\mathbb E_{\mathcal B_s}\!\left[
        \sum_p\Delta_p(\bm x',y',s)
        \bm{\mathscr g}_p^\flat(z')\Phi^\sharp(z,z')\right]d\tau,\\
 -d\bm{\mathscr g}_q^\flat(\tau,z)
 &=dr_q(\tau,z)
   +\beta_0\sum_{z'}
      \mathbb E\!\left[
        \frac{\delta\bm{\mathscr g}_q^\sharp(z)}
             {\delta u^\sharp(z')}\right]
      \bm{\mathscr h}^\flat(z')\,d\tau\\
 &\quad+\eta_\uparrow\gamma_0\beta_0^2
      \sum_{s=0}^{t-1}\mathbb E_{\mathcal B_s}\!\left[
        \sum_p\Delta_p(\bm x',y',s)
        \bm{\mathscr h}^\flat(z')G_{qp}^\sharp(z,z')\right]d\tau.
\end{aligned}
\end{equation}
The minus sign records the backward difference
$\bm{\mathscr g}_q^{\ell\flat}-\bm{\mathscr g}_q^{(\ell+1)\flat}$.
The hidden variables do not accumulate over depth; at each slice they obey
\begin{equation}
\label{eq:sgd-depth-hidden}
\begin{aligned}
 \bm{\mathscr h}^\sharp(\tau,z)
 &=u^\sharp(\tau,z)
   +\beta_0\alpha\sum_{p,z'}
      \mathbb E\!\left[
        \frac{\delta\bm{\mathscr h}^\flat(z)}
             {\delta\,dr_p(z')}\right]
      \bm{\mathscr g}_p^\sharp(z')\\
 &\quad+\eta_\uparrow\gamma_0\beta_0
      \sum_{s=0}^{t-1}\mathbb E_{\mathcal B_s}\!\left[
        \sum_p\Delta_p(\bm x',y',s)
        \bm{\mathscr g}_p^\sharp(z')\Phi^\flat(z,z')\right],\\
 \bm{\mathscr g}_q^\sharp(\tau,z)
 &=\phi'(\bm{\mathscr h}^\sharp(\tau,z))
   \Bigg\{r_q^\sharp(\tau,z)
   +\beta_0\alpha\sum_{z'}
      \mathbb E\!\left[
        \frac{\delta\bm{\mathscr g}_q^\flat(z)}
             {\delta\,du(z')}\right]
      \phi(\bm{\mathscr h}^\sharp(z'))\\
 &\qquad+\eta_\downarrow\gamma_0\beta_0
      \sum_{s=0}^{t-1}\mathbb E_{\mathcal B_s}\!\left[
        \sum_p\Delta_p(\bm x',y',s)
        \phi(\bm{\mathscr h}^\sharp(z'))G_{qp}^\flat(z,z')\right]\Bigg\}.
\end{aligned}
\end{equation}
Responses differentiate the full perturbed computation at fixed
population kernels and loss feedback.
In particular, differentiating $\bm{\mathscr g}_q^\sharp$ includes its
factor $\phi'(\bm{\mathscr h}^\sharp)$.
Derivatives with respect to $du$ and $dr$ mean perturbing a single
depth increment before taking the continuous-depth limit:
\begin{equation}
\label{eq:sgd-depth-response-convention}
\begin{aligned}
 \mathbb E\frac{\delta\bm{\mathscr h}^\flat(\tau,z)}
                    {\delta\,dr_p(\tau,z')}
 &=\lim_{\substack{L\to\infty\\\ell/L\to\tau}}
   \mathbb E\frac{\partial\bm{\mathscr h}^{\ell\flat}(z)}
                    {\partial r_p^{\ell\flat}(z')},\\
 \mathbb E\frac{\delta\bm{\mathscr g}_q^\flat(\tau,z)}
                    {\delta\,du(\tau,z')}
 &=\lim_{\substack{L\to\infty\\\ell/L\to\tau}}
   \mathbb E\frac{\partial\bm{\mathscr g}_q^{(\ell+1)\flat}(z)}
                    {\partial u^{(\ell+1)\flat}(z')}.
\end{aligned}
\end{equation}
The finite-slice sources already include the residual coupling, so
this change of notation introduces no additional scaling factor.

Training the input and readout maps supplies the boundary conditions.
Let $u_{\rm in}(\bm x)$ and $r_{{\rm out},q}$ be centered Gaussian
sources, independent of each other and of the bulk sources, with
\begin{equation}
\label{eq:sgd-depth-boundary-sources}
\begin{aligned}
 \mathbb E[u_{\rm in}(\bm x)u_{\rm in}(\bm x')]
 &=\kappa_{\rm in}(\bm x,\bm x')
   =\frac{\bm x^{\mathsf T}\bm x'}{D_{\mathrm i}},
 &\mathbb E[r_{{\rm out},q}r_{{\rm out},p}]&=\delta_{qp}.
\end{aligned}
\end{equation}
The same sources are reused at every training step. The boundary
updates and output are
\begin{equation}
\label{eq:sgd-depth-boundaries}
\begin{aligned}
 \bm{\mathscr h}^\flat(0,z)
 &=u_{\rm in}(\bm x)
   +\eta_{\rm in}\gamma_0
      \sum_{s=0}^{t-1}\mathbb E_{\mathcal B_s}\!\left[
        \sum_p\Delta_p(\bm x',y',s)
        \bm{\mathscr g}_p^\flat(0,z')\kappa_{\rm in}(\bm x,\bm x')\right],\\
 \bm{\mathscr g}_q^\flat(1,z)
 &=r_{{\rm out},q}
   +\eta_{\rm out}\gamma_0
      \sum_{s=0}^{t-1}\mathbb E_{\mathcal B_s}\!\left[
        \Delta_q(\bm x',y',s)\bm{\mathscr h}^\flat(1,z')\right],\\
 f_q(z)
 &=\frac1{\gamma_0}
      \mathbb E[\bm{\mathscr h}^\flat(1,z)\bm{\mathscr g}_q^\flat(1,z)]\\
 &=\frac1{\gamma_0}
      \mathbb E[\bm{\mathscr h}^\flat(1,z)r_{{\rm out},q}]
   +\eta_{\rm out}
      \sum_{s=0}^{t-1}\mathbb E_{\mathcal B_s}\!\left[
        \Delta_q(\bm x',y',s)\Phi^\flat(1;z,z')\right].
\end{aligned}
\end{equation}
The output determines the loss feedback $\Delta_p$ for the next
simultaneous SGD update, closing the system.
\end{subequations}

With deterministic population coefficients, the residual system and
its boundary conditions are linear in the jointly Gaussian sources.
A unique solution selected by the finite-step computation therefore
has jointly Gaussian residual histories, despite its nonlinear
hidden populations and changing covariance kernels.
The finite identities and Gaussian weight averages are exact; the
population description uses their self-consistent saddle. One route
takes width large at fixed $L$, with $N_\flat/N_\sharp=(N_\flat/K)L$, then increases $L$.
Uniform control of the remainders connects this construction to
simultaneous growth, including bounded hidden width. The same weight
integral extends to tensor-valued channels below, followed by the
corresponding Muon and Adam optimizer-state histories.

\endgroup

\begingroup
\color{black}
\subsection{Attention and other finite-dimensional channels}
\label{app:attention-channel-structure}

System~\eqref{eq:sgd-depth-system} extends to finite-dimensional nonlinear
channels by retaining token and internal coordinates as explicit tensor
axes, separate from the residual and channel populations being averaged.
The scalar activation and its backward derivative become a channel map
and its full adjoint Jacobian; kernels retain both token arguments and
contract internal coordinates without averaging over them.
An entire attention head is one such channel.

\subsubsection{Tensor population equations}
\label{app:sgd-tensor-dmft}
The up--nonlinearity--down structure also describes attention when one
channel contains a full head.
Fix the number of tokens $S$ and the head dimensions
$d_{\rm k},d_{\rm v}$, and let $N_{\sharp\mathrm A}$ be the number of heads in
each of $L$ attention blocks.
Write $K_{\rm A}=LN_{\sharp\mathrm A}$ and set
$d_\uparrow=2d_{\rm k}+d_{\rm v}$.
For token states $\bm h_{\mathfrak{s}}^\flat\in\mathbb R^{N_\flat}$, take
$\bm W_a^\uparrow\in\mathbb R^{d_\uparrow\times {N_\flat}}$ and
$\bm W_a^\downarrow\in\mathbb R^{{N_\flat}\times d_{\rm v}}$.
Suppressing the layer index, one block is
\begin{equation}
\label{eq:attention-channel-finite}
\begin{aligned}
 \bm h_{a,\mathfrak{s}}^{\sharp\mathrm A}&=\frac1{\sqrt {N_\flat}}\bm W_a^\uparrow\bm h_{\mathfrak{s}}^\flat,
 &(\bm o_{a,\mathfrak{s}})_{\mathfrak{s}=1}^{S}
   &=\psi_S((\bm h_{a,\mathfrak{s}}^{\sharp\mathrm A})_{\mathfrak{s}=1}^{S}),\\
 \bm h_{\mathfrak{s}}^{+\flat}
   &=\bm h_{\mathfrak{s}}^\flat+
     \frac{\beta_0\sqrt {N_\flat}}{K_{\rm A}}
     \sum_{a=1}^{N_{\sharp\mathrm A}}\bm W_a^\downarrow\bm o_{a,\mathfrak{s}}.
\end{aligned}
\end{equation}
The input and output maps retain the main-text conventions, applied
tokenwise or followed by a fixed readout over tokens.
All entries of $\bm W_a^\uparrow,\bm W_a^\downarrow$ may be initialized as independent
standard Gaussians; fixed factors associated with the head dimensions
can be included in $\psi_S$ or in the projection constants.

\par
For a general channel map
$\psi_S:\mathbb R^{S\times d_\uparrow}\to\mathbb R^{S\times d_\downarrow}$,
use $K=N_\sharp L$.
For attention, $N_\sharp,K$ denote the head count and budget $N_{\sharp\mathrm A},K_{\rm A}$
in \eqref{eq:attention-channel-finite}.
The numbers $S,d_\uparrow,d_\downarrow$ are fixed.
An attention head is one channel, including its complete token-coupled
map $\psi_S$; its internal dimension is not an additional width limit.
Let $\upsilon=(z,\mathfrak{s})$, where $z=(\bm x,t)$ and $\mathfrak{s}$ is a token index.
Output indices $q,p$ are distinct from token indices.
For a representative channel, introduce a pre-Jacobian cotangent
$R_{\ell,q}(z)\in\mathbb R^{S\times d_\downarrow}$, so that
\begin{equation}
\label{eq:sgd-tensor-channel-adjoint}
 \bm{\mathscr g}_{q,\mathfrak{s} b}^{\ell\sharp}(z)
 =\sum_{\mathfrak{p}=1}^{S}\sum_{c=1}^{d_\downarrow}
   \frac{\partial \psi_S(\bm{\mathscr h}^{\ell\sharp}(z))_{\mathfrak{p} c}}
        {\partial \bm{\mathscr h}_{\mathfrak{s} b}^{\ell\sharp}(z)}
   R_{\ell,q,\mathfrak{p} c}(z).
\end{equation}
This retains the full channel Jacobian, including its cross-token
entries.

The four population kernels are
\begin{equation}
\label{eq:sgd-tensor-kernels}
\begin{aligned}
 \Psi^{\ell\flat}(\upsilon,\upsilon')
   &=\mathbb E[\bm{\mathscr h}^{\ell\flat}(\upsilon)\bm{\mathscr h}^{\ell\flat}(\upsilon')],&
 \Psi^{\ell\sharp}(\upsilon,\upsilon')
   &=\mathbb E\sum_{c=1}^{d_\downarrow}
       \psi_S(\bm{\mathscr h}^{\ell\sharp}(z))_{\mathfrak{s} c}\psi_S(\bm{\mathscr h}^{\ell\sharp}(z'))_{\mathfrak{s}' c},\\
 G_{qp}^{\ell\flat}(\upsilon,\upsilon')
   &=\mathbb E[\bm{\mathscr g}_{q}^{\ell\flat}(\upsilon)\bm{\mathscr g}_{p}^{\ell\flat}(\upsilon')],&
 G_{qp}^{\ell\sharp}(\upsilon,\upsilon')
   &=\mathbb E\sum_{b=1}^{d_\uparrow}
       \bm{\mathscr g}_{q,b}^{\ell\sharp}(\upsilon)\bm{\mathscr g}_{p,b}^{\ell\sharp}(\upsilon').
\end{aligned}
\end{equation}
The corresponding empirical kernels divide by ${N_\flat}$ or ${N_\sharp}$, respectively,
with no division by $S,d_\uparrow$, or $d_\downarrow$.
Thus each kernel retains both token arguments, while the displayed
internal-coordinate contractions are unnormalized traces.

The centered Gaussian sources have nonzero covariances
\begin{equation}
\label{eq:sgd-tensor-source-covariances}
\begin{aligned}
 \mathbb E[u_{b}^{\ell\sharp}(\upsilon)
           u_{b'}^{\ell\sharp}(\upsilon')]
   &=\delta_{bb'}\Psi^{\ell\flat}(\upsilon,\upsilon'),\\
 \mathbb E[r_{q,c}^{\ell\sharp}(\upsilon)
           r_{p,c'}^{\ell\sharp}(\upsilon')]
   &=\delta_{cc'}G_{qp}^{(\ell+1)\flat}(\upsilon,\upsilon'),\\
 \mathbb E[u^{(\ell+1)\flat}(\upsilon)u^{(\ell+1)\flat}(\upsilon')]
   &=\frac{\beta_0^2N_\flat}{KL}\Psi^{\ell\sharp}(\upsilon,\upsilon'),\\
 \mathbb E[r_{q}^{\ell\flat}(\upsilon)
           r_{p}^{\ell\flat}(\upsilon')]
   &=\frac{\beta_0^2N_\flat}{KL}G_{qp}^{\ell\sharp}(\upsilon,\upsilon').
\end{aligned}
\end{equation}
Different source families and layers are independent in this
representation; the resulting trained fields are not.
In particular, $r_{q}^{\ell\sharp}$ has $d_\downarrow$ coordinates
before the Jacobian in \eqref{eq:sgd-tensor-channel-adjoint}.

The four averaged responses are
\begin{equation}
\label{eq:sgd-tensor-responses}
\begin{aligned}
 A_{p}^{\ell\flat}(\upsilon,\upsilon')
   &=\mathbb E\frac{\partial \bm{\mathscr h}^{\ell\flat}(\upsilon)}
                         {\partial r_{p}^{\ell\flat}(\upsilon')},&
 A_{p}^{\ell\sharp}(\upsilon,\upsilon')
   &=\mathbb E\sum_{c=1}^{d_\downarrow}
        \frac{\partial \psi_S(\bm{\mathscr h}^{\ell\sharp}(z))_{\mathfrak{s} c}}
             {\partial r_{p,c}^{\ell\sharp}(\upsilon')},\\
 B_{q}^{\ell\flat}(\upsilon,\upsilon')
   &=\mathbb E\frac{\partial \bm{\mathscr g}_{q}^{(\ell+1)\flat}(\upsilon)}
                         {\partial u^{(\ell+1)\flat}(\upsilon')},&
 B_{q}^{\ell\sharp}(\upsilon,\upsilon')
   &=\mathbb E\sum_{b=1}^{d_\uparrow}
        \frac{\partial \bm{\mathscr g}_{q,b}^{\ell\sharp}(\upsilon)}
             {\partial u_{b}^{\ell\sharp}(\upsilon')}.
\end{aligned}
\end{equation}
As in \eqref{eq:sgd-population-responses}, these derivatives hold the
population kernels and loss feedback fixed and differentiate the
entire perturbed computation.
The two hidden responses trace the matching internal index but retain
the two token arguments, including cross-token responses.
The self-consistent hidden and residual equations are
\begin{equation}
\label{eq:sgd-tensor-hidden}
\begin{aligned}
 \bm{\mathscr h}_{\mathfrak{s} b}^{\ell\sharp}(z)
 &=u_{\mathfrak{s} b}^{\ell\sharp}(z)
   +\beta_0\frac{N_\flat}{K}\sum_{p,z',\mathfrak{r}}
       A_{p}^{\ell\flat}(z,\mathfrak{s};z',\mathfrak{r})
       \bm{\mathscr g}_{p,\mathfrak{r} b}^{\ell\sharp}(z')\\
 &\quad+\eta_\uparrow\gamma_0\beta_0\,
     \mathcal S_t\!\left[\sum_{\mathfrak{r}}
       \Psi^{\ell\flat}(z,\mathfrak{s};z',\mathfrak{r})
       \bm{\mathscr g}_{p,\mathfrak{r} b}^{\ell\sharp}(z')\right],\\
 R_{\ell,q,\mathfrak{s} c}(z)
 &=r_{q,\mathfrak{s} c}^{\ell\sharp}(z)
   +\beta_0\frac{N_\flat}{K}\sum_{z',\mathfrak{r}}
       B_{q}^{\ell\flat}(z,\mathfrak{s};z',\mathfrak{r})
       \psi_S(\bm{\mathscr h}^{\ell\sharp}(z'))_{\mathfrak{r} c}\\
 &\quad+\eta_\downarrow\gamma_0\beta_0\,
     \mathcal S_t\!\left[\sum_{\mathfrak{r}}
       G_{qp}^{(\ell+1)\flat}(z,\mathfrak{s};z',\mathfrak{r})
       \psi_S(\bm{\mathscr h}^{\ell\sharp}(z'))_{\mathfrak{r} c}\right].
\end{aligned}
\end{equation}
\begin{equation}
\label{eq:sgd-tensor-residual}
\begin{aligned}
 \bm{\mathscr h}_{\mathfrak{s}}^{(\ell+1)\flat}(z)-\bm{\mathscr h}_{\mathfrak{s}}^{\ell\flat}(z)
 &=u_{\mathfrak{s}}^{(\ell+1)\flat}(z)
   +\frac{\beta_0}{L}\sum_{p,z',\mathfrak{r}}
       A_{p}^{\ell\sharp}(z,\mathfrak{s};z',\mathfrak{r})
       \bm{\mathscr g}_{p,\mathfrak{r}}^{(\ell+1)\flat}(z')\\
 &\quad+\frac{\eta_\downarrow\gamma_0\beta_0^2}{L}\,
     \mathcal S_t\!\left[\sum_{\mathfrak{r}}
       \Psi^{\ell\sharp}(z,\mathfrak{s};z',\mathfrak{r})
       \bm{\mathscr g}_{p,\mathfrak{r}}^{(\ell+1)\flat}(z')\right],\\
 \bm{\mathscr g}_{q,\mathfrak{s}}^{\ell\flat}(z)-\bm{\mathscr g}_{q,\mathfrak{s}}^{(\ell+1)\flat}(z)
 &=r_{q,\mathfrak{s}}^{\ell\flat}(z)
   +\frac{\beta_0}{L}\sum_{z',\mathfrak{r}}
       B_{q}^{\ell\sharp}(z,\mathfrak{s};z',\mathfrak{r})
       \bm{\mathscr h}_{\mathfrak{r}}^{\ell\flat}(z')\\
 &\quad+\frac{\eta_\uparrow\gamma_0\beta_0^2}{L}\,
     \mathcal S_t\!\left[\sum_{\mathfrak{r}}
       G_{qp}^{\ell\sharp}(z,\mathfrak{s};z',\mathfrak{r})
       \bm{\mathscr h}_{\mathfrak{r}}^{\ell\flat}(z')\right].
\end{aligned}
\end{equation}
The response sums run over the same finite query set as in the MLP
construction. Only the history terms carry the loss-weighted,
strictly earlier-step sum $\mathcal S_t$.

For completeness, a shared input map and a fixed linear token readout
make the boundary contractions explicit. Let
$h^{0\flat}_{i\mathfrak{s}}=D_{\mathrm i}^{-1/2}\sum_j
W^0_{ij}x_{\mathfrak{s} j}$ and
$f_q=(\gamma_0{N_\flat})^{-1}\sum_{i,\mathfrak{s}}
W^L_{qi}\omega_{\mathfrak{s}} h^{L\flat}_{i\mathfrak{s}}$,
where $\omega_{\mathfrak{s}}$ is fixed. With
$\kappa_{\rm in}(\bm x,\mathfrak{s};\bm x',\mathfrak{r})
=\bm x_{\mathfrak{s}}^{\mathsf T}\bm x'_{\mathfrak{r}}/D_{\mathrm i}$, the counterparts
of \eqref{eq:sgd-dmft-boundaries} are
\begin{equation}
\label{eq:sgd-tensor-boundaries}
\begin{aligned}
 \bm{\mathscr h}_{\mathfrak{s}}^{0\flat}(z)
 &=u_{{\rm in},\mathfrak{s}}(\bm x)
   +\eta_{\rm in}\gamma_0\,
      \mathcal S_t\!\left[\sum_{\mathfrak{r}}
       \kappa_{\rm in}(\bm x,\mathfrak{s};\bm x',\mathfrak{r})
       \bm{\mathscr g}_{p,\mathfrak{r}}^{0\flat}(z')\right],\\
 \bm{\mathscr g}_{q,\mathfrak{s}}^{L\flat}(z)
 &=\omega_{\mathfrak{s}}\left\{
    r_{{\rm out},q}
    +\eta_{\rm out}\gamma_0\,
       \mathcal S_t\!\left[\delta_{pq}
                   \sum_{\mathfrak{r}}\omega_{\mathfrak{r}} \bm{\mathscr h}_{\mathfrak{r}}^{L\flat}(z')\right]\right\},\\
 f_q(z)&=\gamma_0^{-1}\mathbb E
                    \sum_{\mathfrak{s}} \bm{\mathscr h}_{\mathfrak{s}}^{L\flat}(z)\bm{\mathscr g}_{q,\mathfrak{s}}^{L\flat}(z).
\end{aligned}
\end{equation}
Here $u_{\rm in}$ has covariance $\kappa_{\rm in}$ and
$r_{{\rm out},q}$ are independent standard Gaussians, reused throughout
training as in the MLP construction.
Other fixed token readouts give their corresponding finite contractions.
Equations \eqref{eq:sgd-tensor-kernels}--\eqref{eq:sgd-tensor-boundaries},
including the loss derivatives obtained from this readout, form the
general-channel population closure.
\par
Setting $S=d_\uparrow=d_\downarrow=1$, $\omega_1=1$, and $\psi_1=\phi$
identifies these $\Psi$ kernels with the MLP $\Phi$ kernels and recovers
\eqref{eq:sgd-dmft-hidden}--\eqref{eq:sgd-dmft-boundaries}.
The continuous-depth passage uses the same equations with the
displayed token kernels: residual source covariances become
$\beta_0^2\alpha\Psi^\sharp\,d\tau$ and
$\beta_0^2\alpha G^\sharp\,d\tau$, and each residual reaction or
training term proportional to $1/L$ becomes its corresponding drift.
The hidden equations remain local to the depth $\tau$, with $N_\flat/K$
replaced by $\alpha$.
The fixed internal dimensions remain inside $\psi_S$, its Jacobian,
and the unnormalized traces throughout.

The suppression of local learning can also be read directly from
these general-channel equations.
At fixed $c=\beta_0^2\alpha$, the hidden reaction coefficient is
$\beta_0\alpha=c/\beta_0$, while the two direct SGD history
coefficients are
$\eta_{{\rm eff},j}=\eta_j\gamma_0\beta_0$,
$j\in\{\uparrow,\downarrow\}$.
Along a family with $\eta_{{\rm eff},j}=O(\beta_0^{-1})$,
\eqref{eq:sgd-tensor-hidden} gives, on a finite query and training
window with bounded residual kernels, responses, loss feedback,
and channel second moments,
\begin{equation}
\label{eq:tensor-semifreeze-deformation}
 \|\bm{\mathscr h}_\beta^\sharp-u_\beta^\sharp\|_{L^2}
 +\|R_\beta-r_\beta^\sharp\|_{L^2}
 =O(\beta_0^{-1}).
\end{equation}
Thus both the direct learning histories and the reaction corrections
disappear from the local hidden equations.
This applies to a finite-dimensional channel map $\psi_S$, including
attention, through its full Jacobian rather than a coordinatewise
activation derivative.
The channel output $\psi_S(\bm{\mathscr h}_\beta^\sharp)$ and its backward Jacobian action
can remain nonlinear, and the Gaussian-source covariance may still
evolve through the residual dynamics. The degeneration concerns
direct local learning, not the disappearance of all activation changes.
\par
\par

\subsubsection{Gaussian weight calculation}
The Gaussian weight integral makes the required contractions explicit.
For the finite fields at layer $\ell$, use
$h_{i\mathfrak{s}}^{\ell\flat},g_{q,i\mathfrak{s}}^{\ell\flat},h_{a\mathfrak{s} b}^{\ell\sharp},\psi_S(\bm h_a^{\ell\sharp})_{\mathfrak{s} c},g_{q,a\mathfrak{s} b}^{\ell\sharp}$
for the finite fields before selecting representative coordinates.
The four initial-weight projections are
\begin{equation}
\label{eq:sgd-tensor-initial-projections}
\begin{aligned}
 \chi^\sharp_{a\mathfrak{s} b}
   &={N_\flat}^{-1/2}\sum_i W^{\ell\uparrow}_{abi}(0)h_{i\mathfrak{s}}^{\ell\flat},&
 \xi^\sharp_{q,a\mathfrak{s} c}
   &={N_\flat}^{-1/2}\sum_i W^{\ell\downarrow}_{iac}(0)g_{q,i\mathfrak{s}}^{(\ell+1)\flat},\\
 \chi^\flat_{i\mathfrak{s}}
   &=(\beta_0\sqrt {N_\flat}/K)\sum_{a,c}W^{\ell\downarrow}_{iac}(0)\psi_S(\bm h_a^{\ell\sharp})_{\mathfrak{s} c},&
 \xi^\flat_{q,i\mathfrak{s}}
   &=(\beta_0\sqrt {N_\flat}/K)\sum_{a,b}W^{\ell\uparrow}_{abi}(0)g_{q,a\mathfrak{s} b}^{\ell\sharp}.
\end{aligned}
\end{equation}
The backward residual field is evaluated at $\ell+1$, and the residual projections already include
their branch coefficient. After introducing these constraints and
their Fourier fields, the coefficients of the individual Gaussian
entries are
\begin{equation}
\label{eq:sgd-tensor-entry-sources}
\begin{aligned}
 J^\uparrow_{abi}
 &=\sum_{z,\mathfrak{s}}\left[
     \frac{\widehat\chi^\sharp_{a\mathfrak{s} b}(z)h_{i\mathfrak{s}}^{\ell\flat}(z)}{\sqrt {N_\flat}}
     +(\beta_0\sqrt {N_\flat}/K)\sum_q
       \widehat\xi^\flat_{q,i\mathfrak{s}}(z)g_{q,a\mathfrak{s} b}^{\ell\sharp}(z)\right],\\
 J^\downarrow_{iac}
 &=\sum_{z,\mathfrak{s}}\left[
     (\beta_0\sqrt {N_\flat}/K)\widehat\chi^\flat_{i\mathfrak{s}}(z)\psi_S(\bm h_a^{\ell\sharp}(z))_{\mathfrak{s} c}
     +\frac1{\sqrt {N_\flat}}\sum_q
       \widehat\xi^\sharp_{q,a\mathfrak{s} c}(z)g_{q,i\mathfrak{s}}^{(\ell+1)\flat}(z)\right].
\end{aligned}
\end{equation}
Consequently the exact initial-weight average in this constrained
integral is
\begin{equation}
\label{eq:sgd-tensor-gaussian-integral}
 \mathbb E_{\bm W^{\ell\uparrow}(0),\bm W^{\ell\downarrow}(0)}
 e^{-i\sum_{a,b,i}W^{\ell\uparrow}_{abi}(0)J^\uparrow_{abi}
    -i\sum_{i,a,c}W^{\ell\downarrow}_{iac}(0)J^\downarrow_{iac}}
 =\!
 \exp\!\left[
   -\frac12\sum_{a,b,i}(J^\uparrow_{abi})^2
   -\frac12\sum_{i,a,c}(J^\downarrow_{iac})^2\right].
\end{equation}
The same weight is used at every token, so these squares contain
double token sums, rather than a token Kronecker delta.
For example, the forward down-projection term is
\begin{equation}
\label{eq:sgd-tensor-token-contraction}
 -\frac{(\beta_0\sqrt {N_\flat}/K)^2}{2}
 \sum_{\upsilon,\upsilon'}
 \left[\sum_i\widehat\chi^\flat_i(\upsilon)
                  \widehat\chi^\flat_i(\upsilon')\right]
 \left[\sum_{a,c}\psi_S(\bm h_a^{\ell\sharp}(z))_{\mathfrak{s} c}\psi_S(\bm h_a^{\ell\sharp}(z'))_{\mathfrak{s}' c}\right].
\end{equation}
The mixed terms have coefficient $(\beta_0\sqrt {N_\flat}/K)/\sqrt {N_\flat}=\beta_0/K$;
they generate the weight-reuse responses in
\eqref{eq:sgd-tensor-responses}.
Summing the mixed weight term over residual coordinates gives
$\beta_0 {N_\flat}/K=\beta_0\frac{N_\flat}{K}$; summing it over channels gives
$\beta_0{N_\sharp}/K=\beta_0/L$.

The training terms follow from synchronous SGD without taking a
gradient-flow limit. In the mathematical weight coordinates,
\eqref{eq:attention-channel-parameter-gradients} gives the exact histories
\begin{equation}
\label{eq:sgd-tensor-weight-histories}
\begin{aligned}
 W^{\ell\uparrow}_{abi}(t)
 &=W^{\ell\uparrow}_{abi}(0)
   +\frac{\eta_\uparrow\gamma_0\beta_0}{\sqrt {N_\flat}}
     \mathcal S_t\!\left[
       \sum_{\mathfrak{r}=1}^S
       g^{\ell\sharp}_{p,a\mathfrak{r} b}(z')h^{\ell\flat}_{i\mathfrak{r}}(z')\right],\\
 W^{\ell\downarrow}_{iac}(t)
 &=W^{\ell\downarrow}_{iac}(0)
   +\frac{\eta_\downarrow\gamma_0\beta_0}{\sqrt {N_\flat}}
     \mathcal S_t\!\left[
       \sum_{\mathfrak{r}=1}^S
       g^{(\ell+1)\flat}_{p,i\mathfrak{r}}(z')\psi_S(\bm h_a^{\ell\sharp}(z'))_{\mathfrak{r} c}\right].
\end{aligned}
\end{equation}
All fields inside $\mathcal S_t$ are evaluated before the corresponding
simultaneous update, as specified in \eqref{eq:sgd-history-sum}.
Substituting these histories and decoupling the initial-weight integral
gives \eqref{eq:sgd-tensor-hidden}--\eqref{eq:sgd-tensor-boundaries}.
\par
\par

\subsubsection{Attention-specific structure}
For masked softmax attention, split $\bm h_{a,\mathfrak{s}}^{\sharp\mathrm A}$ into
$(\bm q_{a,\mathfrak{s}},\bm k_{a,\mathfrak{s}},\bm v_{a,\mathfrak{s}})$ with dimensions
$(d_{\rm k},d_{\rm k},d_{\rm v})$ and define
\begin{equation}
\label{eq:attention-channel-softmax}
\begin{aligned}
 p_{a,\mathfrak{s}\mathfrak{r}}
 &=\frac{M_{\mathfrak{s}\mathfrak{r}}\exp(\bm q_{a,\mathfrak{s}}^{\mathsf T}\bm k_{a,\mathfrak{r}}/\sqrt{d_{\rm k}})}
         {\sum_{\mathfrak{r}'=1}^{S}M_{\mathfrak{s}\mathfrak{r}'}
          \exp(\bm q_{a,\mathfrak{s}}^{\mathsf T}\bm k_{a,\mathfrak{r}'}/\sqrt{d_{\rm k}})},\\
 \bm o_{a,\mathfrak{s}}&=\sum_{\mathfrak{r}=1}^{S}p_{a,\mathfrak{s}\mathfrak{r}}\bm v_{a,\mathfrak{r}}.
\end{aligned}
\end{equation}
Here $M$ is a fixed binary mask with at least one visible token per row.
The hidden computation mixes tokens and head coordinates, while the
residual coordinate is accessed only through the two linear maps.
The count corresponding to $N_\sharp$ is the number of heads, not the
head dimension. In a network alternating MLP and attention updates,
the two populations have separate counts $N_{\sharp\mathrm M},N_{\sharp\mathrm A}$ and
budgets $K_{\rm M}=L_{\rm M}N_{\sharp\mathrm M}$,
$K_{\rm A}=L_{\rm A}N_{\sharp\mathrm A}$, with $L_{\rm M},L_{\rm A}$ counting
their respective residual updates.

The backward variables use the same rescaling as in the MLP.
Regard $\bm h_a^{\sharp\mathrm A}$ and $\bm o_a$ as arrays over tokens, and let
$D\psi_S(\bm h_a^{\sharp\mathrm A})^*$ be the adjoint Jacobian of the head computation.
For output component $q$, set
$\bm g_{q,\mathfrak{s}}^\flat=\gamma_0{N_\flat}\,\partial f_q/\partial\bm h_{\mathfrak{s}}^\flat$
and
$\bm g_{q,a,\mathfrak{s}}^{\sharp\mathrm A}=(\gamma_0K_{\rm A}/\beta_0)
\,\partial f_q/\partial\bm h_{a,\mathfrak{s}}^{\sharp\mathrm A}$.
The chain rule gives
\begin{equation}
\label{eq:attention-channel-backward}
\begin{aligned}
 \bm g_{q,a}^{\sharp\mathrm A}
 &=\frac1{\sqrt {N_\flat}}D\psi_S(\bm h_a^{\sharp\mathrm A})^*
       \bigl(((\bm W_a^\downarrow)^{\mathsf T}\bm g_{q,\mathfrak{s}}^{+\flat})_{\mathfrak{s}=1}^{S}\bigr),\\
 \bm g_{q,\mathfrak{s}}^\flat
 &=\bm g_{q,\mathfrak{s}}^{+\flat}
   +\frac{\beta_0\sqrt {N_\flat}}{K_{\rm A}}
           \sum_{a=1}^{N_{\sharp\mathrm A}}(\bm W_a^\uparrow)^{\mathsf T}\bm g_{q,a,\mathfrak{s}}^{\sharp\mathrm A}.
\end{aligned}
\end{equation}
Thus the hidden derivative is the Jacobian of a finite-dimensional
channel, in place of multiplication by the scalar $\phi'$.
The two parameter Jacobians are
\begin{equation}
\label{eq:attention-channel-parameter-gradients}
\begin{aligned}
 \frac{\partial f_q}{\partial\bm W_a^\uparrow}
 &=\frac{\beta_0}{\gamma_0K_{\rm A}\sqrt {N_\flat}}
       \sum_{\mathfrak{s}=1}^{S}\bm g_{q,a,\mathfrak{s}}^{\sharp\mathrm A}
                        (\bm h_{\mathfrak{s}}^\flat)^{\mathsf T},\\
 \frac{\partial f_q}{\partial\bm W_a^\downarrow}
 &=\frac{\beta_0}{\gamma_0K_{\rm A}\sqrt {N_\flat}}
       \sum_{\mathfrak{s}=1}^{S}\bm g_{q,\mathfrak{s}}^{+\flat}
                        \bm o_{a,\mathfrak{s}}^{\mathsf T}.
\end{aligned}
\end{equation}
Token summation changes the finite internal contractions, while the
size factors in the forward, backward, and parameter updates retain
the MLP form.

At initialization, this replacement also specifies the covariance
map. Index a finite collection of inputs and tokens by $z=(\bm x,\mathfrak{s})$,
and write $C_{zz'}={N_\flat}^{-1}(\bm h_z^\flat)^{\mathsf T}\bm h_{z'}^\flat$.
Let $Z$ be the Gaussian array with
$\mathbb E[Z_{r,z}Z_{r',z'}]=\delta_{rr'}C_{zz'}$,
where $r=1,\ldots,d_\uparrow$, and define
\begin{equation}
\label{eq:attention-initial-covariance-map}
 [\mathcal V_{\psi}(C)]_{zz'}
 =\mathbb E\!\left[
     \psi_S(Z(\bm x))_{\mathfrak{s}}^{\mathsf T}
     \psi_S(Z(\bm x'))_{\mathfrak{s}'}\right].
\end{equation}
For the block increment $\delta\bm h_z^\flat$,
\begin{equation}
\label{eq:attention-initial-increment}
 \mathbb E[\delta h_{i,z}^\flat\delta h_{j,z'}^\flat
             \mid\bm h^\flat]
 =\delta_{ij}\frac{\beta_0^2({N_\flat}/K_{\rm A})}{L}
                    [\mathcal V_{\psi}(C)]_{zz'}.
\end{equation}
The continuous-depth initialization covariance therefore obeys
$\partial_\tau C=\beta_0^2\alpha_{\rm A}\mathcal V_{\psi}(C)$
when ${N_\flat}/K_{\rm A}\to\alpha_{\rm A}$.
The covariance map keeps the attention nonlinearity and token
interactions; the scalar MLP map is its one-dimensional channel case.

This finite computation has the same residual rotation symmetry as the
MLP example.
For a common orthogonal $\bm R\in\mathbb R^{{N_\flat}\times {N_\flat}}$, transform
every residual vector by $\bm R$, every up map by
$\bm W_a^\uparrow\mapsto\bm W_a^\uparrow\bm R^{\mathsf T}$, and every down map by
$\bm W_a^\downarrow\mapsto\bm R\bm W_a^\downarrow$; the input and output weights transform
as $\bm W^0\mapsto\bm R\bm W^0$ and
$\bm W^L\mapsto\bm W^L\bm R^{\mathsf T}$.
Each $\bm h_{a,\mathfrak{s}}^{\sharp\mathrm A}$, attention matrix, and output is unchanged.
Gradients transform in the corresponding way by the chain rule.
Independent isotropic Gaussian initialization, SGD with scalar rates
for each parameter group, and orthogonally equivariant spectral updates
preserve this symmetry.
Consequently the residual Gaussianity argument applies to this
structure whenever its mixed residual Gram matrices have deterministic
limits. The Gaussian weight calculation above supplies the corresponding
self-consistent training equations.

Normalization and shared parameters must respect the same conditions
before that conclusion is used.
A residual normalization $\mathcal N$ preserves the displayed symmetry
if $\mathcal N(\bm R\bm h)=\bm R\mathcal N(\bm h)$ for every orthogonal
$\bm R$. RMS normalization without coordinatewise affine parameters,
or with a scalar gain, has this property.
Mean-subtracting LayerNorm and coordinatewise gains do not have full
residual rotation equivariance.
Tying input and output weights may still respect the rotation, but
changes their joint initialization and training updates; it is not
covered by the independent-endpoint convention.
Growing token lengths or head dimensions also require a separate scaling
analysis.

For spectral optimizers, the matrix on which the transform acts is an
additional part of the specification.
Packing all heads into a layer matrix gives projection dimensions
proportional to $N_{\sharp\mathrm A}$ and ${N_\flat}$, consistent with the width-dependent
spectral-sampling comparison in the MLP analysis.
Applying the transform separately to each fixed-dimensional head
preserves residual rotation equivariance, but does not enlarge that
local spectral window when $N_{\sharp\mathrm A}$ grows.
The MLP spectral-error estimate therefore cannot be transferred to
per-head updates merely by replacing $N_\sharp$ with $N_{\sharp\mathrm A}$.
\par

\subsection{Spectral updates mix feature histories}
\label{app:spectral-history-bridge}

For the spectral-clipping model of Muon, the explicit SGD learning
histories in system~\eqref{eq:sgd-depth-system} are replaced by
feature-history contractions mixed by a matrix determined by their
Gram matrices. Momentum enlarges these histories without changing the
construction. The initial-weight integral and its source and reaction
prefactors retain their form, with kernels and responses evaluated
under the resulting Muon law.

A spectral update acts within the feature directions present in its
gradient. For a fixed batch and finitely many training steps, collect
these directions in two history matrices. Write
\begin{equation}
\label{eq:spectral-history-factorization}
\begin{aligned}
 D&=s\overline A C\overline B^{\mathsf T},&
 \overline A&=A/\sqrt p,&\overline B&=B/\sqrt q,\\
 P&=A^{\mathsf T}A/p,&Q&=B^{\mathsf T}B/q,&
 T&=P^{1/2}CQ^{1/2},
\end{aligned}
\end{equation}
where $s>0$, $A\in\mathbb R^{p\times r}$,
$B\in\mathbb R^{q\times r}$ and $C\in\mathbb R^{r\times r}$.
The normalized clipping map in \eqref{eq:muon-spectral-clip}
satisfies the exact identity
\begin{equation}
\label{eq:spectral-history-mixing}
 F_\lambda(D)=\overline A\mathcal M\overline B^{\mathsf T},
 \qquad
 \mathcal M=P^{\dagger/2}F_{\lambda/s}(T)Q^{\dagger/2}.
\end{equation}
Here $P^{\dagger/2}$ is the positive square root of the Moore--Penrose
inverse. The formula includes singular Gram matrices.
For an up-projection gradient, the columns of $A,B$ are respectively
$\bm g_q^{\ell\sharp}$ and $\bm h^{\ell\flat}$, indexed by batch
examples and output components. For a down-projection they are
$\bm g_q^{(\ell+1)\flat}$ and $\phi(\bm h^{\ell\sharp})$.
The diagonal entries of $C$ are loss derivatives divided by batch size.
Equation~\eqref{eq:sgd-parameter-jacobians} gives, for both projections,
\begin{equation}
 s=\frac{\beta_0\sqrt {N_\sharp}}{\gamma_0K},\qquad
 \lambda=\lambda_0\sqrt {N_\sharp}/K,\qquad
 \frac{\lambda}{s}=\frac{\lambda_0\gamma_0}{\beta_0}.
\end{equation}
Thus the central spectral map has no remaining size factor.
Linear momentum also has this representation: concatenating the
feature columns from steps $0,\ldots,t$ and using coefficient blocks
$(1-\mu)\mu^{t-j}C_j$ gives
$M_t=\mu M_{t-1}+(1-\mu)D_t$, with $M_{-1}=0$.
No independence of the trained histories is used in these identities.

On a fixed nonzero spectral support with eigenvalues bounded away from
zero, concentration of the history Gram matrices and loss coefficients
determines the limit of the mixing matrix $\mathcal M$.
The finite identity itself does not require this restriction.

For bulk rates $\eta_M=\rho\sqrt{N_\sharp}$, the accumulated
forward and backward histories take the following form.
At update $j$, label the history columns by $z_a=(\bm x_a,t_a)$
and their output components by $p_a$; momentum includes columns
from all steps through $j$. Suppressing the layer index only on
the mixing matrices, define
\begin{equation}
\label{eq:muon-history-contractions}
\begin{aligned}
 \bm{\mathcal T}_\uparrow(z)
 &=\sum_{j<t}\sum_{a,b}
   \bm g_{p_a}^{\ell\sharp}(z_a)
   (\mathcal M_{\uparrow,j})_{ab}\Phi^{\ell\flat}(z,z_b),\\
 \bm{\mathcal T}_\downarrow(z)
 &=\sum_{j<t}\sum_{a,b}
   \bm g_{p_a}^{(\ell+1)\flat}(z_a)
   (\mathcal M_{\downarrow,j})_{ab}\Phi^{\ell\sharp}(z,z_b),\\
 \bm{\mathcal V}_{\downarrow,q}(z)
 &=\sum_{j<t}\sum_{a,b}
   \phi(\bm h^{\ell\sharp}(z_b))
   (\mathcal M_{\downarrow,j})_{ab}G_{qp_a}^{(\ell+1)\flat}(z,z_a),\\
 \bm{\mathcal V}_{\uparrow,q}(z)
 &=\sum_{j<t}\sum_{a,b}
   \bm h^{\ell\flat}(z_b)
   (\mathcal M_{\uparrow,j})_{ab}G_{qp_a}^{\ell\sharp}(z,z_a).
\end{aligned}
\end{equation}
The kernels here are the empirical inner products, and all fields
are evaluated in the trained network. With the initial-weight
projections already defined, the four exact identities are
\begin{equation}
\label{eq:muon-four-histories}
\begin{aligned}
 \bm h^{\ell\sharp}
 &=\bm\chi^{\ell\sharp}-\rho_\uparrow\bm{\mathcal T}_\uparrow,\\
 \bm h^{(\ell+1)\flat}-\bm h^{\ell\flat}
 &=\bm\chi^{(\ell+1)\flat}
       -\frac{\beta_0\rho_\downarrow}{L}\bm{\mathcal T}_\downarrow,\\
 \bm g_q^{\ell\sharp}
 &=\phi'(\bm h^{\ell\sharp})\odot
     \left(\bm\xi_q^{\ell\sharp}
                  -\rho_\downarrow\bm{\mathcal V}_{\downarrow,q}\right),\\
 \bm g_q^{\ell\flat}-\bm g_q^{(\ell+1)\flat}
 &=\bm\xi_q^{\ell\flat}
       -\frac{\beta_0\rho_\uparrow}{L}\bm{\mathcal V}_{\uparrow,q}.
\end{aligned}
\end{equation}
The minus signs reflect the positive loss gradients in the spectral
factorization; $\mathcal S_t$ in the SGD formulas instead includes
negative loss feedback.

The initial-weight integral
\eqref{eq:sgd-pi-weight-average} is unchanged: only the learning
histories attached to its projection constraints differ.
It gives the source covariances
\eqref{eq:sgd-population-sources} and reaction prefactors
\eqref{eq:sgd-pi-current-coefficients}, with kernels and responses
evaluated under the Muon law.
In \eqref{eq:sgd-dmft-hidden}--\eqref{eq:sgd-dmft-residual},
replace the four explicit SGD training terms by the corresponding
contractions in \eqref{eq:muon-four-histories}, replacing empirical
inner products and coordinate fields by their population versions.
The responses differentiate this full computation, including the
spectral update, while macroscopic kernels are fixed.
Continuous depth replaces the residual $1/L$ terms by $d\tau$.
For a nonlinear spectral map, its mixing coefficients depend on both
hidden and residual history Gram matrices; the growing hidden
population is therefore essential to this construction.

\paragraph{Trainable endpoints.}
The same construction replaces the two boundary histories in
\eqref{eq:sgd-depth-boundaries}. Let $D_s^j$ be the positive loss-gradient
matrix of the input or output map, obtained from
\eqref{eq:sgd-parameter-jacobians} by the batch loss contraction, and
$M_s^j=\mu M_{s-1}^j+(1-\mu)D_s^j$, with $M_{-1}^j=0$.
Using $\eta_{M,j}=\rho_j\sqrt{N_\flat}$ and
$\lambda_j=\lambda_{0,j}/\sqrt{N_\flat}$ gives
\begin{equation}
\label{eq:muon-endpoint-histories}
\begin{aligned}
 \bm h^{0\flat}(\bm x,t)
 &=\frac{\bm W^0(0)\bm x}{\sqrt{D_{\mathrm i}}}
   -\frac{\eta_{M,\rm in}}{\sqrt{D_{\mathrm i}}}
       \sum_{s<t}F_{\lambda_{\rm in}}(M_s^{\rm in})\bm x,\\
 \bm g_q^{L\flat}(t)
 &=(\bm W^L(0))^{\mathsf T}\bm e_q
   -\eta_{M,\rm out}\sum_{s<t}
       F_{\lambda_{\rm out}}(M_s^{\rm out})^{\mathsf T}\bm e_q .
\end{aligned}
\end{equation}
Here $\bm e_q$ is the $q$th output basis vector; $\mu=0$ removes momentum.
Applying \eqref{eq:spectral-history-mixing} to these rectangular maps
and replacing residual history Gram matrices by their population
values gives the boundary laws. Task-coordinate sums remain finite.
The readout identity in \eqref{eq:sgd-depth-boundaries} is unchanged.

\paragraph{Factorization and update scales.}
Indeed, polar decomposition gives
$\overline A=J_AP^{1/2}$ and $\overline B=J_BQ^{1/2}$, with partial
isometries $J_A,J_B$. A singular-value decomposition on the supports of
$P,Q$ gives $F_\lambda(D)=J_AF_{\lambda/s}(T)J_B^{\mathsf T}$.
Substituting $J_A=\overline AP^{\dagger/2}$ and
$J_B=\overline BQ^{\dagger/2}$ proves \eqref{eq:spectral-history-mixing}.
In the linear spectral region the product reduces to $D/\lambda$.

For bulk Muon rates $\eta_M=\rho\sqrt {N_\sharp}$, put
$k_B(v)=B^{\mathsf T}v/q$, using the input dimension $q$ of the
relevant projection. Holding the incoming field fixed gives
\begin{equation}
\label{eq:muon-local-history-actions}
\begin{aligned}
 \left.\Delta\bm h^{\ell\sharp}\right|_{\rm up}
 &=-\rho_\uparrow A_\uparrow\mathcal M_\uparrow
       k_{B_\uparrow}(\bm h^{\ell\flat}),\\
 \left.\Delta\bm h^{(\ell+1)\flat}\right|_{\rm down}
 &=-\frac{\rho_\downarrow\beta_0}{L}
       A_\downarrow\mathcal M_\downarrow
       k_{B_\downarrow}(\phi(\bm h^{\ell\sharp})).
\end{aligned}
\end{equation}

\subsection{Adam retains coordinate-pair moment histories}
\label{app:adam-history-bridge}

Adam replaces the explicit SGD learning histories in
system~\eqref{eq:sgd-depth-system} by population averages of normalized
moment updates for residual--hidden coordinate pairs.
The moment state is computed from each pair's gradient history before
averaging over the source population. These contractions depend on
the full history law, rather than only its second-moment kernels,
and can be nonlinear in the remaining residual history.

Adam normalizes gradient entries separately. Write a parameter group's
gradient as $D_t^j=d_jR_t^j$, with gradient-entry scale $d_j>0$,
and use a stabilizer $\epsilon_j=d_j\epsilon_0$, with $\epsilon_0>0$.
The normalized moment states, initially zero, obey
\begin{equation}
\label{eq:adam-normalized-state}
\begin{aligned}
 M_t^j&=a_1M_{t-1}^j+(1-a_1)R_t^j,&
 V_t^j&=a_2V_{t-1}^j+(1-a_2)(R_t^j)^{\odot2},\\
 \widehat M_t^j&=\frac{M_t^j}{1-a_1^{t+1}},&
 \widehat V_t^j&=\frac{V_t^j}{1-a_2^{t+1}},\\
 \mathcal U_t^j&=\frac{\widehat M_t^j}
                         {\sqrt{\widehat V_t^j}+\epsilon_0},&
 W^j(t+1)&=W^j(t)-\eta_j^{\rm W}\mathcal U_t^j.
\end{aligned}
\end{equation}
The square, square root and division are entrywise.
This is an exact change of gradient units, with fixed moment
coefficients $a_1,a_2\in[0,1)$.
For the bulk projections, $d_\uparrow=d_\downarrow=
\beta_0/(\gamma_0K\sqrt {N_\flat})$, and
\begin{equation}
\label{eq:adam-normalized-bulk-gradients}
\begin{aligned}
 (R_t^\uparrow)_{ai}
 &=\frac1{B_t}\sum_{(\bm x,y)\in\mathcal B_t}\sum_q
       \partial_{f_q}\mathcal L_{\rm ex}\,
       g_{q,a}^{\ell\sharp}(\bm x,t)h_i^{\ell\flat}(\bm x,t),\\
 (R_t^\downarrow)_{ia}
 &=\frac1{B_t}\sum_{(\bm x,y)\in\mathcal B_t}\sum_q
       \partial_{f_q}\mathcal L_{\rm ex}\,
       g_{q,i}^{(\ell+1)\flat}(\bm x,t)\phi(h_a^{\ell\sharp}(\bm x,t)).
\end{aligned}
\end{equation}
The bulk rates are $\eta_j^{\rm W}=\rho_j/\sqrt{N_\flat}$.

The population variables for Adam include the moment state of a
residual--hidden coordinate pair. For an up map, its normalized
gradient is the batch average of $\bm{\mathscr g}_q^{\ell\sharp}\bm{\mathscr h}^{\ell\flat}$; for a down map it is the
batch average of $\bm{\mathscr g}_q^{(\ell+1)\flat}\phi(\bm{\mathscr h}^{\ell\sharp})$, with the same loss derivatives as
in \eqref{eq:adam-normalized-bulk-gradients}.
Applying \eqref{eq:adam-normalized-state} to these pair histories
defines $\mathcal U_s^\uparrow(\sharp,\flat)$ and
$\mathcal U_s^\downarrow(\flat,\sharp)$.
Let a tilde denote a source-population cavity representative.
The population row and column averages are
\begin{equation}
\label{eq:adam-population-contractions}
\begin{aligned}
 \mathcal A_\uparrow(z)
 &=\sum_{s<t}\mathbb E_{\widetilde\flat}
    [\mathcal U_s^\uparrow(\sharp,\widetilde\flat)\widetilde{\bm{\mathscr h}}^{\ell\flat}(z)],\\
 \mathcal A_\downarrow(z)
 &=\sum_{s<t}\mathbb E_{\widetilde\sharp}
    [\mathcal U_s^\downarrow(\flat,\widetilde\sharp)
                                       \phi(\widetilde{\bm{\mathscr h}}^{\ell\sharp}(z))],\\
 \mathcal B_{\downarrow,q}(z)
 &=\sum_{s<t}\mathbb E_{\widetilde\flat}
    [\mathcal U_s^\downarrow(\widetilde\flat,\sharp)
                                       \widetilde{\bm{\mathscr g}}_{q}^{(\ell+1)\flat}(z)],\\
 \mathcal B_{\uparrow,q}(z)
 &=\sum_{s<t}\mathbb E_{\widetilde\sharp}
    [\mathcal U_s^\uparrow(\widetilde\sharp,\flat)
                                       \widetilde{\bm{\mathscr g}}_{q}^{\ell\sharp}(z)].
\end{aligned}
\end{equation}
Each average holds the target-coordinate history fixed and averages
the source histories with their cavity law. It is the limit of the
corresponding empirical row or column average, rather than an
independence assertion about trained coordinates.
The training terms for $\bm{\mathscr h}^{\ell\sharp}$, $\bm{\mathscr h}^{(\ell+1)\flat}-\bm{\mathscr h}^{\ell\flat}$,
the bracket multiplying $\phi'(\bm{\mathscr h}^{\ell\sharp})$ in $\bm{\mathscr g}_{q}^{\ell\sharp}$,
and $\bm{\mathscr g}_{q}^{\ell\flat}-\bm{\mathscr g}_{q}^{(\ell+1)\flat}$ become, respectively,
\begin{equation}
\label{eq:adam-population-training-terms}
 -\rho_\uparrow\mathcal A_\uparrow,\qquad
 -\frac{\beta_0\rho_\downarrow}{L}\mathcal A_\downarrow,\qquad
 -\rho_\downarrow\mathcal B_{\downarrow,q},\qquad
 -\frac{\beta_0\rho_\uparrow}{L}\mathcal B_{\uparrow,q}.
\end{equation}
The Gaussian integral over the initial weights has the same form as
\eqref{eq:sgd-pi-weight-average}, with the same source and reaction
prefactors. The optimizer dependence resides in the paired moment
histories: their population contractions depend on the full
representative-history law, not just its four second-moment kernels.
Kernels and responses are evaluated self-consistently from this
augmented Adam law.
In continuous depth the residual factors $1/L$ become $d\tau$.
Unlike the spectral history contractions, these pair averages are
generally nonlinear functions of the remaining residual history.
Gaussian initial sources therefore need not give Gaussian residual
statistics after Adam updates.

\paragraph{Trainable endpoints.}
For the input and output maps, use the same moment recursion with
$d_{\rm in}=1/(\gamma_0N_\flat\sqrt{D_{\mathrm i}})$,
$d_{\rm out}=1/(\gamma_0N_\flat)$, and
$\eta_j^{\rm W}=\rho_j$. Their normalized gradients are
\begin{equation}
\label{eq:adam-normalized-endpoint-gradients}
\begin{aligned}
 (R_t^{\rm in})_{ik}
 &=\frac1{B_t}\sum_{(\bm x,y)\in\mathcal B_t}\sum_q
       \partial_{f_q}\mathcal L_{\rm ex}\,
       g_{q,i}^{0\flat}(\bm x,t)x_k,\\
 (R_t^{\rm out})_{qi}
 &=\frac1{B_t}\sum_{(\bm x,y)\in\mathcal B_t}
       \partial_{f_q}\mathcal L_{\rm ex}\,h_i^{L\flat}(\bm x,t).
\end{aligned}
\end{equation}
The resulting boundary histories are
\begin{equation}
\label{eq:adam-endpoint-histories}
\begin{aligned}
 \bm h^{0\flat}(\bm x,t)
 &=\frac{\bm W^0(0)\bm x}{\sqrt{D_{\mathrm i}}}
   -\frac{\rho_{\rm in}}{\sqrt{D_{\mathrm i}}}
       \sum_{s<t}\mathcal U_s^{\rm in}\bm x,\\
 \bm g_q^{L\flat}(t)
 &=(\bm W^L(0))^{\mathsf T}\bm e_q
   -\rho_{\rm out}\sum_{s<t}
       (\mathcal U_s^{\rm out})^{\mathsf T}\bm e_q .
\end{aligned}
\end{equation}
For a representative residual coordinate, the input row and output
column retain their finitely many task-coordinate moment states.
These histories replace the boundary learning terms in
\eqref{eq:sgd-depth-boundaries}; its readout identity is unchanged.

\paragraph{Finite-network histories.}
The coefficients in \eqref{eq:adam-population-training-terms} follow
directly from the finite matrix updates.
Accumulating the update matrices gives the exact forward histories
\begin{equation}
\label{eq:adam-finite-forward-history}
\begin{aligned}
 \bm h^{\ell\sharp}(z)
 &=\frac{\bm W^{\ell\uparrow}(0)\bm h^{\ell\flat}(z)}{\sqrt {N_\flat}}
   -\frac{\rho_\uparrow}{{N_\flat}}
           \sum_{s<t}\mathcal U_s^{\ell\uparrow}\bm h^{\ell\flat}(z),\\
 \bm h^{(\ell+1)\flat}(z)
 &=\bm h^{\ell\flat}(z)
   +\frac{\beta_0\sqrt {N_\flat}}{K}\bm W^{\ell\downarrow}(0)
           \phi(\bm h^{\ell\sharp}(z))\\
 &\quad-\frac{\beta_0\rho_\downarrow}{K}
           \sum_{s<t}\mathcal U_s^{\ell\downarrow}\phi(\bm h^{\ell\sharp}(z)).
\end{aligned}
\end{equation}
All query fields are evaluated in the trained network at $z=(\bm x,t)$.
The backward histories follow from the transposed update matrices:
\begin{equation}
\label{eq:adam-finite-backward-history}
\begin{aligned}
 \bm g_q^{\ell\sharp}(z)
 &=\phi'(\bm h^{\ell\sharp}(z))\odot
   \left[\bm\xi_q^{\ell\sharp}(z)
      -\frac{\rho_\downarrow}{{N_\flat}}\sum_{s<t}
        (\mathcal U_s^{\ell\downarrow})^{\mathsf T}
                         \bm g_q^{(\ell+1)\flat}(z)\right],\\
 \bm g_q^{\ell\flat}(z)-\bm g_q^{(\ell+1)\flat}(z)
 &=\bm\xi_q^{\ell\flat}(z)
      -\frac{\beta_0\rho_\uparrow}{K}\sum_{s<t}
        (\mathcal U_s^{\ell\uparrow})^{\mathsf T}
                         \bm g_q^{\ell\sharp}(z).
\end{aligned}
\end{equation}

\par
\endgroup

\subsection{Joint Gaussianity of residual histories}
\label{app:gaussian-residuals}

\begingroup
\color{black}
SGD and orthogonally equivariant spectral updates preserve a common
rotation of the entire residual history.
When mixed inner products concentrate at deterministic limits, any fixed
finite collection of forward residual states and output Jacobians has a
joint Gaussian limit across inputs, depths, output components, and
training steps, even when hidden variables are non-Gaussian.
Adam can break this property after one update while retaining
self-averaging statistics.
The population equations determine the mixed inner products; the rotation
argument below identifies the joint law.
Numerical illustrations then examine these distributions after learning
in a nonlinear logic task and a complete GPT-2-style training run.
\par
\endgroup

\subsubsection{One rotation acts on the entire training history}

\begingroup
\color{black}
Fix an orthogonal matrix $R\in O(N_\flat)$.
Apply the same transformation to every layer:
\begin{equation}
\label{eq:optimizer-residual-rotation}
\begin{aligned}
\bm W^0&\mapsto R\bm W^0,&
\bm W^{\ell\uparrow}&\mapsto \bm W^{\ell\uparrow}R^{\mathsf T},\\
\bm W^{\ell\downarrow}&\mapsto R\bm W^{\ell\downarrow},&
\bm W^L&\mapsto \bm W^LR^{\mathsf T}.
\end{aligned}
\end{equation}
For the MLP residual network, all hidden activations and outputs remain
unchanged, while
$\bm h^{\ell\flat}\mapsto R\bm h^{\ell\flat}$ and
$\bm g_q^{\ell\flat}\mapsto R\bm g_q^{\ell\flat}$.
In particular, an arbitrary hidden nonlinearity does not obstruct this
symmetry: its input is unchanged. The identity skip connections require
the same $R$ at every depth.

Fix the data and batch sequence, independently of the residual coordinate
system, and let the loss depend on the network through its output.
The induced parameter transformation is orthogonal in the Frobenius
inner product. Invariance of the loss therefore implies equivariance of
its gradients. SGD with a scalar rate for each parameter group preserves
\eqref{eq:optimizer-residual-rotation}, even when the rates differ between
groups or change with training step.

The singular-value definition of $F_\lambda$ gives
\begin{equation}
\label{eq:spectral-equivariance}
 F_\lambda(QDP^{\mathsf T})
 =QF_\lambda(D)P^{\mathsf T}
\end{equation}
for compatible orthogonal $Q,P$.
Thus the spectral update preserves the same residual rotation.
The same identity holds for odd spectral polynomials and their finite
compositions. Linear momentum also preserves it when the momentum state
is transformed with the gradient. A threshold or scalar multiplier
chosen from rotation-invariant quantities preserves equivariance.
None of these statements requires the update to be linear or homogeneous
as a function of the gradient.

Independent centered Gaussian weight matrices, with a common variance
within each group, have an initialization law invariant under
\eqref{eq:optimizer-residual-rotation}.
Induction over training steps consequently gives a common rotational
invariance of the whole residual history.
The result applies when all four parameter groups use compatible
updates; a hybrid method that applies coordinatewise Adam to the
input or readout does not follow from this argument.
\par
\endgroup

\begingroup
\color{black}
Residual-coordinate operations, such as learned diagonal normalization
gains, require a separate symmetry check.
\par
\endgroup

\subsubsection{From deterministic mixed inner products to a joint Gaussian law}

\begingroup
\color{black}
Collect $r$ residual vectors from any fixed finite set of the labels
above into an ${N_\flat}\times r$ matrix $Y_{N_\flat}$, and define
\begin{equation}
\label{eq:history-gram}
        \Gamma_{N_\flat}=\frac1{N_\flat}Y_{N_\flat}^{\mathsf T}Y_{N_\flat}.
\end{equation}
Suppose that $\Gamma_{N_\flat}$ converges in probability to a deterministic,
finite, positive semidefinite matrix $\Gamma$.
Then, for every fixed $k$, the first $k$ rows of $Y_{N_\flat}$ converge jointly
to $k$ independent $\mathcal N(0,\Gamma)$ vectors.

To see this, conditional on $\Gamma_{N_\flat}$, rotational invariance makes
$Y_{N_\flat}$ uniform on the orthogonal orbit with that Gram matrix.
Let $Z_{N_\flat}$ be an independent ${N_\flat}\times r$ standard Gaussian matrix.
For ${N_\flat}\ge r$, a representation of this conditional law is
\begin{equation}
\label{eq:history-gaussian-representation}
Y_{N_\flat}\ \overset{d}{=}\
 Z_{N_\flat}(Z_{N_\flat}^{\mathsf T}Z_{N_\flat})^{-1/2}({N_\flat}\Gamma_{N_\flat})^{1/2}
 =
 Z_{N_\flat}\left(\frac{Z_{N_\flat}^{\mathsf T}Z_{N_\flat}}{{N_\flat}}\right)^{-1/2}
 \Gamma_{N_\flat}^{1/2}.
\end{equation}
Only $Z_{N_\flat}^{\mathsf T}Z_{N_\flat}$ is inverted, so $\Gamma_{N_\flat}$ may be singular.
The middle factor converges in probability to $I_r$, and
$\Gamma_{N_\flat}^{1/2}\to\Gamma^{1/2}$.
The first $k$ rows of $Z_{N_\flat}$ are independent standard Gaussian vectors.
Slutsky's theorem proves the stated joint limit.

The columns of $Y_{N_\flat}$ include the complete finite selection of histories
from the beginning. Thus the proof establishes Gaussianity jointly
across training steps, inputs, depth, and forward/backward variables.
It does not add separately Gaussian terms and infer that their sum is
Gaussian. Different entries within a row can be arbitrarily correlated;
their correlations are the off-diagonal entries of $\Gamma$.
The Gaussian matrix in \eqref{eq:history-gaussian-representation} is a
distributional representation, not noise injected during training.
\par
\endgroup

\begingroup
\color{black}
A random limiting Gram matrix would instead give a Gaussian mixture.
The deterministic-Gram condition is therefore substantive.
The argument identifies finite-dimensional distributions; process
convergence and growing training durations additionally require
tightness and uniform estimates.
\par
\endgroup

\subsubsection{Adam can destroy residual Gaussianity in one step}
\label{app:adam-counterexample}

\begingroup
\color{black}
Consider the zero-block network with scalar input $1$, scalar target $1$,
and two trainable linear maps:
\begin{equation}
\label{eq:adam-linear-example}
 f_{N_\flat}=\frac1{N_\flat}\sum_{i=1}^{N_\flat} a_iw_i,\qquad
 \mathcal L=\frac12(f_{N_\flat}-1)^2,\qquad
 a_i(0),w_i(0)\overset{\mathrm{iid}}{\sim}\mathcal N(0,1).
\end{equation}
In the standard-Gaussian coordinates used here,
$h_i^\flat=w_i$ and the normalized output gradient is $g_{1,i}^\flat=a_i$.
Adam starts from zero moment estimates, uses bias correction, learning
rate $\eta>0$, and stabilizer $\epsilon_{N_\flat}=e/{N_\flat}$ with fixed $e\ge0$.
Regardless of its fixed momentum coefficients in $[0,1)$, the first
update is the gradient divided by its absolute value plus $\epsilon_{N_\flat}$.
Since $f_{N_\flat}(0)\to0$, every fixed set of coordinates has the limit
\begin{equation}
\label{eq:adam-first-update}
\begin{aligned}
 w_i(1)&=w_i(0)+\eta\frac{a_i(0)}{|a_i(0)|+e},\\
 a_i(1)&=a_i(0)+\eta\frac{w_i(0)}{|w_i(0)|+e}.
\end{aligned}
\end{equation}
The common factor $f_{N_\flat}(0)-1$ is included in the finite update before
taking this limit; it is not replaced separately at every coordinate.

When $e=0$, either marginal in \eqref{eq:adam-first-update} is
\begin{equation}
\label{eq:adam-gaussian-mixture}
       \tfrac12\mathcal N(\eta,1)+\tfrac12\mathcal N(-\eta,1).
\end{equation}
Its variance is $1+\eta^2$ and its fourth cumulant is $-2\eta^4$.
For $e>0$, the added term is an independent, bounded, nonconstant
random variable. If its sum with the independent standard Gaussian
were Gaussian, division of their characteristic functions would make
this added term Gaussian as well, which is impossible.
Thus a positive stabilizer does not restore Gaussianity in this scaling.
The residual marginal itself fails to be Gaussian, so joint
Gaussianity fails too.

This example still has self-averaging population statistics.
At $e=0$, for instance,
\begin{equation}
 \frac1{N_\flat}\sum_i w_i(1)^2\longrightarrow1+\eta^2,\qquad
 f_{N_\flat}(1)\longrightarrow2\eta\sqrt{\frac2\pi}.
\end{equation}
The common initial output tends to zero and the limiting coordinate
pairs are functions of independent initial pairs.
Non-Gaussianity here is not caused by a random limiting covariance.
\par
\endgroup

\begingroup
\color{black}
To retain this update at large width, $\epsilon$ must follow the
gradient scale. A fixed positive standard-coordinate $\epsilon$
would make the first update vanish.
In code coordinates, the conversion rules give rate and stabilizer $\eta/{N_\flat}$ and $e$
for the output code weight $a_i/{N_\flat}$; for the input code weight
$w_i$, they remain $\eta$ and $e/{N_\flat}$.
\par
\endgroup

\subsubsection{Numerical illustrations after learning}
\label{app:gaussian-numerics}

\begingroup
\color{black}
After learning AND, OR, and XOR, the nonlinear logic network displays
approximately Gaussian joint clouds of residual activations and actual
output Jacobians under SGD and the calibrated Taylor spectral update,
whereas Adam produces a visibly non-Gaussian distribution
(Fig.~\ref{fig:logic-actual-backward}).
The Jacobians include both the nonlinear branch and the skip path.
Protocols and quantitative diagnostics are given in
Appendix~\ref{app:l01-logic}.
\par
\endgroup

\begin{figure}[!ht]
\centering
\color{black}
\includegraphics{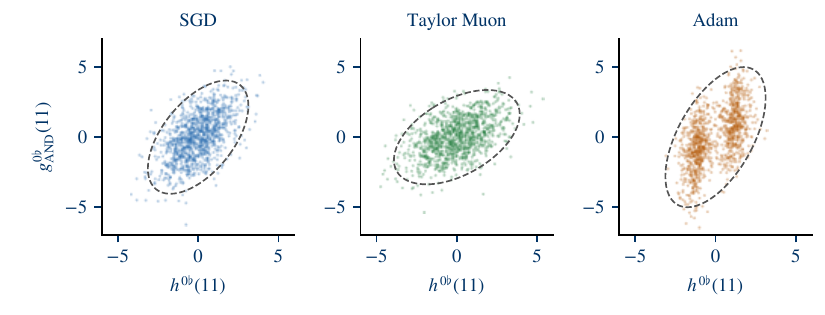}
\caption{Residual activation--Jacobian pairs after learning the nonlinear
logic task.
Each axis is centered, with its original scale retained. Ellipses enclose
95\% of the fitted Gaussian reference; no whitening is applied.}
\label{fig:logic-actual-backward}
\end{figure}

\begingroup
\color{black}
Approximately Gaussian residual activations and output Jacobians also
persist through a complete GPT-2-style run using practical NS5 Muon
on every matrix (Fig.~\ref{fig:ns5-diagnostics}).
At an intermediate checkpoint, these residual shapes coexist with
strongly non-Gaussian first-block pre-GELU activations
(Fig.~\ref{fig:gpt2-theory-illustration}, right).
Training and measurement details are given in
Appendix~\ref{app:gpt2-ns5}.
\par
\endgroup

\begin{figure}[t]
\centering
\color{black}
\includegraphics{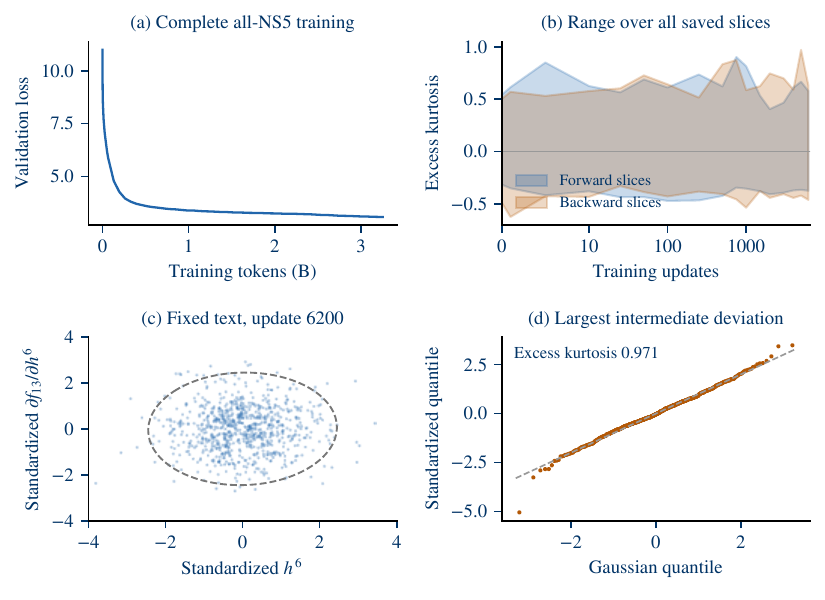}
\caption{All-NS5 training and distributional diagnostics.
Top: complete validation trajectory and excess-kurtosis ranges over
the predefined slices at each measurement time; the bands are ranges,
not confidence intervals. Bottom: the final forward/backward pair
for the first predefined text, with a fitted Gaussian 95\% ellipse,
and quantiles for the retained backward slice with the largest
intermediate excess kurtosis. Each plotted vector is standardized
separately, without whitening.}
\label{fig:ns5-diagnostics}
\end{figure}

\newpage
\section{Zero- and one-step checkpoints}
\label{app:checkpoints}
\label{app:single-linear}

\subsection{Initialization and the intrinsic ratio}
\label{app:sgd-initialization}

\begingroup
\color{black}
Before training, the current block's matrices are independent of its
incoming residual states. For a finite input collection, let
$C_\ell(\bm x,\bm x')=
{N_\flat}^{-1}(\bm h^{\ell\flat}(\bm x,0))^{\mathsf T}
             \bm h^{\ell\flat}(\bm x',0)$.
Each up-projection row produces a Gaussian input vector of covariance
$C_\ell$. Define the nonlinearity's second-moment map by
\begin{equation}
\label{eq:sgd-nonlinearity-covariance}
 [\mathcal V_\phi(C)]_{\bm x,\bm x'}
   =\mathbb E[\phi(Z_{\bm x})\phi(Z_{\bm x'})],
   \qquad Z\sim\mathcal N(0,C).
\end{equation}
If $\delta h_i^{\ell\flat}(\bm x)=
h_i^{(\ell+1)\flat}(\bm x,0)-h_i^{\ell\flat}(\bm x,0)$,
averaging over the independent current up- and down-projections gives
\begin{equation}
\label{eq:sgd-initial-increment-covariance}
\begin{aligned}
 \mathbb E[\delta h_i^{\ell\flat}(\bm x)\mid\bm h^{\ell\flat}]&=0,\\
 \mathbb E[\delta h_i^{\ell\flat}(\bm x)\delta h_j^{\ell\flat}(\bm x')
                         \mid\bm h^{\ell\flat}]
 &=\delta_{ij}\frac{\beta_0^2({N_\flat}/K)}{L}
                  [\mathcal V_\phi(C_\ell)]_{\bm x,\bm x'},\\
 \mathbb E[C_{\ell+1}\mid\bm h^{\ell\flat}]
 &=C_\ell+\frac{\beta_0^2({N_\flat}/K)}{L}\mathcal V_\phi(C_\ell).
\end{aligned}
\end{equation}
These finite-size identities retain the dependence of each block's
conditional variance on its incoming state.
\par
\endgroup

\begingroup
\color{black}
When empirical covariances concentrate, moments remain controlled, and
the depth sums converge, the initialization covariance satisfies
\begin{equation}
\label{eq:sgd-initialization-continuum}
 \frac{dC(\tau)}{d\tau}
   =\beta_0^2\alpha\,\mathcal V_\phi(C(\tau)),
 \qquad
 C(0)_{\bm x,\bm x'}
   =\frac{\bm x^{\mathsf T}\bm x'}{D_{\mathrm i}}.
\end{equation}
The limiting residual process is centered Gaussian with independent
increments over disjoint depth intervals. Its initial value is independent
of these increments, and its covariance across both depth and inputs is
\begin{equation}
\label{eq:sgd-initialization-depth-covariance}
\begin{aligned}
\mathbb E\!\left[
\bm{\mathscr h}^{\flat}(\tau;\bm x,0)
\bm{\mathscr h}^{\flat}(s;\bm x',0)\right]
&=C(\min\{\tau,s\})_{\bm x,\bm x'},\\
\mathbb E[du(\tau;\bm x,0)\,du(\tau;\bm x',0)]
&=c\,[\mathcal V_\phi(C(\tau))]_{\bm x,\bm x'}\,d\tau,
\qquad c=\beta_0^2\alpha.
\end{aligned}
\end{equation}
At $t=0$, the learning histories and forward response kernels
$A^\flat,A^\sharp$ vanish: no parameter update has yet coupled the
forward fields to backward sources.
Equation~\eqref{eq:sgd-dmft-continuous} therefore reduces to
$d\bm{\mathscr h}^{\flat}=du$, with the covariance above.
Thus the deterministic covariance ODE describes the law of a random
residual trajectory, not a deterministic trajectory.

For a fixed input, the diagonal $C_h(\tau;0)=C(\tau)_{\bm x,\bm x}$
satisfies
$\partial_\tau C_h=c\,\mathbb E[\phi(\sqrt{C_h}Z)^2]$,
where $Z\sim\mathcal N(0,1)$.
Unit pre-RMSNorm instead normalizes the branch input before the
up-projection, leaving the residual stream itself unnormalized.
With no gain and with the normalization epsilon set to zero, it replaces
$C(\tau)$ in $\mathcal V_\phi$ by the correlation matrix
$R(\tau)_{\bm x,\bm x'}=
C(\tau)_{\bm x,\bm x'}/
\sqrt{C(\tau)_{\bm x,\bm x}C(\tau)_{\bm x',\bm x'}}$.
Consequently $\partial_\tau C_h=c\,\mathbb E[\phi(Z)^2]$ is constant.
With a nonzero normalization epsilon $\varepsilon$, the diagonal
intensity is instead
$c\,\mathbb E[\phi(\sqrt{C_h/(C_h+\varepsilon)}Z)^2]$.
For ReLU, the zero-epsilon solutions are
\begin{equation}
\label{eq:sgd-initialization-relu-solutions}
C_h(\tau;0)=
\begin{cases}
C_h(0;0)e^{c\tau/2},&\text{without RMSNorm},\\
C_h(0;0)+c\tau/2,&\text{with unit pre-RMSNorm}.
\end{cases}
\end{equation}
For $\phi(z)=z$, the factor $1/2$ is replaced by $1$ in both solutions;
in particular a normalized input without RMSNorm gives
$C_h(\tau;0)=e^{c\tau}$.
At fixed $\beta_0$, the random depth contribution disappears as
$\alpha\to0$.

\par
\endgroup

{\color{black}
\subsection{Effective step and the linear checkpoint}
\label{app:effective-step}

\subsubsection{Setting and parameter coordinates}
\label{app:linear-setting}
As above, $K=N_\sharp L$.
There is one deterministic input
$\bm x\in\mathbb R^{D_{\mathrm i}}$ with
$\|\bm x\|^2/D_{\mathrm i}=1$ and one scalar target $y=1$.
At initialization,
\begin{equation}
\label{eq:linear-finite-model}
\begin{aligned}
\bm h^{0\flat}&=\bm W^0\bm x/\sqrt{D_{\mathrm i}},&
\bm A_\ell&=I+\frac{\beta_0}{K}\bm W^{\ell\downarrow}\bm W^{\ell\uparrow},\\
\bm h^{(\ell+1)\flat}&=\bm A_\ell\bm h^{\ell\flat},&
f_0&=\frac{\bm W^L\bm h^{L\flat}}{\gamma_0 {N_\flat}}.
\end{aligned}
\end{equation}
Here $\bm W^{\ell\uparrow}\in\mathbb R^{{N_\sharp}\times {N_\flat}}$,
$\bm W^{\ell\downarrow}\in\mathbb R^{{N_\flat}\times {N_\sharp}}$,
and $\bm W^L\in\mathbb R^{1\times {N_\flat}}$.
All entries are independent standard Gaussians.
In particular, $\bm h^{0\flat}$ is a standard Gaussian vector independent
of the residual weights and $(\bm W^L)^{\mathsf T}$. The input and readout remain frozen.
All gradients below are evaluated at this same initial network.

{\color{black}
For fixed $\beta_0,\gamma_0>0$, write the SGD rate of each residual
projection in standard-Gaussian coordinates as
\begin{equation}
\label{eq:linear-lr}
\eta_{\mathrm W}=\eta_0\gamma_0^2K
  =\frac{\eta_{\rm eff}\gamma_0K}{\beta_0},
\qquad
\eta_{\rm eff}=\eta_0\gamma_0\beta_0.
\end{equation}
The coefficient $\eta_{\rm eff}$ controls the direct local
hidden-feature update. Both residual projections are updated
simultaneously below, so their joint effect and its propagation
through the network are retained.}
If the scale factors are instead absorbed into the stored matrices,
\begin{equation}
\begin{aligned}
\texttt{Wup}^{\ell}&=\bm W^{\ell\uparrow}/\sqrt {N_\flat},&
\texttt{Wdown}^{\ell}&=\beta_0\sqrt {N_\flat}\,\bm W^{\ell\downarrow}/K ,
\end{aligned}
\end{equation}
the equivalent SGD rates are
\begin{equation}
\label{eq:linear-code-lr}
\eta_{\texttt{Wup}}
=\eta_{\rm eff}\gamma_0K/(\beta_0 {N_\flat}),
\qquad
\eta_{\texttt{Wdown}}
=\eta_{\rm eff}\gamma_0\beta_0 {N_\flat}/K.
\end{equation}
These rates follow from the displayed up- and down-projection factors.
Both endpoints remain fixed.

\subsubsection{The exact finite update}
\label{app:linear-exact}
Define the backward vectors
\begin{equation}
\bm g_1^{L\flat}=(\bm W^L)^{\mathsf T},\qquad
\bm g_1^{\ell\flat}=\bm A_\ell^\top\bm g_1^{(\ell+1)\flat},
\qquad
\delta=1-f_0.
\end{equation}
For the loss $\mathcal L=\frac12(f_0-1)^2$, direct differentiation gives
\begin{equation}
\label{eq:linear-gradients}
\begin{aligned}
\nabla_{\bm W^{\ell\uparrow}}\mathcal L
&=-\frac{\delta\beta_0}{\gamma_0 {N_\flat}K}
   ((\bm W^{\ell\downarrow})^\top\bm g_1^{(\ell+1)\flat})(\bm h^{\ell\flat})^\top,\\
\nabla_{\bm W^{\ell\downarrow}}\mathcal L
&=-\frac{\delta\beta_0}{\gamma_0 {N_\flat}K}
   \bm g_1^{(\ell+1)\flat}(\bm W^{\ell\uparrow}\bm h^{\ell\flat})^\top.
\end{aligned}
\end{equation}
The simultaneous parameter update is
\begin{equation}
\label{eq:linear-exact-update}
\begin{aligned}
\Delta\bm W^{\ell\uparrow}
&=\frac{\eta_{\rm eff}\delta}{{N_\flat}}
   ((\bm W^{\ell\downarrow})^\top\bm g_1^{(\ell+1)\flat})(\bm h^{\ell\flat})^\top,\\
\Delta\bm W^{\ell\downarrow}
&=\frac{\eta_{\rm eff}\delta}{{N_\flat}}
   \bm g_1^{(\ell+1)\flat}(\bm W^{\ell\uparrow}\bm h^{\ell\flat})^\top,\\
\bm A_\ell^+
&=\bm A_\ell+\frac{\beta_0}{K}
  \left(\Delta\bm W^{\ell\downarrow}\bm W^{\ell\uparrow}+
        \bm W^{\ell\downarrow}\Delta\bm W^{\ell\uparrow}+
        \Delta\bm W^{\ell\downarrow}\Delta\bm W^{\ell\uparrow}\right),\\
(\bm h^{L\flat})^+&=\bm A_{L-1}^+\cdots\bm A_0^+\bm h^{0\flat} .
\end{aligned}
\end{equation}
In particular, a finite SGD step includes the last, quadratic term
and products of updates at different layers. Both are retained in
the finite-network simulations.

For comparison, let $\bm z_0=0$ and define the linear response with
the output error set to its limiting value $\delta=1$:
\begin{equation}
\label{eq:linear-tangent}
\begin{aligned}
\bm z_{\ell+1}&=\bm A_\ell\bm z_\ell+
 \frac{\eta_{\rm eff}\beta_0}{{N_\flat}K}
 \left[
 \|\bm W^{\ell\uparrow}\bm h^{\ell\flat}\|^2\bm g_1^{(\ell+1)\flat}
 +\|\bm h^{\ell\flat}\|^2\bm W^{\ell\downarrow}(\bm W^{\ell\downarrow})^\top\bm g_1^{(\ell+1)\flat}
 \right].
\end{aligned}
\end{equation}
This auxiliary response is not the definition of the finite training
step. Its relation to \eqref{eq:linear-exact-update} is justified below.

\subsubsection{Initialization and the finite-step limit}
\label{app:linear-limits}
Set $s=1+\beta_0^2N_\flat/(KL)$.
Independence of the current block from its input gives the exact identities
\begin{equation}
\label{eq:linear-init-exact}
\E\frac{\|\bm h^{\ell\flat}\|^2}{{N_\flat}}=s^\ell,
\qquad
\E f_0^2=\frac{s^L}{\gamma_0^2{N_\flat}}.
\end{equation}
Thus $\delta\to1$ as width grows, and the continuum initialization
profile is $\exp(\beta_0^2\alpha\tau)$.
Conditioned on the residual weights and $\bm h^{0\flat}$, the vector
$\bm z_\ell$ is linear in the independent readout $(\bm W^L)^{\mathsf T}$.
It follows that
$\E[(\bm h^{\ell\flat})^\top\bm z_\ell]=0$ exactly.

The limiting statement first sends
${N_\flat},{N_\sharp}\to\infty$ at each fixed $L$, with ${N_\flat}/({N_\sharp}L)\to\alpha>0$,
and then sends $L\to\infty$.
At fixed depth, the finite-step remainder is controlled as follows.
Write $\widehat{\bm W}^{\ell\uparrow}=\bm W^{\ell\uparrow}/\sqrt {N_\flat}$,
$\widehat{\bm W}^{\ell\downarrow}=\bm W^{\ell\downarrow}/\sqrt {N_\flat}$,
$u=\bm h^{0\flat}/\sqrt {N_\flat}$, and $v=(\bm W^L)^{\mathsf T}/\sqrt {N_\flat}$. Their entries are independent
Gaussians of variance $1/{N_\flat}$.
At fixed $L$, the aspect ratio ${N_\sharp}/{N_\flat}$ stays in a compact subset of
$(0,\infty)$, so all fixed-order moments of
$\|\widehat{\bm W}^{\ell\uparrow}\|_{\mathrm{op}},
\|\widehat{\bm W}^{\ell\downarrow}\|_{\mathrm{op}},\|u\|,\|v\|$
are uniformly bounded.

Expand the product in \eqref{eq:linear-exact-update}.
Every term beyond \eqref{eq:linear-tangent} contains at least one
normalized mixed contraction between the independent initial state
and readout, of the form ${N_\flat}^{-1}(\bm h^{0\flat})^\top B(\bm W^L)^{\mathsf T}$,
with the matrix $B$ determined by the residual weights.
Conditioned on those weights and averaging over both $\bm h^{0\flat}$
and $(\bm W^L)^{\mathsf T}$, its second moment is ${N_\flat}^{-2}\tr(BB^\top)$;
its higher even moments have the corresponding Gaussian bounds.
The normalized matrix factors in $B$ have bounded operator-norm moments.
H\"older's inequality therefore controls each higher insertion,
including its dependent remaining factors, in normalized mean square
by $C_L/{N_\flat}$. The number of terms is finite at fixed $L$.
The same argument handles $\delta-1=-f_0$. Consequently,
for $j\leq L$,
\begin{equation}
\label{eq:linear-remainder}
\E\frac1{N_\flat}\|(\bm h^{j\flat})^+-\bm h^{j\flat}-\bm z_j\|^2\leq \frac{C_L}{{N_\flat}}.
\end{equation}
Here and below $C_L$ may depend on
$L,\alpha,\eta_{\rm eff},\beta_0,\gamma_0$, but not on ${N_\flat},{N_\sharp}$ along the
fixed-depth width limit.

To justify self-averaging over the weights as well as over the readout,
apply the Gaussian Poincar\'e inequality to
$\|\bm h^{j\flat}\|^2/{N_\flat}$, $\|\bm z_j\|^2/{N_\flat}$, and
$(\bm h^{j\flat})^\top\bm z_j/{N_\flat}$.
They are finite-degree scalar polynomials in
$\widehat{\bm W}^{\ell\uparrow},\widehat{\bm W}^{\ell\downarrow},u,v$.
Differentiating their vector and matrix products bounds the expected
sum of squared vector gradients and squared Frobenius matrix gradients
by $C_L$, using the norm-moment bounds above.
Each independent Gaussian coordinate has variance $1/{N_\flat}$.
Since the mixed energy has mean zero, this gives
\begin{equation}
\label{eq:linear-concentration}
\operatorname{Var}\!\left(\frac{\|\bm h^{j\flat}\|^2}{{N_\flat}}\right)
+\operatorname{Var}\!\left(\frac{\|\bm z_j\|^2}{{N_\flat}}\right)
+\E\left[\left(\frac{(\bm h^{j\flat})^\top\bm z_j}{{N_\flat}}\right)^2\right]
\leq\frac{C_L}{{N_\flat}}.
\end{equation}
Equations \eqref{eq:linear-remainder} and
\eqref{eq:linear-concentration} identify the limiting finite-step
energy with the initialization energy plus the averaged tangent
energy, in probability.
This argument allows fixed, finite $\eta_{\rm eff},\beta_0,\gamma_0$;
it is not a small-learning-rate expansion.

{\color{black}
The finite-step reduction takes widths to infinity at fixed depth.
The moment recursion below also has a direct joint width--depth limit;
extending the remainder bound to that limit requires control of $C_L$
as depth grows.}

\subsubsection{A closed moment calculation}
\label{app:linear-moments}
The second moment of \eqref{eq:linear-tangent} can be evaluated without
discarding the factorization $\bm W^{\ell\downarrow}\bm W^{\ell\uparrow}$.
Put $\varepsilon=\beta_0/K$, and define
\begin{equation}
\label{eq:linear-M}
\begin{aligned}
q_\ell&=\|\bm h^{\ell\flat}\|^2/{N_\flat},&
r_\ell&=\|\bm W^{\ell\uparrow}\bm h^{\ell\flat}\|^2/{N_\flat},\\
\bm M_0&=0,&
\bm M_{\ell+1}
&=\bm A_\ell\bm M_\ell\bm A_\ell^\top
  +\varepsilon(r_\ell I+q_\ell\bm W^{\ell\downarrow}(\bm W^{\ell\downarrow})^\top).
\end{aligned}
\end{equation}
If $\bm F_j=\bm A_{L-1}\cdots\bm A_j$, with $\bm F_L=I$, then
$\bm z_j=\eta_{\rm eff}\bm M_j\bm F_j^\top(\bm W^L)^{\mathsf T}$.
The past and future are independent, so
\begin{equation}
\label{eq:linear-moment-transfer}
\E\frac{\|\bm z_j\|^2}{{N_\flat}}
=\eta_{\rm eff}^2s^{L-j}S_j,
\qquad
S_j=\E\frac{\tr(\bm M_j^2)}{{N_\flat}}.
\end{equation}

To close its evolution, retain five scalar moments:
\begin{equation}
\label{eq:linear-five-moments}
\begin{aligned}
Q_j&=\E q_j^2,&
P_j&=\E\left[q_j\,\tr(\bm M_j)/{N_\flat}\right],&
T_j&=\E\left[(\tr(\bm M_j)/{N_\flat})^2\right],\\
S_j&=\E\tr(\bm M_j^2)/{N_\flat},&
H_j&=\E[(\bm h^{j\flat})^\top\bm M_j\bm h^{j\flat}/{N_\flat}].
\end{aligned}
\end{equation}
They start from $(Q_0,P_0,T_0,S_0,H_0)=(1+2/{N_\flat},0,0,0,0)$.
The factor $1+2/{N_\flat}$ is required for the Gaussian input embedding.

The five moments obey a closed finite-size recursion. Define
\begin{equation}
\label{eq:linear-moment-coefficients}
\begin{aligned}
a_*&=1+2\varepsilon^2{N_\sharp}({N_\flat}+1)
 +\varepsilon^4{N_\flat}{N_\sharp}[({N_\flat}+1)({N_\sharp}+1)+2],\\
b_*&=2\varepsilon^2{N_\sharp}{N_\flat}+\varepsilon^4{N_\flat}^2{N_\sharp}({N_\flat}+{N_\sharp}+1),\\
d_*&=1+2\varepsilon^2{N_\sharp}{N_\flat}+\varepsilon^4({N_\flat}^2{N_\sharp}^2+2{N_\flat}{N_\sharp}),\\
e_*&=4\varepsilon^2{N_\sharp}/{N_\flat}+2\varepsilon^4{N_\sharp}({N_\flat}+{N_\sharp}+1),\\
g_*&=1+2\varepsilon^2{N_\sharp}({N_\flat}+2)
 +\varepsilon^4{N_\sharp}{N_\flat}({N_\sharp}+2)({N_\flat}+2).
\end{aligned}
\end{equation}
All quantities on the right below are evaluated at index $j$:
\begin{equation}
\label{eq:linear-exact-moments}
\begin{aligned}
Q_{j+1}&=g_*Q_j,\\
P_{j+1}&=d_*P_j+e_*H_j
 +[2\varepsilon {N_\sharp}+2\varepsilon^3{N_\sharp}\{{N_\flat}({N_\sharp}+1)+1\}]Q_j,\\
H_{j+1}&=a_*H_j+b_*P_j
 +[2\varepsilon {N_\sharp}+\varepsilon^3{N_\sharp}{N_\flat}({N_\flat}+2{N_\sharp}+3)]Q_j,\\
T_{j+1}&=d_*T_j+e_*S_j
 +[4\varepsilon {N_\sharp}+4\varepsilon^3{N_\sharp}({N_\flat}{N_\sharp}+1)]P_j\\
 &\quad+4\varepsilon^3N_\sharp H_j
 +\varepsilon^2(4{N_\sharp}^2+2{N_\sharp}+2{N_\sharp}/{N_\flat})Q_j,\\
S_{j+1}&=a_*S_j+b_*T_j
 +[4\varepsilon {N_\sharp}+2\varepsilon^3{N_\sharp}{N_\flat}({N_\flat}+2{N_\sharp}+1)]P_j\\
 &\quad+4\varepsilon^3N_\sharp H_j+\varepsilon^2{N_\sharp}({N_\flat}+4{N_\sharp}+3)Q_j.
\end{aligned}
\end{equation}
These identities follow by conditioning on $\bm h^{j\flat},\bm M_j$
and pairing the independent Gaussian entries of the current $\bm W^{j\uparrow}$
and $\bm W^{j\downarrow}$. For example, for fixed symmetric $M$,
\begin{equation}
\E\frac{\tr[(AMA^\top)^2]}{N_\flat}
=a_*\frac{\tr(M^2)}{N_\flat}+
b_*\left(\frac{\tr M}{N_\flat}\right)^2.
\end{equation}
For one block,
$S_1=\varepsilon^2{N_\sharp}({N_\flat}+4{N_\sharp}+3)(1+2/{N_\flat})$.
The first trace moment provides another check:
\begin{equation}
\label{eq:linear-first-trace}
\E\frac{\tr M_j}{N_\flat}=\frac{2\beta_0 j}{L}s^{j-1}.
\end{equation}

An important fourth-order contribution is visible in $b_*$:
\begin{equation}
L\varepsilon^4{N_\flat}^3{N_\sharp}\longrightarrow\beta_0^4\alpha^3.
\end{equation}
It survives the depth limit. Thus replacing $\bm W^{\ell\downarrow}\bm W^{\ell\uparrow}$ by a matrix with
independent Gaussian entries and the same entrywise covariance would
not preserve this calculation.

\subsubsection{Solving the continuum moments}
\label{app:linear-solution}
Let $c=\beta_0^2\alpha$ and $\tau=j/L$. Taking the limits of
\eqref{eq:linear-exact-moments} yields the closed subsystem
\begin{equation}
\label{eq:linear-moment-ode}
\begin{aligned}
\dot Q&=2cQ,&Q(0)&=1,\\
\dot P&=2cP+2\beta_0Q,&P(0)&=0,\\
\dot T&=2cT+4\beta_0P,&T(0)&=0,\\
\dot S&=2cS+c(2+c\alpha)T
 +2\beta_0(2+c\alpha)P+cQ,&S(0)&=0.
\end{aligned}
\end{equation}
The moment $H$ drops out of this subsystem in the limit.
In fact these deterministic moment equations also follow along
joint sequences with ${N_\flat},L\to\infty$ and $N_\flat/K\to\alpha<\infty$,
without requiring ${N_\sharp}\to\infty$.
The nonnegative finite-size recursions give, by discrete Gr\"onwall,
$\sup_{j\leq L}(Q_j+P_j+T_j+S_j+H_j)\leq C$ for bounded $N_\flat/K$
and fixed $\beta_0$. The omitted coefficients vanish after
multiplication by $L$, and the resulting bounded linear Euler
recursion converges to \eqref{eq:linear-moment-ode}.

The first three solutions are
$Q=e^{2c\tau}$,
$P=2\beta_0\tau e^{2c\tau}$, and
$T=4\beta_0^2\tau^2e^{2c\tau}$.
Substituting them into the last equation gives
\begin{equation}
\frac{d}{d\tau}\left(e^{-2c\tau}S\right)
=c+4\beta_0^2(2+c\alpha)\tau
 +4c\beta_0^2(2+c\alpha)\tau^2,
\end{equation}
and hence
\begin{equation}
\label{eq:linear-S-solution}
S(\tau)=e^{2c\tau}
\left[c\tau+2\beta_0^2(2+c\alpha)\tau^2
 +\frac43c\beta_0^2(2+c\alpha)\tau^3\right].
\end{equation}
Combining \eqref{eq:linear-moment-transfer},
\eqref{eq:linear-remainder}, and
\eqref{eq:linear-S-solution} proves the general form of the
iterated-limit finite-step result:
\begin{equation}
\label{eq:linear-general-beta}
\begin{aligned}
C_h(\tau;0)&=e^{c\tau},\\
C_h(\tau;1)-C_h(\tau;0)
&=\eta_{\rm eff}^2e^{c(1+\tau)}
\left[c\tau+2(2\beta_0^2+c^2)\tau^2
              \left(1+\frac{2c\tau}{3}\right)\right].
\end{aligned}
\end{equation}
Here $\beta_0^2(2+c\alpha)=2\beta_0^2+c^2$, so the expression is
an exact regrouping of the moment solution.
Setting $\beta_0=1$ and $\eta_{\rm eff}=\gamma_0$ gives the
baseline used in the numerical checks:
\begin{equation}
\label{eq:linear-one-step}
\begin{aligned}
C_h(\tau;0)&=e^{\alpha\tau},\\
C_h(\tau;1)&=e^{\alpha\tau}
 +\gamma_0^2e^{\alpha(1+\tau)}
 \left[\alpha\tau+2(\alpha^2+2)\tau^2
 +\frac43(\alpha^3+2\alpha)\tau^3\right].
\end{aligned}
\end{equation}
Its continuous extension to $\alpha=0$ is
$C_h(\tau;0)=1$ and $C_h(\tau;1)=1+4\gamma_0^2\tau^2$.
For general $\beta_0$, replacing $\alpha$ by $\beta_0^2\alpha$
in this baseline alone would miss the separate coupling dependence
of the learning-induced term.

The calculation concerns feature motion, not a guarantee of a decrease
in squared loss. For instance, \eqref{eq:linear-first-trace} gives
$f_1\to2\eta_{\rm eff}\beta_0e^c/\gamma_0$ in the same iterated limit.
With $\beta_0=1$ and $\eta_{\rm eff}=\gamma_0$, this overshoots the target.
Training the input or readout, using multiple samples, or adding a
nonlinearity changes the calculation.

\begingroup
\color{black}
\subsubsection{The same checkpoint from the DMFT}
\label{app:linear-dmft-checkpoint}

The population equations recover the full one-step formula directly.
Keep the normalized-input, linear, frozen-endpoint setting above,
with the same $\eta_{\rm eff}$ for both residual projections.
The input is fixed; the second argument of each representative field is the training step $t=0,1$.
At initialization the forward and backward processes are independent,
with
\begin{equation}
\label{eq:linear-dmft-initial-covariances}
\begin{aligned}
 \mathbb E[\bm{\mathscr h}^\flat(\tau;0)\bm{\mathscr h}^\flat(s;0)]&=e^{c\min(\tau,s)},\\
 \mathbb E[\bm{\mathscr g}_1^\flat(\tau;0)\bm{\mathscr g}_1^\flat(s;0)]&=e^{c(1-\max(\tau,s))}.
\end{aligned}
\end{equation}
Thus the limiting initial output is zero and its loss feedback is one.
The sources retain their correlations between training steps in
\eqref{eq:sgd-population-sources}; different forward/backward source
families remain independent.
Since $\bm{\mathscr g}_1^\sharp(\tau;0)=r^\sharp(\tau;0)$ and $\bm{\mathscr h}^\sharp(\tau;0)=u^\sharp(\tau;0)$, the hidden cross moment
satisfies $\Phi^\sharp_{10}=\Phi^\flat_{10}$.
The residual cross moment then obeys
$\partial_\tau\Phi^\flat_{10}=c\Phi^\flat_{10}$, with initial value
one, hence $\Phi^\flat_{10}=e^{c\tau}$.

Let $A(\tau)=A^\flat(\tau;1,0)$. There is no forward response to
current backward sources. The hidden equation and the residual
equation at step one reduce to
\begin{equation}
\label{eq:linear-dmft-one-step-fields}
\begin{aligned}
 \bm{\mathscr h}^\sharp(\tau;1)&=u^\sharp(\tau;1)+\lambda_h(\tau)\bm{\mathscr g}_1^\sharp(\tau;0),
 &\lambda_h=A^\sharp_{10}
     &=\beta_0\alpha A+\eta_{\rm eff}e^{c\tau},\\
 d\bm{\mathscr h}^\flat(\tau;1)&=du(\tau;1)+k(\tau)\bm{\mathscr g}_1^\flat(\tau;0)\,d\tau,
 &k&=cA+2\beta_0\eta_{\rm eff}e^{c\tau}.
\end{aligned}
\end{equation}
The two direct-learning terms in $k$ come from the up-projection
response and the down-projection learning history.
A unit initial backward-source increment at depth $\tau$ changes
$\bm{\mathscr g}_1^\flat(s;0)$ by one for $s<\tau$, so
$A(\tau)=\int_0^\tau k(s)\,ds$. Consequently
\begin{equation}
\label{eq:linear-dmft-response-solution}
\begin{aligned}
 A(\tau)&=2\beta_0\eta_{\rm eff}\tau e^{c\tau},&
 \lambda_h(\tau)&=\eta_{\rm eff}e^{c\tau}(1+2c\tau).
\end{aligned}
\end{equation}
Source independence gives
$\Phi^\sharp_{11}=C_h(\tau;1)
 +\eta_{\rm eff}^2e^{c(1+\tau)}(1+2c\tau)^2$, whereas
\eqref{eq:linear-dmft-initial-covariances} gives
$\mathbb E[\bm{\mathscr h}^\flat(\tau;1)\bm{\mathscr g}_1^\flat(\tau;0)]
 =e^{c(1-\tau)}A(\tau)=2\beta_0\eta_{\rm eff}\tau e^c$.
Thus the trained fields are correlated even though their source
families are independent. The variance equation is
\begin{equation}
\label{eq:linear-dmft-energy-ode}
\begin{aligned}
 \partial_\tau C_h(\tau;1)
 &=cC_h(\tau;1)+\eta_{\rm eff}^2e^{c(1+\tau)}
    \left[c(1+2c\tau)^2+8\beta_0^2\tau(1+c\tau)\right],\\
 C_h(0;1)&=1.
\end{aligned}
\end{equation}
Integration recovers \eqref{eq:linear-general-beta}.
The frozen readout also gives $f_1=2\beta_0\eta_{\rm eff}e^c/\gamma_0$.
This calculation retains a finite SGD step; it does not expand in
$\eta_{\rm eff}$.
\par
\endgroup

\begingroup
\color{black}
\subsubsection{Coupling and direct local hidden learning}
\label{app:linear-local-learning}

Fix $c>0$, $\gamma_0>0$, and $\tau\in(0,1]$.
Within the limiting family, set $\alpha=c/\beta_0^2$ and allow
the positive step multiplier to depend on $\beta_0$.
Equation~\eqref{eq:linear-general-beta} gives
\begin{equation}
\label{eq:linear-large-beta}
\frac{C_h(\tau;1)-C_h(\tau;0)}
     {(\eta_{\rm eff}\beta_0)^2}
\longrightarrow
4e^{c(1+\tau)}\tau^2\left(1+\frac{2c\tau}{3}\right)
\quad\text{as }\beta_0\to\infty.
\end{equation}
The limit is strictly positive. Keeping the first-step residual energy
bounded therefore requires $\eta_{\rm eff}=O(\beta_0^{-1})$.
The output identity gives the same scale:
$f_1\to2(\eta_{\rm eff}\beta_0/\gamma_0)e^c$.
These are comparisons between limits taken at fixed constants, without
interchanging growing coupling with the width and depth limits.

For a block $\ell$, hold the incoming residual vector $\bm h^{\ell\flat}$
and outgoing backward vector $\bm g_1^{(\ell+1)\flat}$ at their initial values.
The linear hidden variables are
$\bm h^{\ell\sharp}=\bm W^{\ell\uparrow}\bm h^{\ell\flat}/\sqrt {N_\flat}$ and
$\bm g_1^{\ell\sharp}=(\bm W^{\ell\downarrow})^\top\bm g_1^{(\ell+1)\flat}/\sqrt {N_\flat}$.
The parameter increments in \eqref{eq:linear-exact-update} give
the exact local changes
\begin{equation}
\label{eq:linear-local-hidden-update}
\begin{aligned}
\Delta_{\rm loc}\bm h^{\ell\sharp}
&=\frac{\Delta\bm W^{\ell\uparrow}\bm h^{\ell\flat}}{\sqrt {N_\flat}}
 =\eta_{\rm eff}\delta\,\frac{\|\bm h^{\ell\flat}\|^2}{{N_\flat}}
        \bm g_1^{\ell\sharp},\\
\Delta_{\rm loc}\bm g_1^{\ell\sharp}
&=\frac{\Delta(\bm W^{\ell\downarrow})^\top\bm g_1^{(\ell+1)\flat}}{\sqrt {N_\flat}}
 =\eta_{\rm eff}\delta\,\frac{\|\bm g_1^{(\ell+1)\flat}\|^2}{{N_\flat}}
        \bm h^{\ell\sharp}.
\end{aligned}
\end{equation}
At initialization, the current block's matrices are independent of
these neighboring vectors. For $\ell/L\to\tau$, their norm limits are
$\|\bm h^{\ell\flat}\|^2/{N_\flat}\to e^{c\tau}$ and
$\|\bm g_1^{(\ell+1)\flat}\|^2/{N_\flat}\to e^{c(1-\tau)}$.
The backward limit follows by applying the initialization calculation
to the transposed blocks in reverse order.
Since $\delta\to1$, conditional Gaussian averaging over the hidden
coordinates gives
\begin{equation}
\label{eq:linear-local-hidden-energy}
\begin{aligned}
{N_\sharp}^{-1}\|\Delta_{\rm loc}\bm h^{\ell\sharp}\|^2
 &\longrightarrow\eta_{\rm eff}^2e^{c(1+\tau)},\\
{N_\sharp}^{-1}\|\Delta_{\rm loc}\bm g_1^{\ell\sharp}\|^2
 &\longrightarrow\eta_{\rm eff}^2e^{c(2-\tau)}.
\end{aligned}
\end{equation}
Both direct local changes therefore vanish in the bounded-energy
large-$\beta_0$ family.
The full hidden activation can still move through its changing input:
\begin{equation}
\label{eq:linear-hidden-total-change}
\bm h^{\ell\sharp}(1)-\bm h^{\ell\sharp}(0)
=\Delta_{\rm loc}\bm h^{\ell\sharp}+
\frac{(\bm W^{\ell\uparrow})^+[\bm h^{\ell\flat}(1)-\bm h^{\ell\flat}(0)]}{\sqrt {N_\flat}}.
\end{equation}
The second term is not controlled by the vanishing local step alone.
The backward hidden variable likewise receives changes through the
updated downstream backward vector.

The coupling factor is also visible in a nonlinear block.
Holding its incoming residual state fixed, the down-projection update
on one scalar loss contributes
\begin{equation}
\label{eq:nonlinear-down-coupling}
 \left.\Delta h_i^{(\ell+1)\flat}\right|_{\rm down}
 =-\frac{\beta_0\eta_{\rm eff}}{L}
 \frac{\partial\mathcal L}{\partial f}\,
 g_i^{(\ell+1)\flat}
 \frac1{N_\sharp}\sum_{a=1}^{N_\sharp}\phi(h_a^{\ell\sharp})^2.
\end{equation}
The linear checkpoint includes both projection updates and their
propagation through all subsequent blocks.
\par
\endgroup

\begingroup
\color{black}
\subsubsection{Gaussian hidden preactivations along the SGD semi-freeze family}
\label{app:semifreeze-hidden-gaussianity}

The same suppression of local learning also removes the non-Gaussian
deformation of hidden preactivations in the regular SGD semi-freeze
family. Take the population and depth limits at fixed $\beta_0$ first,
then consider
\begin{equation}
\label{eq:semifreeze-sgd-family}
 \alpha=\frac{c}{\beta_0^2},\qquad c>0,\qquad
 \eta_{{\rm eff},\uparrow}=\eta_\uparrow\gamma_0\beta_0
       =O(\beta_0^{-1}),\qquad
 \eta_{{\rm eff},\downarrow}=\eta_\downarrow\gamma_0\beta_0
       =O(\beta_0^{-1}).
\end{equation}
Fix a depth and a finite training horizon $T$. Let $\mathcal X$
contain all training and probe inputs, and set
$\mathcal Z=\mathcal X\times\{0,\ldots,T\}$.
Work on a regular training window where the deterministic
residual kernels, averaged responses, and loss feedback remain
bounded along this family. The endpoint learning rates may remain
fixed.

Stack the hidden fields on this collection into $\bm{\mathscr h}_\beta^\sharp,\bm{\mathscr g}_\beta^\sharp$.
Equation~\eqref{eq:sgd-dmft-hidden} has the form
\begin{equation}
\label{eq:semifreeze-hidden-operators}
\begin{aligned}
 \bm{\mathscr h}_\beta^\sharp&=\bm u_\beta+\mathsf E_\beta\bm{\mathscr g}_\beta^\sharp,
 &\mathsf E_\beta
   &=\frac{c}{\beta_0}\mathsf A_\beta
        +\eta_{{\rm eff},\uparrow}\mathsf S_{\Phi,\beta},\\
 \bm{\mathscr g}_\beta^\sharp
   &=D_\phi(\bm{\mathscr h}_\beta^\sharp)
         [\bm r_\beta+\mathsf F_\beta\phi(\bm{\mathscr h}_\beta^\sharp)],
 &\mathsf F_\beta
   &=\frac{c}{\beta_0}\mathsf B_\beta
        +\eta_{{\rm eff},\downarrow}\mathsf S_{G,\beta}.
\end{aligned}
\end{equation}
Here $\mathsf A_\beta,\mathsf B_\beta$ collect the residual responses,
$\mathsf S_{\Phi,\beta},\mathsf S_{G,\beta}$ are the loss-weighted
history maps, and $D_\phi$ is diagonal with the activation derivatives,
repeated over output components.
The centered Gaussian sources $\bm u_\beta,\bm r_\beta$ have the
residual covariances in \eqref{eq:sgd-population-sources}.
In particular, the reaction term
$(c/\beta_0)\mathsf A_\beta\bm{\mathscr g}_\beta^\sharp$ accounts for dependence
between the initial up weights and their trained residual inputs.
Freezing the up weights alone would not justify discarding that term.

Write $e_\beta=\|\mathsf E_\beta\|_{\rm op}$,
$f_\beta=\|\mathsf F_\beta\|_{\rm op}$,
$u_*=\|\bm u_\beta\|_{L^2}$, and
$r_*=\|\bm r_\beta\|_{L^2}$.
The regular-window bounds give $e_\beta+f_\beta=O_T(\beta_0^{-1})$
and $u_*+r_*=O_T(1)$.
For an activation with bounded derivative $M=\|\phi'\|_\infty$,
including ReLU, put $b=\|\phi(\bm 0)\|$.
Pointwise, $\|\bm{\mathscr h}_\beta^\sharp\|\le\|\bm u_\beta\|+e_\beta\|\bm{\mathscr g}_\beta^\sharp\|$
and
$\|\bm{\mathscr g}_\beta^\sharp\|\le
M[\|\bm r_\beta\|+f_\beta(b+M\|\bm{\mathscr h}_\beta^\sharp\|)]$.
Solving these inequalities before taking $L^2$ norms gives,
for sufficiently large $\beta_0$,
\begin{equation}
\label{eq:semifreeze-hidden-moment-bound}
 \|\bm{\mathscr g}_\beta^\sharp\|_{L^2}
 \le
 \frac{M\{r_*+f_\beta(b+Mu_*)\}}
      {1-M^2e_\beta f_\beta}
 =O_T(1).
\end{equation}
This controls the hidden field multiplying the small coefficients
without an independence assumption.
The fields and sources already share a probability space, so
\begin{equation}
\label{eq:semifreeze-hidden-gaussian-bound}
\begin{aligned}
 W_2\!\left(\mathcal L(\bm{\mathscr h}_\beta^\sharp),
                \mathcal N(0,\Sigma_\beta)\right)
 &\le \|\bm{\mathscr h}_\beta^\sharp-\bm u_\beta\|_{L^2}
 \le e_\beta\|\bm{\mathscr g}_\beta^\sharp\|_{L^2}
 =O_T(\beta_0^{-1}),\\
 (\Sigma_\beta)_{zz'}&=\Phi^\flat_\beta(z,z').
\end{aligned}
\end{equation}
Here $W_2$ is the quadratic Wasserstein distance.
If $\Sigma_\beta$ converges, this gives a joint Gaussian limit over
the selected inputs and training steps. Its covariance may change
by order one during training; the preactivations need not remain
at their initial values.
The pointwise bound also controls every fixed $L^p$ norm, so
fixed-order moments approach those of the Gaussian source.

\paragraph{Pre-RMS normalization.}
For a residual branch using
$\mathcal N(\bm h)=\bm h/\nu_{N_\flat}(\bm h)$ with
$\nu_{N_\flat}(\bm h)^2=\|\bm h\|^2/{N_\flat}+\epsilon$ and unit gain,
the source covariance and the direct-learning history kernel are both
\begin{equation}
\label{eq:semifreeze-rms-kernel}
 \Phi_\beta^{\mathcal N}(z,z')
 =\frac{\Phi^\flat_\beta(z,z')}
        {\nu_\beta(z)\nu_\beta(z')},\qquad
 \nu_\beta(z)^2=\Phi^\flat_\beta(z,z)+\epsilon.
\end{equation}
The reaction also changes, because backpropagation through the up map
$\bm W^{\ell\uparrow}$ contains the full normalization Jacobian:
\begin{equation}
\label{eq:semifreeze-rms-jacobian}
\begin{aligned}
 D\mathcal N(\bm h)
 &=\frac{I}{\nu_{N_\flat}(\bm h)}
   -\frac{\bm h\bm h^{\mathsf T}}{{N_\flat}\,\nu_{N_\flat}(\bm h)^3},\\
 \bm g_q^\flat-\bm g_q^{+\flat}
 &=\frac{\beta_0\sqrt {N_\flat}}{K}
       D\mathcal N(\bm h)(\bm W^{\ell\uparrow})^{\mathsf T}\bm g_q^\sharp.
\end{aligned}
\end{equation}
Let $\mathsf T_{p,{N_\flat}}(z,z')$ be the full response of
$\bm h^\flat(z)$ to an additive perturbation of its residual
backward variable at $z'$, with population coefficients fixed.
The raw initial up-transpose source enters that variable through
$D\mathcal N(z')$, while the queried normalized input has derivative $D\mathcal N(z)$.
The response replacing $A^\flat_p$ is therefore
\begin{equation}
\label{eq:semifreeze-rms-response}
 A^R_{\beta,p}(z,z')
 =\lim_{{N_\flat}\to\infty}\frac1{N_\flat}\mathbb E\,
       \operatorname{tr}\!\left[
          D\mathcal N(z)\mathsf T_{p,{N_\flat}}(z,z')D\mathcal N(z')\right].
\end{equation}
This retains the rank-one term in both Jacobians.
Contraction over an exposed initial up-weight entry contributes
$(\beta_0/K)\mathbb E\operatorname{tr}[D\mathcal N(z)\mathsf T_{p,{N_\flat}}D\mathcal N(z')]$,
so the population-counting factor remains $\beta_0 {N_\flat}/K$.
Since $\|D\mathcal N\|_{\rm op}\le\nu_{N_\flat}^{-1}$, a lower bound
$\nu_{N_\flat}\ge\nu_{\min}>0$ gives
\begin{equation}
 |A^R_{\beta,p}(z,z')|
 \le \nu_{\min}^{-2}
       \limsup_{{N_\flat}\to\infty}
       \left(\frac1{N_\flat}\mathbb E
                    \|\mathsf T_{p,{N_\flat}}(z,z')\|_{\rm F}^2\right)^{1/2}.
\end{equation}
Thus a bounded normalized response leaves the reaction of order
$\beta_0\alpha=c/\beta_0$.
A fixed positive $\epsilon$ bounds the inverse RMS factors directly;
for $\epsilon=0$ the regular window requires nonvanishing RMS scales
and controlled inverse RMS factors.
The hidden forward equation becomes
\begin{equation}
\label{eq:semifreeze-rms-hidden}
 \bm{\mathscr h}_\beta^\sharp(z)=u^R_\beta(z)
   +\frac{c}{\beta_0}\sum_{p,z'}A^R_{\beta,p}(z,z')\bm{\mathscr g}_{\beta,p}^\sharp(z')
   +\eta_{{\rm eff},\uparrow}
        \mathcal S_t[\bm{\mathscr g}_{\beta,p}^\sharp(z')\Phi_\beta^{\mathcal N}(z,z')].
\end{equation}
Pre-RMS leaves the down-projection gradient and hidden backward
identity unchanged. The second equation in
\eqref{eq:sgd-dmft-hidden} therefore retains its form: its Gaussian
source covariance and direct history still use $G^\flat$, while
$B^\flat$ is evaluated through the full normalized network.
Writing that response as $\mathsf B^R_\beta$ gives
$\mathsf F_\beta=(c/\beta_0)\mathsf B^R_\beta+
\eta_{{\rm eff},\downarrow}\mathsf S_{G,\beta}$.
The same moment bound and Gaussian approximation consequently apply
with $\Sigma_\beta=\Phi_\beta^{\mathcal N}$.
For unit RMS with $\epsilon=0$, each limiting marginal has variance
one, while its correlations across inputs and training steps can
still evolve. With $\epsilon>0$, the marginal variance is
$\Phi^\flat_\beta(z,z)/(\Phi^\flat_\beta(z,z)+\epsilon)$.

This conclusion concerns preactivations $\bm{\mathscr h}^\sharp$, not
$\phi(\bm{\mathscr h}^\sharp)$ or the hidden backward variable containing $\phi'(\bm{\mathscr h}^\sharp)$.
Resolved non-Gaussianity in a fixed-input, fixed-depth preactivation
slice of nonvanishing variance is a one-way diagnostic against
having reached this regular semi-freeze description on the measured
time scale. Approximate Gaussianity alone does not establish
semi-freeze. The bound depends on the training window, so substantial
deformation at finite coupling or after longer training is compatible
with this limiting statement.
\par
\endgroup

\begingroup
\color{black}
\subsubsection{Finite-network illustrations}
\label{app:linear-numerics}
Figures~\ref{fig:linear-profiles} and \ref{fig:linear-stress} compare the
full finite SGD step with the continuum prediction.
At larger $\alpha$, the discrepancies include both a finite-depth
correction to the tangent energy and a higher-order remainder of the
full update. The latter is measured by
\par
\endgroup
\begin{equation}
\label{eq:linear-remainder-measured}
\mathcal R_L=
\frac1{N_\flat}\|\bm h^{L\flat}(1)-\bm h^{L\flat}-(1-f_0)\bm z_L\|^2.
\end{equation}
\begingroup
\color{black}
This squared state remainder is not the total energy correction,
which also contains cross terms and the random factor $1-f_0$.
Implementation checks, sampling conventions, and numerical tables
are given in Appendix~\ref{app:linear-verification}.
\par
\endgroup

\begin{figure}[ht]
\color{black}
\centering
\includegraphics{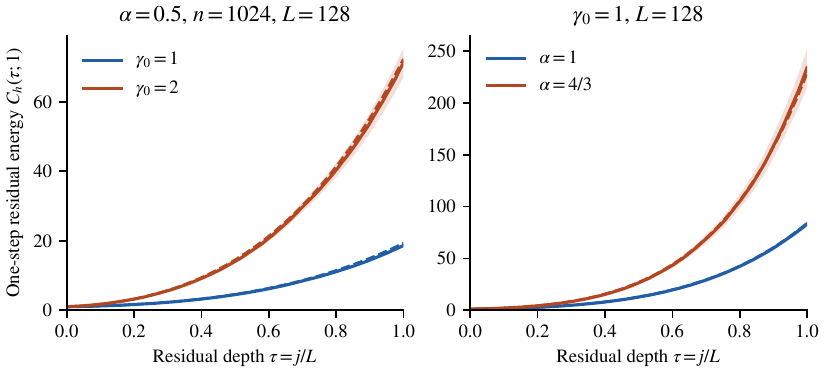}
\caption{Solid curves: full finite-step means; dashed curves:
\eqref{eq:linear-one-step}. Shading is $\pm2$ Monte Carlo standard
errors, not a simultaneous confidence band.
All runs freeze both endpoints and update both residual projections.}
\label{fig:linear-profiles}
\end{figure}

\begin{figure}[ht]
\color{black}
\centering
\includegraphics{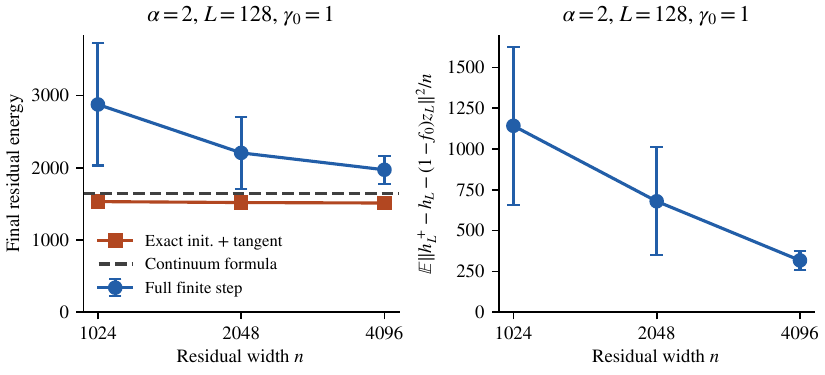}
\caption{A stress test at $\alpha=2$.
Left: the actual finite-step energy, the exact
initialization-plus-unit-error tangent energy, and the continuum
formula. Right: the directly measured higher-order remainder
\eqref{eq:linear-remainder-measured}.
Error bars are $\pm2$ Monte Carlo standard errors.
The tangent curve is a diagnostic, not a replacement training rule.}
\label{fig:linear-stress}
\end{figure}

}
\newpage
\section{Finite-size errors of macroscopic quantities}
\label{app:macroscopic-errors}

\begingroup
\color{black}
For a scalar observable $O$ at a fixed input and training step, compare
the finite model to the population value $o_\infty$ of the same
optimizer. Averaging over initialization separates mean bias from
centered variability:
\begin{equation}
\label{eq:macro-bias-variance}
       \mathbb E|O-o_\infty|^2
       =|\mathbb E O-o_\infty|^2+\operatorname{Var}(O).
\end{equation}
Our size comparisons use $o_\infty(N_\flat/K)$ at the finite ratio
$N_\flat/K$. Comparing to a different target $\alpha$ would also
include $|o_\infty(N_\flat/K)-o_\infty(\alpha)|$.
\par
\endgroup

\subsection{How averaging over depth enters}

\begingroup
\color{black}
Consider a local scalar sampling error $Z_\ell$ at a constant
population kernel, as in the spectral calculation below.
If the two coordinate samples are independently
redrawn for every block, then for
$\overline Z_L=L^{-1}\sum_{\ell=0}^{L-1}Z_\ell$,
\begin{equation}
\label{eq:independent-block-average}
 \mathbb E\overline Z_L=\mathbb E Z_0,\qquad
 \operatorname{Var}(\overline Z_L)
 =\operatorname{Var}(Z_0)/L.
\end{equation}
The inverse-width mean bias survives; the centered standard deviation
is reduced by $L^{-1/2}$.
If the blocks share residual samples $X$ and redraw only the hidden
samples, conditional independence gives instead
\begin{equation}
\label{eq:shared-block-average}
 \operatorname{Var}(\overline Z_L)
 =\operatorname{Var}\bigl(\mathbb E[Z_0\mid X]\bigr)
  +\frac1L\mathbb E\bigl[\operatorname{Var}(Z_0\mid X)\bigr].
\end{equation}
The shared residual randomness leaves a fluctuation floor.
\par
\endgroup

\begingroup
\color{black}
The population saddle averages over ${N_\flat}$ complete residual histories
and ${N_\sharp}$ hidden samples per block.
Hidden kernels and responses enter the residual equations with
weight $1/L$; for example, the forward-source covariance changes by
$(\beta_0^2N_\flat/(KL))\,\delta\Phi^{\ell\sharp}$.
Residual sampling therefore retains its ${N_\flat}^{-1/2}$ scale, while
local hidden sampling errors enter through a depth average.
This averaging also applies directly to hidden observables integrated
over depth, including weighted averages
$L^{-1}\sum_\ell w(\ell/L)\widehat F_\ell$, where
$\widehat F_\ell$ averages a channel observable within block $\ell$
and $w$ is a fixed bounded weight.
A single block has only the local ${N_\sharp}^{-1/2}$ sampling scale:
when ${N_\sharp}=1$, its hidden observable can remain random even as the
depth-integrated observable concentrates.
The Gaussian sources themselves are part of the limiting dynamics;
finite-size error concerns their empirical statistics and the
self-consistent coefficients.

These errors perturb both the population kernels and the loss
feedback. Collect the kernels, responses, and outputs in $\Theta$
and write their closure as $\Theta=\mathcal F_L(\Theta)$.
To first order in the empirical sampling defect $\varepsilon_L$,
\begin{equation}
\label{eq:macro-full-response}
 (I-D\mathcal F_L(\Theta_*))\,\delta\Theta=\varepsilon_L.
\end{equation}
The propagated contributions below include this full response.
The source responses in \eqref{eq:sgd-population-responses} hold
population kernels and loss feedback fixed; they enter this
linearization but are not themselves the full macroscopic response.

For a trained network, the averaging condition must concern each
block's propagated effect on the observable.
In a sampling representation that separates the shared residual sample
from the hidden samples, let $\mathcal R$ denote the shared information
before the hidden samples are revealed. It is not the complete trained
residual trajectory, which already depends on those samples.
Write
\begin{equation}
\label{eq:macro-conditional-split}
\begin{aligned}
 b&=\mathbb EO-o_\infty(N_\flat/K),&
 S&=\mathbb E[O\mid\mathcal R]-\mathbb EO,&
 H&=O-\mathbb E[O\mid\mathcal R].
\end{aligned}
\end{equation}
Conditional centering gives the exact identity
$\mathbb E|O-o_\infty(N_\flat/K)|^2=b^2+\mathbb ES^2+\mathbb EH^2$.
Suppose the shared component has root-mean-square size $O({N_\flat}^{-1/2})$.
For the hidden contribution, write
$H=L^{-1}\sum_\ell\xi_\ell$ with
$\mathbb E[\xi_\ell\mid\mathcal R]=0$.
A sufficient averaging condition is
\begin{equation}
\label{eq:depth-covariance-condition}
 \mathbb E\sum_{\ell,k=0}^{L-1}
 \bigl|\operatorname{Cov}(\xi_\ell,\xi_k\mid\mathcal R)\bigr|
 \le C L/{N_\sharp}.
\end{equation}
It gives $\mathbb EH^2\le C/({N_\sharp}L)$, separately from the shared
${N_\flat}^{-1/2}$ contribution.
The variables $\xi_\ell$ include propagation through the trained
network, with any random response weight included before centering.
Thus \eqref{eq:depth-covariance-condition} is a condition on the
propagated fluctuations, not independence of trained layers.
\par
\endgroup

\subsection{Spectral bias from a finite hidden population}
\label{app:muon-coupled-tests}

\begingroup
\color{black}
The spectral comparison uses scalar actions against coupled test
functions. Sampling the kernel and both test functions on the same
coordinates gives a common observable at every size, even though the
matrices themselves act on different spaces.
This calculation isolates the bias introduced by a finite spectral
window.
\par
\endgroup

\subsubsection{An exact coupled comparison}

\begingroup
\color{black}
Consider a real finite-rank kernel
\begin{equation}
    k(y,x)=\sum_{j=1}^r s_j\psi_j(y)\varphi_j(x),\qquad
    S=\operatorname{diag}(s_1,\ldots,s_r),
\end{equation}
where $s_j>0$ and the two function systems are orthonormal in their
respective probability measures.
Sample $y_1,\ldots,y_{N_\sharp}$ and $x_1,\ldots,x_{N_\flat}$ independently and set
\begin{equation}
\label{eq:coupled-sampled-kernel}
\begin{aligned}
 Y_{aj}&=\psi_j(y_a),& X_{ij}&=\varphi_j(x_i),\\
 P&=Y^{\mathsf T}Y/{N_\sharp},&Q&=X^{\mathsf T}X/{N_\flat},&
 M&=YSX^{\mathsf T}/\sqrt{{N_\sharp}{N_\flat}}.
\end{aligned}
\end{equation}
For fixed test coefficients $v,w\in\mathbb R^r$, let
$v_{N_\sharp}=Yv/\sqrt {N_\sharp}$ and $w_{N_\flat}=Xw/\sqrt {N_\flat}$.
Write $F_c$ for the singular-value map $f_c(s)=\min(s/c,1)$ with fixed
$c>0$. Its empirical action and population target are
\begin{equation}
\label{eq:coupled-scalar-observable}
 A=v_{N_\sharp}^{\mathsf T}F_c(M)w_{N_\flat},\qquad
 a_0=v^{\mathsf T}F_c(S)w.
\end{equation}
Polar factorization gives the exact identity
\begin{equation}
\label{eq:coupled-gram-identity}
 A=v^{\mathsf T}P^{1/2}
       F_c(P^{1/2}SQ^{1/2})Q^{1/2}w.
\end{equation}
For singular Gram matrices the same identity follows by using partial
isometries. The square roots outside $F_c$ are not inverse square roots.

The population-transformed kernel can also be evaluated at the same
sampling points:
\begin{equation}
\label{eq:coupled-paired-reference}
 B=v^{\mathsf T}P F_c(S)Qw,\qquad Z=A-B.
\end{equation}
Independence of the two sample sets gives
$\mathbb E B=a_0$, hence
$\mathbb E Z=\mathbb E A-a_0$.
This paired difference isolates the additional spectral-processing bias.
Its variance need not equal the variance of the physical action $A$.

If the sampled and population spectra both lie in the linear region,
then, for every realization,
\begin{equation}
\label{eq:muon-linear-exact}
       A=v^{\mathsf T}PSQw/c=B,\qquad Z=0.
\end{equation}
The condition must hold for both spectra; a population singular value
below the threshold does not prohibit a finite sample from crossing it.
In update coordinates the same check is
$-\eta_M F_\lambda(D)=-(\eta_M/\lambda)D$.
Thus the effective SGD rate is $\eta_M/\lambda$.
Above threshold each update singular value is $\eta_M$; this is
amplitude normalization, not equality with Adam's update direction.
\par
\endgroup

\subsubsection{Mean bias and centered fluctuations have different orders}

\begingroup
\color{black}
For fixed rank, bounded feature functions, and a spectrum separated
from $c$, the scalar map
\[
 \mathcal A(P,Q)=
 v^{\mathsf T}P^{1/2}F_c(P^{1/2}SQ^{1/2})Q^{1/2}w
\]
has three bounded derivatives near $(P,Q)=(I,I)$.
Repeated singular values away from the clipping threshold are allowed.
Let
$\Xi_\sharp(y)=\psi(y)\psi(y)^{\mathsf T}-I$ and
$\Xi_\flat(x)=\varphi(x)\varphi(x)^{\mathsf T}-I$.
A second-order expansion and independence give
\begin{equation}
\label{eq:muon-spectral-bias}
\begin{aligned}
 \mathbb E Z
 &=\frac{B_\sharp}{{N_\sharp}}+\frac{B_\flat}{{N_\flat}}
   +O({N_\sharp}^{-3/2}+{N_\flat}^{-3/2}),\\
 B_\sharp
 &=\tfrac12\mathbb E\,
   D^2_{PP}\mathcal A(I,I)[\Xi_\sharp,\Xi_\sharp],\\
 B_\flat
 &=\tfrac12\mathbb E\,
   D^2_{QQ}\mathcal A(I,I)[\Xi_\flat,\Xi_\flat].
\end{aligned}
\end{equation}
Centering eliminates the first-order terms; independence eliminates
the mixed second-order mean. The remaining coefficients depend on
the spectrum, clip, tests, and sampling law, and can vanish.
For bounded fixed-dimensional sample covariances,
$\mathbb E\|P-I\|^3=O({N_\sharp}^{-3/2})$ and
$\mathbb E\|P-I\|^4=O({N_\sharp}^{-2})$, and likewise for $Q$.
The third moments control the local Taylor remainder.
The fourth moments bound the probability of leaving the smooth
neighborhood by $O({N_\sharp}^{-2}+{N_\flat}^{-2})$; bounded features control the
observable and polynomial on that event. This proves the displayed
remainder for the scalar action.

For the centered first-order part of $Z$, define
\[
 \xi_\sharp(y)=D_P(\mathcal A-B)(I,I)[\Xi_\sharp(y)],
 \qquad
 \xi_\flat(x)=D_Q(\mathcal A-B)(I,I)[\Xi_\flat(x)].
\]
The same expansion yields
\begin{equation}
\label{eq:muon-centered-variance}
 \operatorname{Var}(Z)
 =\frac{\operatorname{Var}(\xi_\sharp)}{N_\sharp}
  +\frac{\operatorname{Var}(\xi_\flat)}{N_\flat}
  +o({N_\sharp}^{-1}+{N_\flat}^{-1}).
\end{equation}
Thus centered fluctuations and mean bias generally have different
width powers. Either leading coefficient can vanish for particular
spectra or tests.
\par
\endgroup

\subsubsection{A continuous-kernel case with an infinite spectral tail}

\begingroup
\color{black}
The inverse-width bias is not restricted to kernels of small finite
rank. Here is a one-sided version that retains an infinite tail.
Assume $\sum_j s_j^2<\infty$ and
$\sup_{j,y}|\psi_j(y)|<\infty$, sample the hidden side, and keep the
residual side at its population value.
For a fixed mode $j$ with $s_j>0$, write
\begin{equation}
 u(y)=(s_k\psi_k(y))_{k\ge1},\qquad
 S_{N_\sharp}=\frac1{N_\sharp}\sum_{a=1}^{N_\sharp} u(y_a)\otimes u(y_a),\qquad
 \mathbb E S_{N_\sharp}=S^2.
\end{equation}
The coupled mode action is exactly
$A_j=g(S_{N_\sharp})_{jj}/s_j$, where
$g(t)=\min(t/c,\sqrt t)$.
Suppose $c^2$ is outside the spectrum of $S^2$.
With repeated arguments interpreted by continuity, the second divided
difference gives
\begin{equation}
\label{eq:one-sided-bias-coefficient}
\begin{aligned}
 B_j
 &=s_j\sum_k s_k^2
   \operatorname{Var}(\psi_j\psi_k)\,
   g^{[2]}(s_j^2,s_k^2,s_j^2),\\
 \mathbb E A_j-f_c(s_j)
 &=\frac{B_j}{{N_\sharp}}+O({N_\sharp}^{-3/2}).
\end{aligned}
\end{equation}
This follows by expanding $g$ at $S^2$.
Indeed,
$\mathbb E[(S_{N_\sharp}-S^2)_{jk}^2]
 =s_j^2s_k^2\operatorname{Var}(\psi_j\psi_k)/{N_\sharp}$.
The second-order contribution to $g(S_{N_\sharp})_{jj}$ is the sum of these
variances times $g^{[2]}(s_j^2,s_k^2,s_j^2)$.
The series converges absolutely because the features are bounded,
the test mode and threshold are fixed, and $\sum_k s_k^2<\infty$.

For completeness, smooth functional calculus is available here even
though $g$ has a kink at $c^2$.
A fixed contour $\Gamma$ enclosing the finitely many eigenvalues above
$c^2$ gives, near $S^2$,
\begin{equation}
 g(T)=T/c+\frac1{2\pi i}\oint_\Gamma
       (\sqrt z-z/c)(zI-T)^{-1}\,dz .
\end{equation}
The contour lies away from zero and from the spectrum.
The resolvent expansion has a third-order operator-norm remainder.
The bounded Hilbert-space random vector $u(y)$ gives
$\mathbb E\|S_{N_\sharp}-S^2\|_{\mathrm{HS}}^3=O({N_\sharp}^{-3/2})$
and fourth moment $O({N_\sharp}^{-2})$.
As above, this controls the local remainder and the exceptional event.
No infinite empirical Gram matrix is assumed to approach the identity
in operator norm.
This proves the one-sided, fixed-test statement, not a uniform
two-sided infinite-rank bound.
\par
\endgroup

\begingroup
\color{black}
\paragraph{A threshold control.}
For a rank-one kernel, take its two singular functions as the tests and
put
$p={N_\sharp}^{-1}\sum_a\psi(y_a)^2$,
$q={N_\flat}^{-1}\sum_i\varphi(x_i)^2$,
$z=pq$, and $r=s/c$.
Equations \eqref{eq:coupled-scalar-observable} and
\eqref{eq:coupled-paired-reference} reduce exactly to
\begin{equation}
\label{eq:rank-one-coupled-clip}
 A=\min(rz,\sqrt z),\qquad
 B=\min(r,1)z .
\end{equation}
Write $v_\sharp=\operatorname{Var}(\psi^2)$,
$v_\flat=\operatorname{Var}(\varphi^2)$, and
$V=v_\sharp/{N_\sharp}+v_\flat/{N_\flat}$.
In a strictly saturated region $r>1$, bounded features and a fixed
threshold gap give
\begin{equation}
       \mathbb E Z
       =-\frac{v_\sharp}{8{N_\sharp}}-\frac{v_\flat}{8{N_\flat}}
          +o({N_\sharp}^{-1}+{N_\flat}^{-1}).
\end{equation}
At the exact threshold $r=1$, however,
$Z=-(z-\sqrt z)_+$.
Its leading term is $-(z-1)_+/2$.
The scalar central limit theorem, with convergence of the corresponding
first absolute moments, gives
\begin{equation}
\label{eq:clip-kink-bias}
       \mathbb E Z
       =-\frac{\sqrt V}{2\sqrt{2\pi}}+o(\sqrt V).
\end{equation}
At the hard-clipping kink this central-limit scaling replaces the
smooth inverse-width expansion.
\par
\endgroup

The sampled-kernel experiments in Appendix~\ref{app:l01-spectral}
test the smooth bias, centered fluctuations, and threshold control.

\subsection{Combining the finite-size contributions}

\begingroup
\color{black}
The depth error concerns population statistics.
At initialization, for example, the finite-depth variance recursion
is the explicit Euler step
$C_{\ell+1}=C_\ell+(c/L)\mathbb E[\phi(\sqrt{C_\ell}Z)^2]$
for the unnormalized equation in
\eqref{eq:initialization-checkpoint}.
Its local error is $O(L^{-2})$, giving $O(L^{-1})$ over the
unit depth interval when the response remains bounded.
For trained networks, the weak depth expansion applies to the
self-consistent kernels and responses in
\eqref{eq:sgd-dmft-hidden}--\eqref{eq:sgd-dmft-boundaries},
retaining the discrete training histories and forward/backward
boundary coupling. A local remainder of order $L^{-2}$ and bounded
propagation give the same depth scale.

Sampling bias comes from the second response of the closed population
equations. Eliminate the local hidden moments in favor of their
sampling measures $\mu_\ell$.
These measures describe complete channel histories, including the
coordinate-pair moment states for Adam.
For SGD and entrywise Adam, nonlinearities and optimizer states act
on individual channels or coordinate pairs before hidden averaging.
At fixed population coefficients, the resulting hidden moments enter
both the residual drift and source covariance linearly, with weight
$\Delta\tau=1/L$.
The local weak operator on the augmented forward/backward histories
therefore has the expansion
\begin{equation}
\label{eq:macro-hidden-weak-slice}
\begin{aligned}
 \mathcal T_\ell(\mu_\ell)
 &=I+\Delta\tau\,\mathcal L_\ell(\mu_\ell)+O(\Delta\tau^2),\\
 D_{\mu_\ell}^2\mathcal L_\ell&=0,\qquad
 D_{\mu_\ell}^2\mathcal T_\ell=O(\Delta\tau^2).
\end{aligned}
\end{equation}
The expansion uses the covariance, not a particular square root of
the Gaussian noise. Two hidden-sampling insertions therefore carry
two depth weights. Their propagation includes changes in the
kernels and loss feedback through \eqref{eq:macro-full-response}.
This counts the effect of a slice on the full history; it does not
require a small pointwise fluctuation of a fixed-width hidden block.

At second order in this population-response expansion, centered
sampling leaves the same-sample contractions
\begin{equation}
\label{eq:macro-sampling-mean}
 \mathbb E O-O_{\mathrm{pop},L}
 =\frac{b_\flat}{{N_\flat}}
  +\frac{1}{{N_\sharp}L^2}\sum_{\ell=0}^{L-1}b_\ell
  +\text{higher-order terms},
\end{equation}
where $O_{\mathrm{pop},L}$ is the finite-depth population value.
The coefficients include the response of the observable itself and
the contraction of a sampling fluctuation with the response
derivative computed from that same sample; the latter need not
vanish. Bounded propagated coefficients give the sampling-bias
scales ${N_\flat}^{-1}$ and $({N_\sharp}L)^{-1}$.
Together with the depth error, these give
$|b|\lesssim L^{-1}+{N_\flat}^{-1}+K^{-1}$ at the retained orders.
Combining this mean bias with the two variance contributions yields
\begin{equation}
\label{eq:macro-three-scales}
 E_{\mathrm{SGD}}
 \lesssim \frac1{\sqrt {N_\flat}}+\frac1L+\frac1{\sqrt{{N_\sharp}L}}.
\end{equation}
The depth term estimates the mean of the selected statistic.
It is distinct from a trajectorywise strong-error estimate.

For Adam, include its first and second moment states and express the
stabilizer in each group's gradient units. When these dynamics satisfy
the same regularity, propagation, and covariance conditions,
\begin{equation}
\label{eq:adam-three-scales}
 E_{\mathrm{Adam}}
 \lesssim \frac1{\sqrt {N_\flat}}+\frac1L+\frac1{\sqrt{{N_\sharp}L}}.
\end{equation}
These conditions concern Adam's own dynamics. Gaussianity is a
separate question from the control of these errors.

Spectral Muon instead transforms an empirical hidden Gram matrix
before its contribution is averaged over depth.
Its local operator is already nonlinear in $\mu_\ell$ at order
$\Delta\tau$, so the same-sample contraction contributes
$\Delta\tau/{N_\sharp}$ per slice and $1/{N_\sharp}$ after summing over depth.
The smooth local calculation
\eqref{eq:muon-spectral-bias} identifies a mean bias of order
${N_\sharp}^{-1}+{N_\flat}^{-1}$. When the propagated spectral contributions retain
this mean order, their depth average also has this order.
Their conditionally centered parts enter $H$, while fluctuations of
their conditional means enter $S$.
Combining these terms with the same depth and covariance control gives
\begin{equation}
\label{eq:muon-macro-scales}
 E_{\mathrm{Muon}}
 \lesssim \frac1{\sqrt {N_\flat}}+\frac1L+\frac1{\sqrt{{N_\sharp}L}}+\frac1{N_\sharp}.
\end{equation}
The residual-side bias $1/{N_\flat}$ is covered by $1/\sqrt {N_\flat}$, and
\begin{equation}
\label{eq:muon-error-absorption}
       \frac1{\sqrt{{N_\sharp}L}}\le\frac12\left(\frac1{N_\sharp}+\frac1L\right).
\end{equation}
This gives the three-term Muon estimate in the main text.
The additional spectral bias vanishes in the fully linear case
\eqref{eq:muon-linear-exact}.
\par
\endgroup

\subsection{Regrouping the same hidden population}
\label{app:swing-errors}

\begingroup
\color{black}
The main-text pairing example preserves the channel budget exactly.
For even $L$, put ${N_\sharp}'=2{N_\sharp}$ and $L'=L/2$, so ${N_\sharp}'L'=K$, and stack
the projection matrices as
\begin{equation}
\label{eq:swing-paired-matrices}
\begin{aligned}
 \widetilde{\bm W}^{r\uparrow}
 &=\begin{pmatrix}
     \bm W^{2r\uparrow}\\ \bm W^{(2r+1)\uparrow}
   \end{pmatrix},
 &\widetilde{\bm W}^{r\downarrow}
 &=\begin{pmatrix}
     \bm W^{2r\downarrow}&\bm W^{(2r+1)\downarrow}
   \end{pmatrix}.
\end{aligned}
\end{equation}
Both hidden populations now read the residual state at the start of
the pair, and their outputs add with the unchanged coefficient
$\beta_0\sqrt {N_\flat}/K$. This is exactly a block of width $2{N_\sharp}$.
Compared with the original sequential network, the second population
reads the state one residual update earlier, as in the main-text
Euler interpretation.
SGD and entrywise Adam updates commute with stacking the paired
computation, since their code-coordinate rates and gradient units
depend on ${N_\flat},K$; corresponding Adam moments are stacked as well.
A nonlinear spectral transform acts on each whole block and need not
commute with stacking. Its finite-${N_\sharp}$ bias is the additional
contribution identified above.

The population coefficients give the same Euler interpretation.
Replacing $({N_\sharp},L)$ by $(2{N_\sharp},L/2)$ leaves $N_\flat/K$ and the hidden
reaction coefficient $\beta_0N_\flat/K$ unchanged.
The residual drift coefficients and source covariances double,
as required for a doubled depth step.
For example, source independence gives
\begin{equation}
\label{eq:swing-source-pairing}
\begin{aligned}
 \operatorname{Cov}(u^{(2r+1)\flat}+u^{(2r+2)\flat})
 &=\frac{\beta_0^2N_\flat}{KL}
       (\Phi^{2r\sharp}+\Phi^{(2r+1)\sharp})\\
 &=\frac{\beta_0^2N_\flat}{KL'}\Phi^{2r\sharp}+O(L^{-2}),
 \qquad L'=L/2,
\end{aligned}
\end{equation}
where the last equality uses regular depth variation of the population
kernels. The backward-source covariance obeys the same calculation
with $G^\sharp$. The comparison concerns the Gaussian sources;
trained residual increments also contain the learning feedback.

Consider two regroupings with the same ${N_\flat},K$, optimizer, data, training
step, and limiting coefficients. Let their depths be $L_1,L_2$,
with hidden widths $N_{\sharp,j}=K/L_j$, and compare the same macroscopic
statistic $O_j$ in each network.
Both approximate the common value $o_\infty({N_\flat}/K)$.
For any coupling of their initializations, including reuse of the same
ordered hidden computations, the triangle inequality gives
\begin{equation}
\label{eq:swing-error-comparison}
\begin{aligned}
 \|O_1-O_2\|_2
 &\le \|O_1-o_\infty({N_\flat}/K)\|_2+\|O_2-o_\infty({N_\flat}/K)\|_2\\
 &\lesssim {N_\flat}^{-1/2}+K^{-1/2}+L_1^{-1}+L_2^{-1}
 \qquad\text{(SGD or Adam)}.
\end{aligned}
\end{equation}
For smooth spectral Muon, add $N_{\sharp,1}^{-1}+N_{\sharp,2}^{-1}$.
This bound does not require or assert trajectorywise agreement of
individual residual coordinates. It compares macroscopic statistics
within one optimizer, and does not exploit any additional cancellation
from coupling the two finite networks.
\par
\endgroup

\begingroup
\color{black}
\section{Half Copy: a correlated-initialization recipe}
\label{app:half-copy-recipe}

Half Copy restores an initial residual-variance scale by correlating
nearby channels, while keeping the network geometry, coupling constants,
and learning-rate scalings unchanged.
The construction has two choices: how many channels share a parent,
and how strongly their down-projections are correlated.
The four-channel example in Eq.~\ref{eq:half-copy-main} is one instance.

\subsection{Layer-local groups and Gaussian weights}

Consider one channel family in Eq.~\ref{eq:transformer-channel-form},
with total population $K$ and a desired variance multiplier $s\geq1$.
For fixed coupling, preserving an anchor's coefficient
$\beta_0^2N_\flat/K$ requires
\begin{equation}
\label{eq:half-copy-target}
s=\frac{(N_\flat/K)_{\rm anchor}}{N_\flat/K}.
\end{equation}
Growing both widths and depth by $R$ therefore gives $s=R$.
Input and readout initialization are left unchanged.

Partition consecutive channels within each layer into disjoint groups
$\mathcal G_g$ of integer sizes $c_g$, and let $p(a)$ identify the
group containing channel $a$.
In mathematical-weight coordinates, initialize
\begin{equation}
\label{eq:half-copy-general}
\begin{aligned}
\bm W_a^\uparrow&=\bm U_{p(a)},\\
\bm W_a^\downarrow
 &=\sqrt{\rho_{p(a)}}\,\bm Z_{p(a)}
   +\sqrt{1-\rho_{p(a)}}\,\bm E_a,\qquad 0\leq\rho_g\leq1.
\end{aligned}
\end{equation}
Parent arrays $\bm U_g,\bm Z_g$ and innovations $\bm E_a$ have
independent standard Gaussian entries and are mutually independent.
Different groups and layers use independent draws.
For an MLP channel, these arrays are an up-projection row and
down-projection columns; their code weights absorb the coefficients
in Eq.~\ref{eq:mlp-forward}.
Each entry retains its standard Gaussian marginal.

\subsection{Matching variance with integer groups}

Channels sharing a parent see the same incoming state and initially
produce the same activation $A_g(\bm x)$.
For one residual coordinate, their summed down-weight has variance
\begin{equation}
\label{eq:half-copy-group-variance}
\operatorname{Var}\!\left(\sum_{a\in\mathcal G_g}W^\downarrow_{ia}\right)
=c_g\bigl[1+(c_g-1)\rho_g\bigr].
\end{equation}
Thus, conditional on the incoming states and up-projections, the
initial residual increment $\Delta h_i^\flat$ has covariance
\begin{equation}
\label{eq:half-copy-increment-covariance}
\begin{aligned}
&\operatorname{Cov}_{\downarrow}
\bigl(\Delta h_i^\flat(\bm x),\Delta h_i^\flat(\bm x')\bigr)
\\
&\qquad=\frac{\beta_0^2N_\flat}{K^2}
\sum_{g\ {\rm in\ layer}}c_g\bigl[1+(c_g-1)\rho_g\bigr]
A_g(\bm x)A_g(\bm x').
\end{aligned}
\end{equation}
Choosing
\begin{equation}
\label{eq:half-copy-correlation-calibration}
\rho_g=\frac{s-1}{c_g-1},\qquad c_g\geq s,\quad c_g>1,
\end{equation}
makes every group's variance multiplier exactly $s$.
After averaging over the up-projection parents, this multiplies the
branch covariance by $s$ without changing the single-channel activation
law. The residual variance coefficient becomes
$\beta_0^2(N_\flat/K)s$.
For $s=1$, use $\rho_g=0$; singleton groups need no correlation.

Integer sizes need not divide the layer width.
For a layer with $m$ channels, choose a number $G$ of parent groups,
set $c=\lfloor m/G\rfloor$ and $r=m-Gc$, and use $r$ groups of size
$c+1$ and $G-r$ groups of size $c$.
Choose $1\leq G\leq\lfloor m/\lceil s\rceil\rfloor$, so that
$c\geq s$, then apply
Eq.~\ref{eq:half-copy-correlation-calibration} separately to the two
sizes. This matches the target exactly despite the uneven partition.
Taking $c>s$ keeps $\rho_g<1$ and leaves independent innovations;
for example, group sizes near $\lambda s$ with fixed $\lambda>1$
leave $1-\rho_g$ of order one.
If a layer has fewer than $s$ channels, this layer-local construction
cannot supply the requested enhancement.

Alternatively, fixing $\rho$ first makes the ideal common group size
$c=1+(s-1)/\rho$.
At $\rho=1/3$, this is $c=3s-2$: $c=4$ when $s=2$, and approximately
$3R$ under conventional growth.
For balanced integer groups with fixed $\rho$, the realized
layerwise multiplier is
\begin{equation}
\label{eq:half-copy-realized-enhancement}
\chi_\ell
=1+\rho\,\frac{\sum_{g\ {\rm in\ layer}}c_g(c_g-1)}{m}.
\end{equation}
The CIFAR and Tiny Shakespeare growth experiments use this fixed-$\rho$
version, recording the remaining integer mismatch rather than
retuning $\rho$.
Their task-specific choices are given in
Appendices~\ref{app:cifar-growth-recipes} and
\ref{app:shakespeare-protocol}.

\subsection{Attention, locality, and learning}

For attention, group whole heads.
Copy the Q, K, and V up-maps within each group, using independent
parents for the three maps, and apply
Eq.~\ref{eq:half-copy-general} to their output-projection matrices.
The grouped heads then compute identical attention outputs at
initialization. In Eq.~\ref{eq:half-copy-increment-covariance}, products
of scalar activations become inner products of the head outputs,
with the head-dimension normalization already included in $\phi_{\rm A}$.
MLP units and attention heads have separate populations and groups;
tokens and coordinates within a head are not grouped as extra channels.

Under conventional growth, group sizes are $\Theta(R)$ while $K=\Theta(R^2)$.
The total number of independent parent groups therefore grows as
$K/R=\Theta(N_\flat)$, and each group's fraction of the ordered hidden
population vanishes.
Groups stay within a single layer, so correlated copies read the same
residual state; copying across layers would not obey the same
finite-network covariance identity.

Copying is an initialization operation, not parameter tying:
materialize every matrix separately and use separate optimizer states
with the rates of Table~\ref{tab:optimizer-scaling}.
When $\rho_g<1$, distinct down-projections give the copied
up-projections distinct backward signals, allowing them to separate
from the first update
(Appendix~\ref{app:half-copy-unlocking}).
The matching calculation fixes the initial variance contribution;
it does not identify the subsequent learning dynamics with those
of independently initialized fixed-ratio networks.
\par
\endgroup

\begingroup
\color{black}
\section{Controlled checks of theoretical predictions}
\label{app:controlled-checks}

These controlled experiments test residual distribution shapes,
the one-step checkpoint, spectral sampling error, and the separation
of initially copied features.
Protocols and quantitative records are collected here, with the
distributional and checkpoint figures alongside the theoretical arguments.

\subsection{A nonlinear logic task with actual backward propagation}
\label{app:l01-logic}

The inputs are $00,01,10,11$, in that order; the output coordinates are
AND, OR, and XOR. Put $n=N_\flat=N_\sharp$ and augment the input as
$x'=(x_1/\sqrt2,x_2/\sqrt2,1)$. The stored matrices compute
\begin{equation}
\begin{aligned}
 h&=Ex', & z&=Uh/\sqrt n,\\
 H&=h+V\tanh(z)/\sqrt n, & f&=A^{\mathsf T}H/n.
\end{aligned}
\end{equation}
The three columns of $E$ have independent Gaussian entries of standard
deviation $0.01$; $U,V,A$ have standard Gaussian entries.
Thus the constant input implements a trained input bias. There are no
other biases, normalization, dropout, or weight decay.
The objective is half the mean squared error over the twelve targets;
binary predictions use threshold $1/2$.
Widths are $64,128,256,512,1024$, with seeds $100001$--$100008$.
Each optimizer runs 1000 full-batch updates in CPU float64.
Optimizers share their initial arrays, and widths use nested prefixes
within each seed. All parameter groups update simultaneously.

For explicit optimizer units, let $\Gamma$ be the stored loss gradient
multiplied by $n$ for $E,A$, or by $n\sqrt n$ for $U,V$.
SGD applies rate $0.1$ to $\Gamma$, with an additional $1/\sqrt n$
for $U,V$. Adam uses rate $0.03$ and the same group factors,
$(\beta_1,\beta_2)=(0.9,0.999)$, bias correction, and
$\epsilon=10^{-8}$ in these normalized-gradient coordinates.

The spectral comparison uses the odd polynomial
$p(s)=15s/8-5s^3/4+3s^5/8$, composed five times:
\begin{equation}
 F(\Gamma)=\frac{\lambda_n}{(15/8)^5}
                  p^{\circ5}(\Gamma/\lambda_n).
\end{equation}
The matrix polynomial acts on singular values, preserving singular vectors.
Its outer learning rate is four times the SGD rate above.
For each group, the fixed bound is 1.25 times the maximum SGD singular
value across all planned widths, seeds, and updates, after dividing spectra
by $\sqrt n$ for $E,A$ and by $n$ for $U,V$.
The resulting dimensionless bounds, in this order, are
$0.7609752890$, $0.1501675784$, $0.06461736165$, and $0.08052997111$.
They are multiplied back by $\sqrt n$ or $n$ to obtain $\lambda_n$.
The largest observed singular-value/bound ratio is $0.839$;
bound violations would stop a run rather than change its recipe.
Exact low-rank evaluation retains all four input contributions and performs
no numerical rank truncation.
The pooled calibration couples widths and seeds: equivariance of a
fixed-bound update does not alone establish rotational invariance after
conditioning on this entire calibration procedure.

The external gain was selected before adding the input bias, using learning
progress and domain safety rather than Gaussianity. That earlier small-input
control is informative: strong spectral compression at unit gain left all
eight largest-width seeds short of solving the task, with mean loss
$0.0428$. Gain four resolved this slow-learning failure.
A separate standard-input control changed the update direction by roughly
$50$--$60\%$ in the four parameter groups relative to the best scalar
multiple of the same SGD gradient. Spectral processing therefore need not
be an almost-linear update, but these differently calibrated methods are
not a matched-cost optimizer benchmark.

We record $h,H,A$, and the actual rescaled Jacobian
$g_q^{0\flat}=n\,\partial f_q/\partial h$ at updates
$0,1,2,4,8,16,32,64,128,256,512,1000$.
The derivative includes the nonlinear block and the identity path;
it excludes the loss derivative.
For a chosen pair, each coordinate vector is centered and divided by its
own empirical standard deviation within each run. With the resulting
$z_i$, covariance $C$, and empirical average over residual coordinates,
we calculate
\begin{equation}
\begin{aligned}
 \kappa_{ijkl}&=\langle z_i z_j z_k z_l\rangle
                 -C_{ij}C_{kl}-C_{ik}C_{jl}-C_{il}C_{jk},\\
 Q_4&=\left(\frac1{16}\sum_{i,j,k,l=1}^{2}\kappa_{ijkl}^2\right)^{1/2}.
\end{aligned}
\end{equation}
We summarize eight run-level values, without pooling seeds or inputs.
Residual coordinates are not treated as independent experimental replicates.
The plotted scatter uses the predetermined seed $100001$, width 1024,
and update 1000; it is centered only, without whitening or rotation.
If shown, covariance ellipses are Gaussian reference contours, not confidence
regions for the observed samples.

All 120 runs across the five widths, eight seeds, and three optimizers
classify all twelve targets correctly.
The joint distributions are illustrated in
Fig.~\ref{fig:logic-actual-backward} in Appendix~\ref{app:gaussian-numerics}.
Table~\ref{tab:l01-cumulants} reports the RMS of standardized joint
fourth cumulants across eight seeds, including forward pairs from
two inputs.

\begin{table}[!ht]
\centering
\color{black}
\begin{tabular}{@{}lccc@{}}
\toprule
Residual pair & SGD & Taylor & Adam\\
\midrule
$h^{0\flat}(11),g_{\mathrm{AND}}^{0\flat}(11)$
 & $.0601\pm.0293$ & $.0687\pm.0327$ & $.6109\pm.0172$\\
$h^{0\flat}(11),h^{0\flat}(00)$
 & $.0667\pm.0216$ & $.0712\pm.0217$ & $1.1086\pm.1195$\\
$h^{1\flat}(11),h^{1\flat}(00)$
 & $.0790\pm.0327$ & $.0692\pm.0303$ & $.3654\pm.0591$\\
\bottomrule
\end{tabular}
\caption{Joint fourth-cumulant RMS after 1000 updates at width 1024,
reported as mean $\pm$ sample standard deviation across eight seeds.}
\label{tab:l01-cumulants}
\end{table}

The complementary two-layer linear control removes the block and input bias,
using $h=E(x_1,x_2)/\sqrt2$, $f=A^{\mathsf T}h/n$, and the same loss,
input standard deviation, optimizer units, eight seeds, and 1000 updates.
Its widths are $256,1024,4096,16384,65536$.
Both SGD and Adam approach the linear loss floor $1/12$ and fail XOR.
At the largest width, their excess kurtoses for $h(11)$ are
$-0.0044\pm0.0279$ and $-1.9412\pm0.0015$, respectively.
SGD's variance grows from approximately $10^{-4}$ to $1.21$, so its Gaussian
appearance does not result from frozen features.
An exact finite-dimensional SGD recurrence matches its trajectory to
$1.09\times10^{-14}$.
The separate one-step linear residual calculation, including its finite-width
discrepancies, is checked in Appendix~\ref{app:linear-verification}.

\paragraph{Raw mixed moments across initializations.}
\label{app:logic-raw-gram}
Gaussian-shaped coordinate clouds need not have deterministic covariance.
We also examine raw mixed inner products in the same logic-task archive,
using all five widths and eight seeds.
At each of updates 0 and 1000 we collect
$h(00),h(11),H(00),H(11),g_{\mathrm{AND}}^{0\flat}(11),
g_{\mathrm{XOR}}^{0\flat}(11),A_{\mathrm{AND}},A_{\mathrm{XOR}}$,
where $h,H$ are the residuals before and after the block and $A_q$ is
the rescaled final-residual output Jacobian.
Chronological concatenation gives $Y\in\mathbb R^{n\times16}$.
We retain $Y^{\mathsf T}Y/n$ and the coordinate means without centering,
whitening, or run-specific rescaling. All 2,880 matching-time readout
values $H^{\mathsf T}A/n$ reproduce the archived predictions exactly.

Across widths 64--1024, the RMS of the 64 entrywise seed standard
deviations in the final-time Gram block decreases from $0.1376$ to
$0.0398$ for SGD, $0.1462$ to $0.0407$ for Taylor, and $0.1925$ to
$0.0596$ for Adam.
The corresponding RMS magnitudes of the seed-mean entries change from
$1.2431$ to $1.2366$, $1.2497$ to $1.2285$, and $1.6173$ to $1.6325$.
Cross-time block dispersions also decrease over this range.
These summaries do not imply monotonic improvement: for example, the
SGD final-residual second-moment SD increases from $0.0727$ at width
512 to $0.0809$ at width 1024, while the aggregate final-time dispersion
barely changes over that last doubling ($0.0403$ to $0.0398$).
The accompanying data retain all entries, coordinate means, and
seed-paired adjacent-width differences; the predefined scalar summaries
include forward/backward, cross-input, cross-output and cross-time pairs.

The observations support increasing reproducibility of these raw moments
over the tested widths, including for non-Gaussian Adam features.
They do not establish convergence to a deterministic Gram matrix, a
concentration rate, or a whole-process limit.
Standard deviations use seeds, not neurons, as replicates; widths are
paired by their nested initializations. Taylor remains descriptive
because its calibration pools these same widths and seeds.

\begingroup
\color{black}
\subsection{Numerical verification of the linear checkpoint}
\label{app:linear-verification}
The simulations implement \eqref{eq:linear-exact-update} in NumPy
float64 on a CPU, retaining all $L+1$ residual states.
Each realization uses an independent Gaussian initialization and one
simultaneous SGD step for the squared loss $\frac12(f-1)^2$.
All finite-step runs in this subsection use $\eta_{\rm eff}=\gamma_0$.
The input embedding can be sampled directly as
$\bm h^{0\flat}\sim\mathcal N(0,I_{N_\flat})$, which has exactly the same distribution
as the frozen normalized Gaussian embedding in
\eqref{eq:linear-finite-model}.

\paragraph{Implementation checks.}
PyTorch automatic differentiation verified
both parameter gradients and the full updated forward pass.
At ${N_\flat}=128$, ${N_\sharp}=8$, $L=32$, and $\beta_0=\gamma_0=1$, the maximum absolute
gradient-update discrepancy was $7.3\times10^{-16}$ and the final-state
discrepancy was $7.2\times10^{-15}$.
A second check at $({N_\flat},{N_\sharp},L,\beta_0,\gamma_0)=(64,8,16,0.7,1.3)$
gave discrepancies below
$7.8\times10^{-16}$ and $3.2\times10^{-15}$, respectively.
No input or readout gradient was applied.

The five-moment recursion was checked separately by constructing
$\bm M_j$ directly in $50{,}000$ Gaussian realizations at
$({N_\flat},{N_\sharp},L,\beta_0)=(8,3,4,0.7)$.
All five terminal means were within $0.79$ Monte Carlo standard
errors of \eqref{eq:linear-exact-moments}.
Across the finite-network runs with at least $32$ realizations,
the measured tangent energies agreed with
\eqref{eq:linear-moment-transfer} to within $2.10$ standard errors.
These comparisons test the implementation and the finite-size
moment identities separately.

\paragraph{Depth and shape conventions.}
Residual state $\bm h^{j\flat}$ is assigned depth $j/L$.
The plots include both endpoints, $j=0,\ldots,L$; a trajectory
containing only $\bm h^{0\flat},\ldots,\bm h^{(L-1)\flat}$ would end at
$(L-1)/L$, not at $1$.
All theoretical predictions use the realized ratio ${N_\flat}/({N_\sharp}L)$.
For example, at ${N_\flat}=1024,L=128$, requested ratios
$1.25,1.5,2.5$ yield ${N_\sharp}=6,5,3$ after integer rounding, corresponding
to actual ratios $4/3,1.6,8/3$.
The large-ratio discrepancies reported below remain after accounting
for the realized dimensions.

\paragraph{Finite-step measurements.}
Table~\ref{tab:linear-finite-checks} reports the energy after the
full finite update. The uncertainty is one Monte Carlo standard
error of the mean. All rows use $L=128$, $\beta_0=1$, and
$\eta_{\rm eff}=\gamma_0$.
Figure~\ref{fig:linear-profiles} in Appendix~\ref{app:linear-numerics}
shows representative complete depth profiles, including the actual
ratio $4/3$. Its left panel uses $128$ seeds per curve; the right
uses $256$, with $(N_\flat,N_\sharp)=(2048,16)$ at $\alpha=1$
and $(1024,6)$ at $\alpha=4/3$.

\begin{table}[ht]
\color{black}
\centering
\small
\begin{tabular}{rrrrrrr}
\toprule
$N_\flat$ & $N_\sharp$ & $N_\flat/K$ & $\gamma_0$ & Seeds
& Full-step $C_h(1;1)$ & Limit \eqref{eq:linear-one-step}\\
\midrule
1024 & 16 & $0.5$ & 1 & 128 & $18.63\pm0.28$ & 19.32\\
1024 & 16 & $0.5$ & 2 & 128 & $71.21\pm2.10$ & 72.32\\
2048 & 16 & $1$ & 1 & 256 & $83.06\pm1.40$ & 84.00\\
1024 & 6 & $4/3$ & 1 & 256 & $233.83\pm9.35$ & 228.38\\
1024 & 4 & $2$ & 1 & 128 & $2876.95\pm423.02$ & 1645.33\\
2048 & 8 & $2$ & 1 & 256 & $2206.47\pm250.25$ & 1645.33\\
4096 & 16 & $2$ & 1 & 256 & $1972.17\pm96.10$ & 1645.33\\
\bottomrule
\end{tabular}
\caption{Residual energy after one finite SGD step for the squared loss.
The continuum formula is not asserted to be exact at these sizes.
In particular, the large-$\alpha$ discrepancy remains visible.}
\label{tab:linear-finite-checks}
\end{table}

\paragraph{Separating two finite-size effects.}
At $\alpha=2,L=128,\gamma_0=1$, the exact tangent energies for
${N_\flat}=1024,2048,4096$ are respectively
$1523.90,1510.75,1504.24$, while the continuum tangent contribution
is $1637.94$. Thus a depth correction is already present before
the finite-step remainder is considered.
At ${N_\flat}=4096$, increasing $L$ from $128$ to $256$ while keeping
$\alpha=2$ changes the exact tangent prediction to $1572.84$,
closer to its continuum value.
These deterministic comparisons use the exact finite-size recurrence,
not a fit to the Monte Carlo curves.

The full step has a further remainder, measured directly by
\eqref{eq:linear-remainder-measured}.
At the three widths above, its means are
$1142.45,\ 679.59,\ 317.05$.
This substantial contribution decreases over the tested width range;
we do not infer a uniform asymptotic rate from three noisy points.
Figure~\ref{fig:linear-stress} displays both effects.
The correction to residual energy is not equal to $\mathcal R_L$:
cross terms and the random factor $1-f_0$ also contribute.

For a milder depth check with ${N_\flat}=1024,\alpha=0.5,\gamma_0=1$,
the exact tangent energies at $L=64,128,256$ are
$17.2195,17.4740,17.6032$; the continuum value is $17.6688$.
The corresponding sampled tangent means are
$17.2566,17.2774,17.6069$.
This comparison distinguishes deterministic finite-size corrections
from the sampling fluctuations in a full-step depth sweep.

Finally, the general-$\beta_0$ formula was checked at
$({N_\flat},{N_\sharp},L,\beta_0,\gamma_0)=(1024,16,128,0.7,1.3)$ with $256$ seeds.
The full-step terminal energy was $8.479\pm0.074$, compared with
$8.629$ from \eqref{eq:linear-general-beta}.
The finite-network mean remains slightly below the continuum prediction.

\par
\endgroup

\subsection{Coupled scalar tests of spectral sampling error}
\label{app:l01-spectral}
\label{app:spectral-bias-experiment}

A sampled kernel isolates spectral processing from network dynamics.
For independent uniform coordinates on $[0,1]$, use
$\psi_j(y)=\sqrt2\sin(2\pi jy)$, $\varphi_j(x)=\sqrt2\cos(2\pi jx)$,
and $s_j=(1+(j/8)^2)^{-5/8}$, giving an infinite square-summable tail.
At any finite spectral truncation, put $Y_{aj}=\psi_j(y_a)$ and
$X_{ij}=\varphi_j(x_i)$. With $S=\operatorname{diag}(s_j)$
and $m,n$ coordinates, set
$M=YSX^{\mathsf T}/\sqrt{mn}$,
$P=Y^{\mathsf T}Y/m$, and $Q=X^{\mathsf T}X/n$.
Let $F_c$ apply $s\mapsto\min(s/c,1)$ to singular values.
With coupled mode-one test vectors
$v=Ye_1/\sqrt m$ and $w=Xe_1/\sqrt n$, compare the scalars
\begingroup
\color{black}
\begin{equation}
 Z=v^{\mathsf T}F_c(M)w-e_1^{\mathsf T}P F_c(S)Qe_1.
\end{equation}
\endgroup
Both terms use the same sampled coordinates. This comparison remains
meaningful when matrix sizes change; it does not subtract unlike matrices.

At $c=0.5$, the population spectrum has a positive gap from the clipping
threshold. Sampling the hidden side with seeds $718000$--$718127$,
using nested prefixes at widths $512,1024,2048,4096$ and keeping the
residual side at its population value
gives mean biases $-0.004095,-0.001977,-0.001001,-0.000509$.
The analytic leading coefficient is $-2.0323153$, rather than a fitted
amplitude. The fitted exponents of absolute bias and centered standard
deviation are $-1.001$ and $-0.480$, respectively.
Their 95\% bootstrap intervals are $[-1.031,-0.969]$ and
$[-0.562,-0.403]$, resampling seed identities jointly across widths.
The mean estimator subtracts the zero-mean control variate
$-\tfrac12(P_{11}-1)$; fluctuations use the original $Z$.
The paired bootstrap uses 500 resamples with seed $230017$.
At fixed $m=128,n=256$, averaging $L=1$--$64$ independent blocks in
64 independent sequences reduces centered dispersion with fitted exponent
$-0.539$ in $L$, while leaving mean bias near $-0.0244$.
Both coordinate samples are redrawn between blocks, unlike shared network
coordinates.

Two controls delimit the inverse-width interpretation. In the entirely
linear region, paired error and regrouping differences are at numerical
roundoff ($5.6\times10^{-17}$ and $1.1\times10^{-16}$).
At an exact clipping kink, a rank-one example instead gives mean-bias
exponent $-0.501$. Thus the smooth-region bias should not be extrapolated
through the kink or equated with a universal trained-network error.
The kernel is evaluated using a $2^{22}$-point Fourier table with linear
interpolation; refinement to $2^{24}$ changes the tested actions by at most
$5.6\times10^{-7}$. Seeds, scalar statistics, and reference coefficients
are retained in the accompanying data.
\subsection{A control for separating initially copied features}
\label{app:half-copy-unlocking}

A ReLU residual-MLP control uses $D_{\textrm{i}}=12$,
$D_{\textrm{o}}=10$, $L=3$, no biases or normalization,
$\beta_0=2.25$, and $\gamma_0=1$.
Float32 SGD on mean cross-entropy uses standard-coordinate rates
$0.002N_\flat$ for input/readout and $0.002K$ for both branch matrices.
With $N_\flat=32,N_\sharp=8$, nine standard-normal inputs and labels
$0,\ldots,8$, groups of four identical up-projections receive
down-projection correlation $1/3$; parameters remain untied.
After subtracting each group's mean, the up-gradient RMS is
$8.98095\times10^{-3}$; one update raises the corresponding
up-weight RMS from zero to $4.31086\times10^{-4}$.

A separate exact-duplication check copies both projections twice within
each block ($N_\flat=16,N_\sharp=4\to8$), with the compensating
doubling of $K$ and branch learning rates.
On seven Gaussian inputs and labels $0,\ldots,6$, logits match within
$9.54\times10^{-7}$ after three updates.
Both use initialization seed 7319; input seeds are 73190091 and
73190090, respectively. The first check demonstrates separation of
initially copied features. The second checks compensated exact
duplication, not equivalence between correlated and independent
initialization laws.

\par
\endgroup

\begingroup
\color{black}
\section{CIFAR-10 residual-MLP experiments}
\label{app:l01-protocols}
\label{app:empirical-protocols}

The CIFAR-10 \citep{krizhevsky2009learning} experiments below separate convergence, scaling recipes,
and learning-rate selection. They use different normalization and
learning-rate settings, specified for each comparison.

\subsection{CIFAR-10 growth and coupled grouping}
\label{app:l01-cifar}

The network takes 3072 flattened pixel values mapped to $[-1,1]$ and returns
ten logits. It uses the four-projection parametrization of
\eqref{eq:mlp-forward}, ReLU, Gaussian initialization, and cross-entropy loss.
There is no augmentation, bias, normalization, momentum, weight decay, or
learning-rate schedule. In standard-Gaussian coordinates, SGD uses
$\eta_0\gamma_0^2N_\flat$ for input/readout and
$\eta_0\gamma_0^2K$ for up/down matrices, with
$\eta_0=0.02$, $\gamma_0=1$, and $\beta_0=2.25$.
Training uses PyTorch \citep{paszke2019pytorch} with MPS float32,
batch size 64, and ten epochs.
Keeping the final batch of 16 gives 782 updates per epoch.
The implementation specifies a common shuffled minibatch sequence;
initialization seeds are separate from data order.
The older checkpoints do not retain individual batch-index arrays or
per-run source revisions.

The growth family uses $N_\flat=q^2$ and $N_\sharp=L=q$.
The tested values are $q=8,10,\ldots,24$ and $q=28,32,40,64,80,96$,
with 32 distinct initialization seeds per size, listed in the data export.
Logit RMSE is computed on the first ten test images per class, over all
$100\times10$ output entries. Each model is compared with the mean output
of the 32 $q=96$ models at the same epoch; distances are then averaged.
Figure~\ref{fig:cifar-convergence} shows twice the standard error of
these run-level distances, conditional
on this finite reference. The $q=96$ self-reference is excluded when fitting
size dependence. Its individual-to-mean RMSE is $0.129$ at epoch one and
$1.272$ at epoch ten, which measures finite reference dispersion, not an
asymptotic error floor. Full training/test losses and accuracies use all
50,000/10,000 images, rather than this probe.
The kernel panel uses the final residual states on the first ten training
images per class. Their uncentered Gram matrix is normalized to unit
diagonal, giving a cosine kernel. Each run's entrywise kernel RMSE is
measured against the mean of the 32 $q=96$ kernels at the same epoch.
The means and twice-standard-error bars are computed across runs,
as for the output errors.

Unweighted least-squares fits of log error against $\log q$ over all
14 plotted sizes ($q=8$--$80$) give output/kernel slopes of
$-0.808/-0.820$ at epoch one and $-0.289/-0.320$ at epoch ten.
Over the larger half of these sizes,
$q\in\{22,24,28,32,40,64,80\}$, the corresponding slopes are
$-0.948/-0.896$ and $-0.230/-0.262$.
These are descriptive fits to finite-reference errors.
The $q^{-1}$ guides in Fig.~\ref{fig:cifar-convergence} indicate the
theoretical exponent; their vertical offsets are chosen for readability.

The grouping study fixes $N_\flat=K=q^2$ for $q=24,36$, using all
21 and 25 divisor depths, respectively. Both families use seed $101234$.
Each family shares its input/output matrices and one ordered pool of hidden
weights, partitioned into consecutive groups of size $K/L$.
It has one coupled realization, not an ensemble of independent depths.
Changing the grouping changes the residual states at which subsequent
groups are evaluated. The $L=K$ model is the finest finite reference;
its zero distance to itself is omitted.
Only $q=24$ appears in Fig.~\ref{fig:cifar-convergence} (right);
both sizes are retained in the numerical supplement.
At fixed $N_\flat$ and $K$, the population-sampling terms in
Theoretical Result~\ref{thm:finite-errors} do not decrease with depth,
so this comparison is not assigned a pure $L^{-1}$ reference slope.
Early output agreement improves towards this reference, but later agreement
can flatten or be nonmonotone. Likewise, raw residual Grams retain amplitude
changes removed by cosine normalization. At epoch one, $q=36$, the mean raw
Gram diagonal at $L=1,36,72,1296$ is $2.56,64.84,66.41,65.37$.
Thus neither every added layer nor every diagnostic improves monotonically.

\paragraph{Direct comparisons of adjacent groupings.}
\label{app:adjacent-swing}
Distances to one reference do not determine how close two other models
are to each other. We therefore also compare the saved raw logits
$F_{q,L,e}\in\mathbb R^{100\times10}$ directly at
$L=18,36,72,144$ and epochs $e=0,1,10$ for both sizes.
These choices were fixed before this additional extraction from the
completed runs. For $\Delta F=F_{q,2L,e}-F_{q,L,e}$, we report
\begin{equation}
\begin{aligned}
 R&=\left(\frac1{1000}\sum_{a=1}^{100}\sum_{c=1}^{10}
                  \Delta F_{a,c}^2\right)^{1/2},\\
 S&=\frac1{100}\sum_{a=1}^{100}
       \left(\Delta F_{a,y_a}-\frac19\sum_{c\ne y_a}\Delta F_{a,c}\right),
\end{aligned}
\end{equation}
where $y_a$ is the saved test-probe label.
The signed contrast $S$ is a fixed linear test of the output change;
its sign does not determine a change in accuracy or cross-entropy.
The raw-logit discrepancy $R$ also counts offsets common to all classes.

\begin{table}[t]
\centering
\color{black}
\caption{Adjacent regroupings of each coupled pool. All selected cases
are shown; epoch zero records the initial disagreement.}
\label{tab:adjacent-swing}
\small
\begin{tabular}{rr*{3}{rr}}
\toprule
& & \multicolumn{2}{c}{$18\to36$}
  & \multicolumn{2}{c}{$36\to72$}
  & \multicolumn{2}{c}{$72\to144$}\\
$q$ & Epoch & $R$ & $S$ & $R$ & $S$ & $R$ & $S$\\
\midrule
24 & 0  & 0.0173 &  0.0023 & 0.0136 &  0.0002 & 0.0106 &  0.0016\\
24 & 1  & 0.2480 &  0.0628 & 0.2073 & -0.0209 & 0.1866 &  0.0369\\
24 & 10 & 1.1092 & -0.1466 & 1.0145 &  0.1718 & 0.9123 & -0.1930\\
36 & 0  & 0.0124 & -0.0005 & 0.0100 &  0.0002 & 0.0082 & -0.0010\\
36 & 1  & 0.2118 &  0.0927 & 0.1550 & -0.0097 & 0.1498 & -0.0211\\
36 & 10 & 1.4441 &  0.1931 & 1.3345 &  0.2075 & 1.3254 & -0.0408\\
\bottomrule
\end{tabular}
\end{table}

The three successive RMS discrepancies decrease within each size and
epoch, but their scale grows substantially during training
(Table~\ref{tab:adjacent-swing}).
These absolute differences do not separate changes in logit amplitude
from sensitivity to grouping.
At epoch ten, $q=36$ has larger discrepancies than $q=24$, and its final
refinement makes little further difference to their magnitude.
The signed contrasts change direction repeatedly, so a small average
contrast cannot stand in for agreement of all outputs.
These comparisons describe local sensitivity to grouping without choosing
the finest partition as a limiting output. They remain one realization
per size, with dependent pairs, and do not estimate an ensemble bias,
centered fluctuations, or a convergence rate. The data export retains all
24 output arrays and 18 difference arrays.

\subsection{CIFAR-10 growth recipes from a common anchor}
\label{app:cifar-growth-recipes}

The CIFAR panel in Fig.~\ref{fig:scaling-recipes} uses the same flattened
inputs and ten-class cross-entropy as the convergence experiment, but
uses non-affine RMSNorm \citep{zhang2019rmsnorm} before each ReLU branch:
$\widehat h=h/\sqrt{\|h\|^2/N_\flat+10^{-6}}$.
The identity path is unchanged; there are no biases, augmentation,
momentum, weight decay, or learning-rate schedule.
Its anchor has $(N_\flat,K,L)=(181,362,16)$, $\beta_0=0.5625$,
$\gamma_0=1$, and endpoint/residual learning-rate multipliers
$2.56/1.28$ in the stored-Gaussian convention: the actual SGD rates are
$\eta_{\rm end}\gamma_0^2N_\flat$ for input/readout and
$\eta_{\rm res}\gamma_0^2K$ for up/down matrices.
There is no additional $0.02$ factor in this experiment.
Three seeds, $100001$--$100003$, share the minibatch order.
Training uses MPS float32 and batches of 64, dropping the last batch.
Each run completes five epochs (3,905 SGD updates); the final
cross-entropy is evaluated on a fixed 10,000-example training subset.
Figure~\ref{fig:scaling-recipes} reports the mean and sample standard
deviation across these three seeds.

Write the anchor dimensions as $(N_0,K_0,L_0)$.
At budget index $j$, the ideal ODE geometry is
$(N_0 2^{j/3},K_0 2^{2j/3},L_0 2^{j/3})$,
whereas the fixed-ratio geometry is
$(N_0 2^{j/2},K_0 2^{j/2},L_0 2^{j/4})$.
Both have residual-block parameter budget $2N_0K_0 2^j$.
Integer widths are distributed across layers to realize the recorded
total $K$; comparisons use actual budgets, not stage names.
The early ODE/semi-freeze/Half Copy point has
$(N_\flat,K,L)=(362,1448,32)$, whereas the early fixed-ratio point is
$(724,1448,32)$, with twice the block parameters.
The subsequent six stages likewise use different budget indices for
the two geometries; exact dimensions and indices accompany the data.
The extra largest fixed-ratio point is omitted from the main panel
but included in the full-range Fig.~\ref{fig:cifar-growth-full}.
Total trainable parameters are $2N_\flat K+3082N_\flat$.
ODE and the fixed-ratio recipe keep their rate multipliers fixed.
Semi-freeze uses the ODE geometry, multiplies $\beta_0$ by
$\sqrt{R_\alpha}$, and divides only the residual rate multiplier by
$R_\alpha$, where
$R_\alpha=(N_0/K_0)/(N_\flat/K)$; endpoint rates remain fixed.
Half Copy keeps the ODE geometry and rates and uses the layer-local
construction in Appendix~\ref{app:half-copy-recipe}.
Its up-projections are identical within each group; down-projections
have correlation $1/3$, and all parameters remain independently trainable.
With group sizes $c_g$ across all layers, the initial variance
enhancement is $1+\sum_g c_g(c_g-1)/(3K)$.
The integer layer-local partitions target $R_\alpha$;
the supplement records their realized enhancement and group sizes.
The older recipe that also reduces endpoint rates is not used in this
figure.

\begin{figure}[htbp]
\centering
\includegraphics{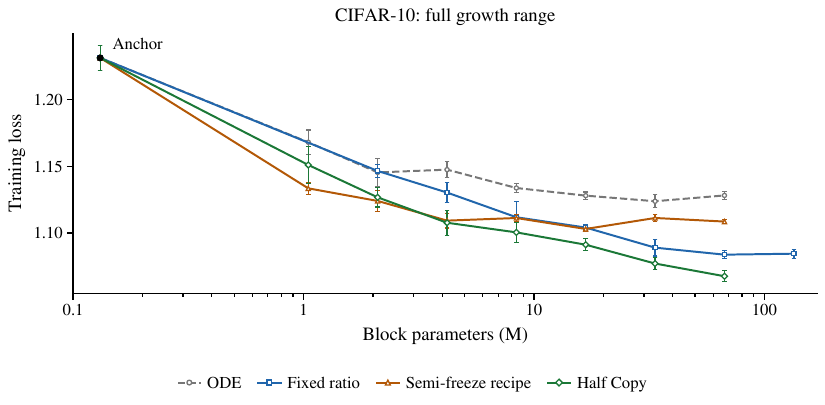}
\caption{CIFAR-10 growth recipes over the full measured range, including
the shared anchor and early points. Final training cross-entropy follows
five SGD epochs; bars are sample standard deviations over three seeds.
All recipes use their actual block parameter counts, including the extra
largest fixed-ratio point omitted from Fig.~\ref{fig:scaling-recipes}.}
\label{fig:cifar-growth-full}
\end{figure}

The full-range curves illustrate the reversal of an initially favorable
semi-freeze comparison; an initial advantage is not required.
In a separate common-anchor comparison with $\beta_0=0.28125$ and
endpoint/residual multipliers $1.28/1.28$, semi-freeze has higher
four-seed mean training loss than the fixed-ratio recipe at all twelve
measured budgets $j=3,\ldots,14$.
The endpoint pairs are $1.19993$ versus $1.19249$ and $1.22033$ versus
$1.14292$, respectively.
This record combines three audited historical seeds with a fourth
seed; one individual seed at the smallest budget slightly favors
semi-freeze.

\paragraph{Learning-rate transfer and local retuning.}
\label{app:cifar-lr-transfer}
A separate 114-run scan tests the older anchor
$\beta_0=0.28125$, $\gamma_0=1$, with
$\eta_{\rm end}=\eta_{\rm res}=1.28m$.
It uses the same five-epoch pre-RMS protocol, three seeds, and
budget indices $j=0,7,14$, but does not reuse the plotted growth runs.
The candidate multipliers are
$m=2^{k/2}$ for $k=-2,\ldots,2$ at $j=0$,
$k=0,\ldots,4$ at $j=7$, and $k=0,\ldots,5$ at $j=14$.
The anchor is shared by ODE, fixed-ratio, and Half Copy recipes.
All selection uses final training loss.

The anchor-selected multiplier is $\sqrt2$.
The grid minimizers at $j=7$ are $2\sqrt2,2,2$ for ODE, fixed ratios,
and Half Copy, respectively; at $j=14$ they are $2\sqrt2,2\sqrt2,2$.
All lie inside their search windows.
Transferring $\sqrt2$ instead of retuning costs $0.008$--$0.016$
in mean training cross-entropy across the six larger-scale comparisons.
Local retuning preserves the ordering Half Copy, fixed ratios, ODE
from lower to higher loss at both sizes
(Fig.~\ref{fig:cifar-lr-landscape}).
These finite grids assess transfer accuracy, not a continuous optimal
learning rate; their anchor and endpoint-rate convention differ from
Fig.~\ref{fig:scaling-recipes}.

The figure also includes an earlier coarse scan of ODE, fixed ratios,
and semi-freeze at the same anchor, seeds, budgets, and five-epoch protocol.
It uses $m=2^k$, $k=-4,\ldots,3$, with 168 physical runs on A100;
the 114-run refinement above was performed on MPS.
The figure shows one curve per recipe, using the refined measurements
where grid points overlap and the earlier scan elsewhere, without pooling seeds.
For semi-freeze, $m$ scales the endpoint rate and the prescribed reduced
residual rate together.
Within the coarse scan, its grid minimum moves from $1$ at the anchor
to $4$ at $j=7$ and $8$ at $j=14$, where it is still at the search boundary.
Transferring its anchor-selected $m=1$ costs $0.052$ and $0.073$ in
mean training loss relative to these measured minima.
Thus semi-freeze provides a case where the prescribed rate scaling
requires appreciably more retuning.

\begin{figure}[t]
\centering
\color{black}
\includegraphics{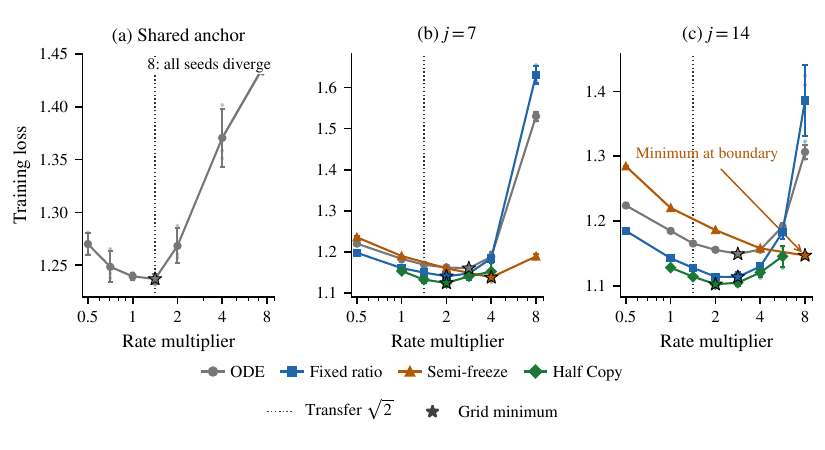}
\par
\includegraphics{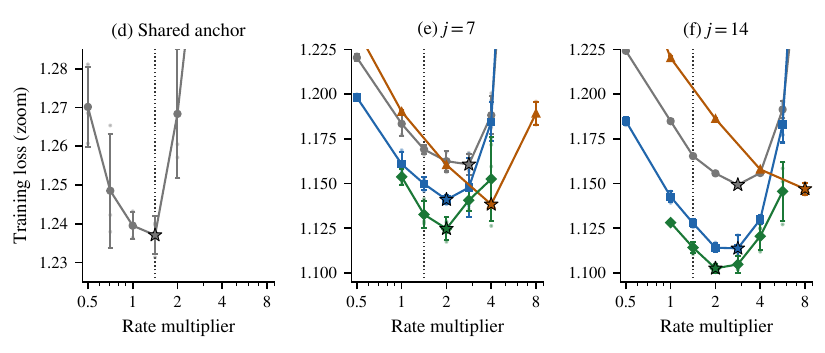}
\caption{CIFAR-10 learning-rate scans, shown for $m\geq1/2$.
Top: scan overview. Bottom: the same data with tighter vertical limits
to resolve the basin bottoms; all six panels use identical horizontal axes.
Colors identify recipes. Small points show three seeds;
bars show sample standard deviations after five epochs.
Stars mark grid minima; the semi-freeze minimum at $j=14$ is not bracketed.
The dotted line marks the anchor-selected $\sqrt2$.
The shared anchor is drawn once. The anchor arrow at $m=8$ marks
three divergent runs, not a finite loss.}
\label{fig:cifar-lr-landscape}
\end{figure}

\par
\endgroup
\begingroup
\color{black}
\section{Tiny Shakespeare Transformer experiments}
\label{app:shakespeare-protocol}

\subsection{Model, training, and evaluation}

The growth experiments use a causal character-level Transformer
\citep{vaswani2017attention} on the Tiny Shakespeare text distributed with
\texttt{char-rnn} \citep{karpathy2015charrnn}, with a 65-character vocabulary
and a contiguous $90/10$ split.
This series evaluates only training windows.
Each of three seeds, $100001$--$100003$, specifies initialization,
training-window sampling, and fixed evaluation windows.
The training and evaluation streams use seed offsets $+1$ and $+2$.
The same seed identifies paired streams across recipes and sizes.
Consequently, variation across seeds includes data sampling as well as
initialization; it is not initialization-only uncertainty.
Each run uses 5,000 Adam updates, effective batch size 32, sequence length
128, and float32 arithmetic, processing 20,480,000 training positions.
Loss is evaluated on 16 fixed batches of 32 training sequences at step 1
and every 500 updates.
All uncertainty reported here is sample standard deviation over three
seeds, not standard error.

Write $n=N_\flat$, let $N_{{\rm b},\ell}$ denote the MLP width and
$N_{{\rm h},\ell}$ the head count in block $\ell$, and set
\[
 K_{\rm b}=\sum_{\ell=0}^{L-1}N_{{\rm b},\ell},
 \qquad K_{\rm h}=\sum_{\ell=0}^{L-1}N_{{\rm h},\ell}.
\]
The head dimension is $d_{\rm h}=16$.
Each block applies causal attention and then a GELU MLP
\citep{hendrycks2016gelu}, with a separate non-affine pre-RMSNorm
\citep{zhang2019rmsnorm} before each branch, using $\epsilon=10^{-6}$.
Queries and keys use RoPE \citep{su2021roformer},
with frequencies $2^{-j}$, $j=0,\ldots,7$.
Attention logits are divided by $\tau\sqrt{d_{\rm h}}$, with $\tau=1$.
There are no biases, dropout, weight decay, gradient clipping, or
learning-rate schedule.
The token embedding and output weights are untied.
A final non-affine RMSNorm precedes the linear readout.
For stored standard-Gaussian weights, all up-maps have factor $1/\sqrt n$,
the MLP residual branch has factor $\beta_{\rm b}\sqrt n/K_{\rm b}$,
the attention output branch has factor
$\beta_{\rm h}\sqrt n/(K_{\rm h}\sqrt{d_{\rm h}})$,
and the readout has factor $1/(\gamma n)$.
Here $\gamma=1$; $d_{\rm h}$ is kept separate from the head-count ratio.

The core and total trainable parameter counts are
\[
 P_{\rm core}=2nK_{\rm b}+4d_{\rm h}nK_{\rm h},
 \qquad P_{\rm total}=P_{\rm core}+2(65)n.
\]
Parameter counts below are realized counts, not nominal targets.
The anchor has $(n,N_{\rm b},N_{\rm h},L)=(128,256,4,4)$,
giving $K_{\rm b}=1024$, $K_{\rm h}=16$, and $P_{\rm core}=393216$.
Its branch coefficients are
$\beta_{\rm b}=1.6771809884309676$ and
$\beta_{\rm h}=0.35617040537656264$.

\subsection{Calibration and eight optimizer groups}

Initial branch energy provides a calibration observable without requiring
a diversity score.
Let $\Delta_{{\rm b},\ell}$ and $\Delta_{{\rm h},\ell}$ be the MLP and
attention additions to the residual stream at one token.
We average the following quantities over tokens and fixed training windows:
\[
 Q_{\rm b}=\sum_\ell\frac{\|\Delta_{{\rm b},\ell}\|^2}{n},
 \quad Q_{\rm h}=\sum_\ell\frac{\|\Delta_{{\rm h},\ell}\|^2}{n},
 \quad Q=Q_{\rm b}+Q_{\rm h},
 \quad S_0=\frac{\|h^0\|^2}{n}.
\]
The anchor calibration uses five seeds and eight batches, with no weight
updates.
A target $Q/S_0=0.3$ gives measured mean $0.325306$ and an
attention-to-MLP energy ratio of $1.16695$.
A stronger target of $3$ gives $5.02973$ and a branch ratio of $2.30879$,
failing the prescribed balance check.
These tests select a usable energy scale, not a performance optimum.
A subsequent training-loss search adjusts bulk, input, and output rates.
The selected anchor has five-seed loss $1.269145\pm0.010729$;
none of the six tested axial neighbors passes the local improvement
criterion.
The endpoint search reaches an output-rate boundary, so this is not a
global hyperparameter optimum.

All experiments in the comparison use Adam \citep{kingma2015adam}
with moments $(0.9,0.999)$
and its usual bias correction.
Table~\ref{tab:shakespeare-adam-groups} specifies the eight groups in
stored Gaussian coordinates.
The constants are
$\eta_{\rm in}=0.064$, $\eta_{\rm out}=0.256$, and
$\eta_{\rm bulk}=0.38159268702343885$.
For each group, the denominator constant is transported as
$\epsilon_j=10^{-8}s_j/s_{j,0}$, where $s_{j,0}$ is its anchor value.
There is no additional factor $128/n$ beyond this ratio.
This deterministic transport is part of the recipe; it does not assert
that every gradient coordinate has identical scaled statistics.

\begin{table}[t]
\centering
\color{black}
\small
\begin{tabular}{lll}
\toprule
Parameter group & Stored-weight Adam rate & Gradient unit $s_j$\\
\midrule
Token embedding & $\eta_{\rm in}\gamma$ & $1/(\gamma n)$\\
Readout & $\eta_{\rm out}\gamma$ & $1/(\gamma n)$\\
MLP up & $\eta_{\rm bulk}\gamma\beta_{\rm b}/\sqrt n$
 & $\beta_{\rm b}/(\gamma K_{\rm b}\sqrt n)$\\
MLP down & $\eta_{\rm bulk}\gamma\beta_{\rm b}/\sqrt n$
 & $\beta_{\rm b}/(\gamma K_{\rm b}\sqrt n)$\\
Attention Q & $\eta_{\rm bulk}\gamma\beta_{\rm h}/\sqrt n$
 & $\beta_{\rm h}/(\gamma K_{\rm h}\sqrt n)$\\
Attention K & $\eta_{\rm bulk}\gamma\beta_{\rm h}/\sqrt n$
 & $\beta_{\rm h}/(\gamma K_{\rm h}\sqrt n)$\\
Attention V & $\eta_{\rm bulk}\gamma\beta_{\rm h}/\sqrt n$
 & $\beta_{\rm h}/(\gamma K_{\rm h}\sqrt n)$\\
Attention output & $\eta_{\rm bulk}\gamma\beta_{\rm h}/\sqrt n$
 & $\beta_{\rm h}/(\gamma K_{\rm h}\sqrt n)$\\
\bottomrule
\end{tabular}
\caption{The eight Adam groups. The compensated-coupling recipe changes the
bulk constants and branch coefficients as specified below.}
\label{tab:shakespeare-adam-groups}
\end{table}

The ratios $\epsilon_j/\sqrt{\widehat v_j}$ are recorded at updates
1, 100, 1000, and 5000 on up to 8192 stratified coordinates per group.
This check motivated replacing an earlier fixed-$\epsilon$ continuation;
its results are not pooled with the transported-$\epsilon$ series here.
Earlier development runs also used learned positional embeddings and
held-out evaluation; they are a different protocol.

\subsection{Growth recipes and correlated initialization}

The main growth series uses $R=4,5,6.25,8,10,12.75$.
The ODE route has
$(n,N_{\rm b},N_{\rm h},L)=(128R,256R,4R,4R)$,
with fixed branch and rate constants.
Thus $n/K_{\rm b}=1/(8R)$ and $n/K_{\rm h}=8/R$.
For the fixed-ratio route, let $q_R$ be the nearest integer to
$16R^{3/2}$ and choose
$(n,K_{\rm b},K_{\rm h},L)=(8q_R,64q_R,q_R,4R)$.
Balanced floor/ceiling allocations preserve these totals when per-block
counts are not integers.
This keeps $(n/K_{\rm b},n/K_{\rm h})=(1/8,8)$.
Compensated coupling keeps the ODE geometry but multiplies both branch
coefficients by $\sqrt R$ and divides the six bulk $\eta$ constants by $R$;
endpoint constants stay fixed.
Its initialization compensates the falling $\beta^2n/K$ scale.
These experiments do not directly measure whether individual features
freeze, and we make no such inference from the recipe's construction.

Half Copy keeps the ODE geometry, branch coefficients, and rates,
using the construction in Appendix~\ref{app:half-copy-recipe}.
The growth series fixes $\rho=1/3$ and uses layer-local balanced groups
targeting $\chi=R$ separately for MLP units and attention heads.
For each family, its reported enhancement is the channel-weighted
average of Eq.~\ref{eq:half-copy-realized-enhancement} over layers.
Integer group sizes cause small departures: at $R=6.25$, the realized
uniform attention enhancement is $6.1712$.

Figure~\ref{fig:transformer-growth-full} also includes the common $R=1$ anchor
and three earlier $R=2$ points, all with the same three seeds and
5,000-update protocol.
At $R=2$, ODE and semi-freeze use
$(n,K_{\rm b},K_{\rm h},L)=(256,4096,64,8)$;
the fixed-ratio point uses $(512,4096,64,8)$.
Thus this earlier fixed-ratio point has twice their block parameters;
it is plotted at its actual budget, not treated as an equal-budget
instance of the allocation above. There is no $R=2$ Half Copy endpoint.
The anchor's initial $Q$ is reconstructed using the pinned original
model code and evaluation windows, without training.

\begin{table}[t]
\centering
\color{black}
\small
\begin{tabular}{r r r r r r r r}
\toprule
& & \multicolumn{3}{c}{ODE / compensated / Half Copy}
& \multicolumn{3}{c}{Fixed ratios}\\
$R$ & $L$ & $n$ & $K_{\rm h}$ & $P_{\rm core}$ (M)
& $n$ & $K_{\rm h}$ & $P_{\rm core}$ (M)\\
\midrule
4 &16 &512 &256 &25.165824 &1024 &128 &25.165824\\
5 &20 &640 &400 &49.152000 &1432 &179 &49.214976\\
6.25 &25 &800 &625 &96.000000 &2000 &250 &96.000000\\
8 &32 &1024 &1024 &201.326592 &2896 &362 &201.283584\\
10 &40 &1280 &1600 &393.216000 &4048 &506 &393.271296\\
12.75 &51 &1632 &2601 &815.007744 &5824 &728 &814.055424\\
\bottomrule
\end{tabular}
\caption{Actual growth geometry and core parameter counts.
Every row has $K_{\rm b}=64K_{\rm h}$.
Add $130n$ for total parameters; the two geometry routes therefore need
not have identical total parameter counts even when their core counts match.}
\label{tab:shakespeare-budgets}
\end{table}

All 72 runs in this six-scale, four-recipe, three-seed comparison complete
5,000 updates.
The initial-energy comparison includes the anchor and early points
as well as the five main scales through $R=10$;
we do not supply an initial-energy value for $R=12.75$.
Initial $Q$ is a sum of squared increments, not the squared total residual
displacement.
For all branch increments collected as $\Delta_j$,
$\|\sum_j\Delta_j\|^2/n=Q+
\sum_{j\ne k}\langle\Delta_j,\Delta_k\rangle/n$.
Accordingly, $Q$ does not measure cross-layer covariance or all sources of
feature diversity.

\begin{figure}[htbp]
\centering
\includegraphics{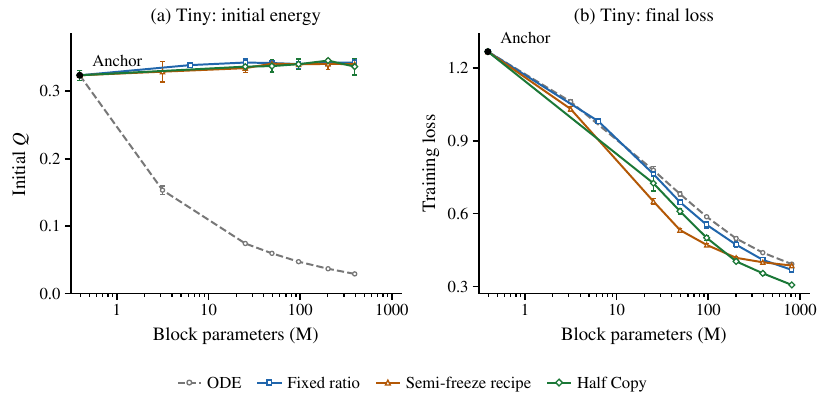}
\caption{Tiny Shakespeare growth recipes over the full measured range,
including the shared anchor and available early points. Left: initial
summed squared residual increments $Q$. Right: training loss after 5,000
Adam updates. Bars are sample standard deviations over three seeds.
Half Copy has no $R=2$ run; losses include $R=12.75$, where initial $Q$
was not measured. Figure~\ref{fig:scaling-recipes} focuses on $R\ge4$.}
\label{fig:transformer-growth-full}
\end{figure}

\subsection{Cross-task parameter savings at comparable training loss}
\label{app:parameter-savings}

The parameter savings quoted in the Introduction compare the final
three-seed mean training losses in Fig.~\ref{fig:scaling-recipes},
after five SGD epochs for CIFAR-10 and 5,000 Adam updates for Tiny
Shakespeare. We count all trainable parameters, including input and
output maps: $P=2N_\flat K+3082N_\flat$ for CIFAR-10 and
$P=P_{\rm core}+130n$ for Tiny Shakespeare.
For an ODE reference with count $P_{\rm ODE}$, the saving is
$1-P_{\rm alt}/P_{\rm ODE}$, where the alternative matches or improves
its mean loss under the same task-specific training schedule.

For Tiny Shakespeare SDE, the matching sizes lie between sampled
networks. With neighboring counts $P_a<P_b$ and mean losses
$\bar\ell_a>\bar\ell_{\rm ODE}>\bar\ell_b$, we interpolate locally:
\begin{equation}
\begin{aligned}
w&=\frac{\log\bar\ell_a-\log\bar\ell_{\rm ODE}}
        {\log\bar\ell_a-\log\bar\ell_b},
&P_{\rm alt}&=P_a^{1-w}P_b^w.
\end{aligned}
\end{equation}
No extrapolation is used.
At ODE reference scales $R=10$ and $12.75$, this gives savings of
$28.1\%$ and $34.7\%$, respectively.
Interpolating loss linearly against either $\log P$ or $P$ instead
gives $27.2\%$ or $23.1\%$ at $R=10$, and $34.1\%$ or $29.6\%$
at $R=12.75$.
For CIFAR-10, a direct comparison needs no interpolation:
SDE uses $12.850$M parameters with mean loss $1.1119$, versus ODE's
$19.581$M and $1.1281$, a $34.4\%$ reduction.

Half Copy also admits direct comparisons on both tasks.
On CIFAR-10, $10.618$M parameters give loss $1.1005$, below the same
$19.581$M ODE reference, using $45.8\%$ fewer parameters.
On Tiny Shakespeare, Half Copy at $R=10$ uses $393.382$M parameters
with loss $0.3539$, compared with $815.220$M and $0.3916$ for ODE at
$R=12.75$, using $51.7\%$ fewer parameters.
All three paired seeds favor the smaller network in each of these
direct comparisons, including the CIFAR SDE comparison.
The introductory $20\%$ and $40\%$ figures round down these estimates
and comparisons at the specified larger reference sizes.
They describe final mean training loss under these schedules, not
confidence bounds or guarantees at every size or seed; for example,
one paired Tiny SDE estimate at $R=10$ is $18.8\%$ under linear
loss--parameter interpolation.

\subsection{Width--depth allocation at fixed capacity}
\label{app:shakespeare-depth}

\paragraph{One to four blocks at fixed capacity.}
A separate Half Copy scan probes shallower allocations at two fixed
budgets. For $R=4$ and $8$, respectively,
$(n,K_{\rm b},K_{\rm h})=(512,16384,256)$ and
$(1024,65536,1024)$, with head dimension $16$.
The total parameter counts are $25{,}232{,}384$ and $201{,}459{,}712$.
All runs use the same Tiny Shakespeare split and Adam recipe described
above: $5{,}000$ updates, sequence length $128$, effective batch $32$,
and seed $100001$. Final cross-entropy averages $512$ fixed training
sequences, not held-out data.
MLP units and attention heads are allocated evenly across blocks, up to
integer rounding. This scan rebuilds the original layer-local
correlated groups at every depth, with $\rho=1/3$.
It therefore changes their joint initialization law as well as depth.
Among the seven completed points, fewer blocks give lower training loss
at each budget; each point uses a single seed.
The original $16$-block model at $R=4$ uses the same layer-local rule,
parameter budget, and training protocol. Its seed-$100001$ loss is
$0.724028$, completing the $1,2,4,16$-layer comparison in
Fig.~\ref{fig:width-depth-allocation}.
These four plotted values evaluate the terminal model on the fixed
training windows; they are not individual optimization-minibatch losses.

\begin{table}[t]
\centering
\color{black}
\begin{tabular}{crrrr}
\toprule
$R$ & $L=1$ & $L=2$ & $L=3$ & $L=4$ \\
\midrule
$4$ & $0.426116$ & $0.595115$ & $0.677284$ & $0.689741$ \\
$8$ & $0.257038$ & $0.339469$ & --- & $0.382781$ \\
\bottomrule
\end{tabular}
\caption{Final fixed-training-window cross-entropy in the separate
low-depth scan. The dash denotes no completed endpoint in this evidence
set.}
\label{tab:shakespeare-low-depth}
\end{table}

\paragraph{Paired four-block and deeper controls.}
For each recipe at $R=4,5,6.25,8,10$, a four-block model preserves
$n,K_{\rm b},K_{\rm h}$, both parameter counts, and the training protocol of
its original $L=4R$ counterpart.
All 60 paired comparisons yield lower final training loss with four blocks
(Fig.~\ref{fig:transformer-paired-depth}).
Table~\ref{tab:shakespeare-depth} gives the largest paired scale.
The networks are initialized and trained separately.
For Half Copy, regrouping can also move correlated groups across layer
boundaries; its comparison does not isolate depth from all changes in the
joint initialization law.
The other three recipes retain independent Gaussian initialization.

\begin{figure}[t]
\centering
\color{black}
\includegraphics{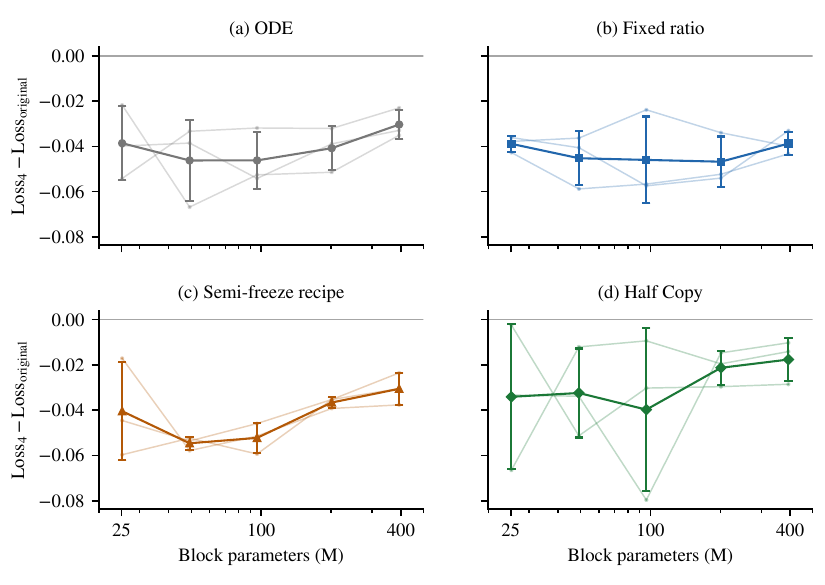}
\caption{Paired depth comparisons on Tiny Shakespeare at five budgets
($R=4,5,6.25,8,10$). Each value is four-block minus original-depth final
training loss. Faint lines show the three paired seeds; thick lines and
bars show their mean and sample standard deviation. All sixty differences
are negative. The separate $R=12.75$ extension is not included.}
\label{fig:transformer-paired-depth}
\end{figure}

\begin{table}[t]
\centering
\color{black}
\small
\begin{tabular}{lrrr}
\toprule
Recipe & $L=40$ & $L=4$ & $R=12.75,\ L=51$\\
\midrule
ODE route & $0.4389\pm0.0003$ & $0.4085\pm0.0067$ & $0.3916\pm0.0010$\\
Fixed ratios & $0.4092\pm0.0058$ & $0.3704\pm0.0085$ & $0.3680\pm0.0025$\\
Compensated coupling & $0.3993\pm0.0054$ & $0.3688\pm0.0059$ & $0.3857\pm0.0025$\\
Half Copy & $0.3539\pm0.0069$ & $0.3362\pm0.0082$ & $0.3059\pm0.0038$\\
\bottomrule
\end{tabular}
\caption{Final training loss, mean and sample standard deviation over
three seeds. The first two columns are paired at $R=10$;
the last column is a larger growth experiment, not part of that pair.}
\label{tab:shakespeare-depth}
\end{table}

Additional deeper allocations at $R=4,5,6.25$ comprise 36 completed runs.
At $R=5$, increasing depth from 20 to 100 raises mean loss from
$0.6805$ to $0.6889$ for the ODE route, from $0.6469$ to $0.6555$ for
fixed ratios, from $0.5319$ to $0.5526$ for compensated coupling,
and from $0.6096$ to $0.6324$ for Half Copy.
The fixed-ratio deeper model here has $1.12\%$ more core parameters due
to rounding; at $R=6.25$ the discrepancy is $-1.59\%$.
The direction is not universal: compensated coupling at $R=4$ gives
$0.6486$ for the deeper allocation and $0.6503$ for the original.
These finite controls support a task- and recipe-specific depth effect,
not a universal ordering by depth.

\subsection{Depth-integrated residual kernels}
\label{app:shakespeare-kernels}

Depth-integrated residual kernels supply a separate diagnostic of the
trained networks. These measurements use the 5,000-update checkpoint:
the integration is over network depth, not training time.
On 16 common training texts at token positions 31, 63, 95, and 127, let
$H^\ell$ be the $64\times n$ residual matrix after block $\ell$ and before
the next normalization.
For each seed we form the raw kernels
\[
 \Phi^\ell=\frac{H^\ell(H^\ell)^\mathsf{T}}n,\qquad
 \overline\Phi=\frac1L\sum_{\ell=1}^L\Phi^\ell.
\]
No diagonal normalization or division by branch strength is applied.
The input kernel is recorded separately.
Across the three seeds, relative Frobenius dispersion of
$\overline\Phi$ decreases from $14.9\%$ to $10.1\%$ between $R=4$ and
$R=10$ for the ODE route, from $15.0\%$ to $10.6\%$ for fixed ratios,
from $13.4\%$ to $8.9\%$ for compensated coupling, and from $17.3\%$ to
$15.3\%$ for Half Copy.
Here relative dispersion is
$\sqrt{\mathbb E_s\|\overline\Phi_s-\mathbb E_s\overline\Phi_s\|_F^2}
/\|\mathbb E_s\overline\Phi_s\|_F$, with the numerator estimated using
sample variance.
Cross-layer covariance accounts for approximately $93$--$95\%$ of the
entrywise variance at $R=10$.
Layers cannot be treated as independent replications, and three seeds
do not establish a fluctuation exponent.
Figure~\ref{fig:transformer-trained-kernels} also retains kernel amplitude
through the mean diagonal. Half Copy has the lowest training loss among
these recipes at $R=10$ but the largest relative kernel dispersion;
the two statistics therefore give different recipe rankings.
Across-seed dispersion is distinct from random variation along depth
within a network.

\begin{figure}[t]
\centering
\color{black}
\includegraphics{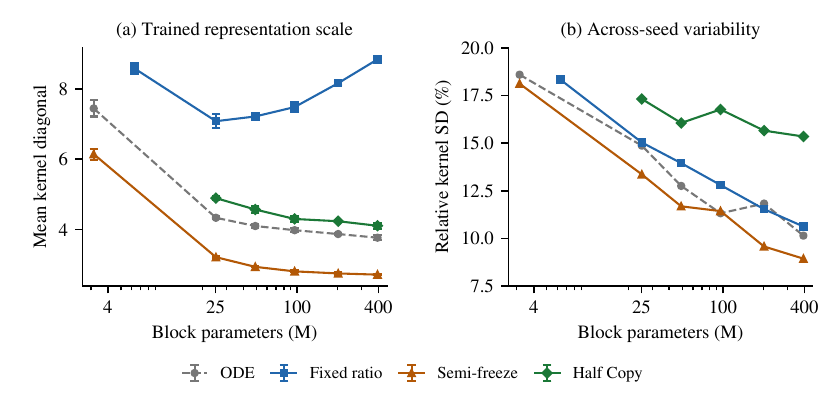}
\caption{Raw residual kernels after 5,000 Tiny Shakespeare updates.
Left: mean diagonal of the depth-integrated kernel, with three-seed sample
standard deviations. Right: relative Frobenius dispersion across those
seeds. For the integer grouping used in this experimental series,
Half Copy measurements begin at $R=4$.}
\label{fig:transformer-trained-kernels}
\end{figure}

\subsection{Exploratory nonuniform allocations}
\label{app:shakespeare-adaptive}

Nonuniform width and nonuniform depth meshes implement different finite
networks.
The width experiment keeps $n,L,K_{\rm b},K_{\rm h}$ fixed while allocating
more MLP units and heads to selected depths.
Writing $d_\ell=N_{{\rm h},\ell}/(K_{\rm h}/L)$, with MLP counts
proportional to head counts, the local residual branch and its
gradient-unit epsilon are multiplied by $1/d_\ell$.
Stored-weight learning rates are unchanged.
The depth-mesh experiment instead keeps per-block capacity fixed and
uses intervals $\delta_\ell$ summing to one, multiplying the same branch
and epsilon factors by $L\delta_\ell$.
For Half Copy, maintaining the corresponding initialization variance
can exceed what layer-local head groups can supply.
At $R=8$, the finest requested interval requires enhancement $27.6283$,
whereas 32 heads with $\rho=1/3$ allow at most $1+31/3=11.3333$
within one layer. The implemented mesh therefore allows groups across
two or three adjacent layers and matches the weighted covariance over
each such group of layers; MLP groups remain layer-local.
This exploratory construction differs from the uniformly spaced,
layer-local growth recipes above.
Table~\ref{tab:shakespeare-adaptive} reports completed training losses;
each row is a single seed, not a three-seed estimate.

\begin{table}[t]
\centering
\color{black}
\small
\begin{tabular}{rrlrrr}
\toprule
$R$ & Seed & Recipe & Uniform & Width allocation & Depth mesh\\
\midrule
8 &100001 &ODE &0.506104 &0.477521 &---\\
8 &100001 &Half Copy &0.397821 &0.380109 &---\\
8 &100002 &ODE &--- &0.483269 &0.488810\\
8 &100002 &Half Copy &--- &0.402959 &0.388330\\
10 &100001 &ODE &--- &0.432979 &0.435738\\
10 &100001 &Half Copy &--- &0.339870 &0.337943\\
\bottomrule
\end{tabular}
\caption{Exploratory nonuniform allocations, each trained for 5,000 updates.
Dashes indicate comparisons not included here, not failed training.
The second $R=8$ seed contributed to allocation selection.}
\label{tab:shakespeare-adaptive}
\end{table}

These allocations were guided by measured activity profiles, not an
identified SDE drift.
A centered-window estimator initially used to interpret those profiles
failed an Ornstein--Uhlenbeck control: it returned approximately
$-0.840$ for a window-averaged squared drift of $0.854$.
Its near-correct constant-drift control therefore did not validate that
interpretation.
The trained models and losses remain valid finite experiments, but the
allocation rule is an exploratory heuristic.
Neither these selected runs nor the branch-energy profiles establish a
general benefit from moving channels toward any particular depth.
\endgroup
\begingroup
\color{black}
\section{GPT-2 protocols and residual measurements}
\label{app:gpt2-protocol}

The AdamW shape and energy experiments and the all-NS5 Muon experiment
use distinct architectures and training protocols, detailed below.
They are not a matched optimizer comparison.

\paragraph{Sources of the theoretical illustration.}
\label{app:gpt2-illustration}
Figure~\ref{fig:gpt2-theory-illustration} combines selected examples
from these two experiments.
On the left, coordinate 6 of the twenty-four-stage AdamW model is
tracked at all 25 residual boundaries for validation text 0,
token 1023, at initialization and after 15,000 updates.
On the right, the all-NS5 model at update 4500 pairs validation
texts 3 and 6 at block 1, token 1023.
All 768 post-block residual coordinates and all 3072 pre-GELU hidden
coordinates are retained.
Each marginal is centered and standardized separately, without
rotation, whitening, or tail removal.
The examples were selected for qualitative illustration; the systematic
distributional measurements below retain the predefined texts and slices.
The covariance-matched Gaussian ellipses are 95\% contours, not
confidence intervals or contours forced to enclose 95\% of the data.

\subsection{AdamW models and finite regrouping}
\label{app:gpt2-shape}

The base implementation of GPT-2 \citep{radford2019language} is
\texttt{build-nanogpt} \citep{karpathy2024buildnanogpt}, commit
\texttt{6104ab1b53920f6e2159}\allowbreak\texttt{749676073ff7d815c1fa}.
It has twelve attention--MLP blocks with pre-LayerNorm
\citep{ba2016layernorm}, residual width 768,
twelve heads of dimension 64, MLP width 3072, approximate GELU, learned
positional embeddings, and a final LayerNorm. Biases and affine
LayerNorm parameters are retained. The token embedding and output
matrix are tied; the vocabulary is padded to 50,304 entries.
There is no dropout. All five constructions have 124,475,904 parameters.

The one-, two-, and four-stage constructions group 12, 6, and 3
consecutive original blocks, respectively. Within each group, the
attention branches, each with its own LayerNorm, read the same incoming
state. Their outputs are summed and added once. The MLP branches then
read this updated state and their outputs are summed and added once.
No averaging coefficient is introduced. The twenty-four-stage model
instead divides each original block into two serial attention--MLP
stages, each with six heads and 1536 MLP units. Matrix slices retain
their original parameters; the stages share their parent block's
LayerNorm parameters, and each contributes half its original
down-projection bias. Up-projection biases are partitioned.
Parameter tensors, optimizer groups, and numerical initial arrays
are unchanged, although the initial functions differ.

All runs use seed 1337 and retain the initialization and optimizer
settings of \citet{karpathy2024buildnanogpt}.
Embeddings and ordinary linear weights have
Gaussian standard deviation 0.02; attention and MLP down-projections
use $0.02/\sqrt{24}$ in every construction. Biases begin at zero and
LayerNorm gains at one. Fused AdamW \citep{loshchilov2019decoupled} uses
$(\beta_1,\beta_2)=(0.9,0.95)$, $\epsilon=10^{-8}$, and weight decay
0.1 on matrices and embeddings, with no decay on biases or LayerNorm.
The gradient norm is clipped at 1. Learning rate increases linearly
over 715 updates to $6\times10^{-4}$, then follows the original cosine
schedule toward $6\times10^{-5}$.
Each update accumulates eight microbatches of 64 sequences of length
1024: 524,288 tokens. The one-, twelve-, and twenty-four-stage runs use
one A100-SXM4-80GB. The two-stage run resumes from its 10,341-update
A100 checkpoint on an H100; the four-stage run uses an H100 throughout.
All use PyTorch 2.8.0, the CUDA 12.8 build, and bfloat16 autocasting,
without \texttt{torch.compile}.

The data are \texttt{HuggingFaceFW/fineweb-edu}
\citep{lozhkov2024finewebedu}, configuration
\texttt{sample-10BT}, revision
\texttt{87f09149ef4734204d70ed1d046ddc9ca3f2b8f9}.
Ordered preprocessing uses the GPT-2 tokenizer
(\href{https://github.com/openai/tiktoken}{\texttt{tiktoken}} 0.11.0),
inserting an end-of-text token before each
document. The first 100-million-token shard is held out; the remaining
99 contain 9,853,989,344 training tokens. Identical shard checksums and
sequential loaders give the same training sequence, cycling when it
is exhausted. Each run completes 19,073 updates, consuming
9,999,745,024 tokens. Validation precedes the final update, so the
reported endpoint is after 19,072 updates and 9,999,220,736 tokens.
Every evaluation resets to the same held-out prefix and averages
twenty microbatches, or 1,310,720 tokens.

Figure~\ref{fig:width-depth-allocation} instead reports training loss,
averaged over the final 100 optimization updates in every construction
(zero-based log steps 18,973--19,072).
Each logged value averages the eight gradient-accumulation microbatches
before the corresponding parameter update; the plotted mean therefore
covers 52,428,800 training tokens along the last 100 updates, rather
than evaluating one frozen terminal model.
For 1, 2, 4, 12, and 24 stages, these means are respectively
3.588404, 3.219410, 3.120185, 3.062130, and 3.049850.
Both panels thus use training data, but the Tiny Shakespeare panel uses
the fixed-window terminal evaluation described in
Appendix~\ref{app:shakespeare-depth}; their estimators and tokenizations differ.
The held-out metrics in Table~\ref{tab:gpt2-shape} retain their original
validation definition. Figure~\ref{fig:gpt2-depth-validation} shows all
recorded validation evaluations, including initialization, with a second
panel resolving the later training differences.

\begin{figure}[t]
\centering
\color{black}
\includegraphics{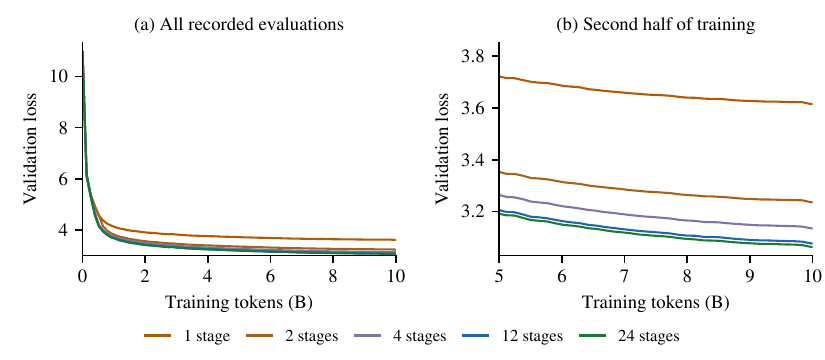}
\caption{Complete validation trajectories for the five AdamW depth
allocations, all with 124,475,904 parameters and seed 1337.
Left: all 78 recorded evaluations per model, including initialization.
Right: an enlarged view of the second half of training.
Lines connect the saved evaluations without smoothing; values retain
the four-decimal precision of the logs.}
\label{fig:gpt2-depth-validation}
\end{figure}
HellaSwag uses all 10,042 validation questions without demonstrations
\citep{zellers2019hellaswag}. The candidate ending with the lowest mean
autoregressive token loss over its completion is selected.
The CE and HellaSwag orderings of the twelve- and twenty-four-stage
models differ; neither single-seed ordering establishes a significant
advantage.

\begin{table}[!ht]
\centering
\color{black}
\begin{tabular}{lrrr}
\toprule
Construction & Stages & Validation CE & HellaSwag\\
\midrule
Parallel regrouping & 1 & 3.6148 & 0.2808\\
Parallel regrouping & 2 & 3.2359 & 0.2927\\
Parallel regrouping & 4 & 3.1347 & 0.2942\\
Original & 12 & 3.0764 & 0.3091\\
Split blocks & 24 & 3.0627 & 0.3061\\
\bottomrule
\end{tabular}
\caption{Completed GPT-2 comparisons at the common evaluation endpoint.}
\label{tab:gpt2-shape}
\end{table}

\subsection{Energy in common residual coordinates}
\label{app:gpt2-energy}

We evaluate four held-out 1024-token sequences beginning at offsets
0, one million, two million, and three million in the validation shard.
Forward hooks record the true residual after every complete
attention--MLP stage, before the next normalization, in CPU float32.
The twenty-four-stage model has 24 measured stages despite retaining
twelve parameter containers. Initialization is reconstructed in the
original PyTorch environment and its parameter-array hash checked
against the training initialization.

For each token, use $N_\flat=768$ and
$\Delta\bm h_\ell=\bm h_{\ell+1}-\bm h_\ell$.
Define $E_0=\|\bm h_0\|^2/N_\flat$,
$Q=\sum_\ell\|\Delta\bm h_\ell\|^2/N_\flat$, and
$C=\|\sum_\ell\Delta\bm h_\ell\|^2/N_\flat-Q$.
The full accounting is
\begin{equation}
\label{eq:gpt2-energy-accounting}
\frac{\|\bm h_L\|^2}{N_\flat}
=E_0+Q+C+
\frac{2\langle\bm h_0,\sum_\ell\Delta\bm h_\ell\rangle}{N_\flat}.
\end{equation}
We compute the quadratic quantities token by token before averaging.
Write $B_0=2\langle\bm h_0,\sum_\ell\Delta\bm h_\ell\rangle/N_\flat$;
the identity is $E_L=E_0+Q+C+B_0$.
The 4096 tokens are descriptive probes, not independent replicates.

\begin{table}[!ht]
\centering
\color{black}
\setlength{\tabcolsep}{4pt}
\begin{tabular}{rrrrrr}
\toprule
Stages & Updates & $E_0$ & $Q$ & $C$ & $B_0$\\
\midrule
12 & 0 & 0.000801 & 0.08099 & $-0.001308$ & $-0.00000351$\\
12 & 15,000 & 0.001385 & 7.84837 & 1.85463 & 0.006261\\
24 & 0 & 0.000801 & 0.08406 & 0.000121 & 0.00000726\\
24 & 15,000 & 0.001394 & 6.40892 & 3.91831 & 0.006519\\
24 & 19,072 & 0.001338 & 5.63515 & 2.70808 & 0.005173\\
\bottomrule
\end{tabular}
\caption{Mean energy components in the same residual coordinates.
The full-precision components satisfy the accounting identity to
less than $10^{-13}$ in the saved aggregates.}
\label{tab:gpt2-energy}
\end{table}

Positive $C$ records aggregate alignment across depth; $Q$ includes
both directed and random increments. We do not identify either with
an independently estimated SDE coefficient. Additional saved profiles
at updates 1000, 2000, and 3000 describe the development of early and
late activity. Local-window comparisons use equal normalized depth
widths, for example two of twelve versus four of twenty-four stages.
Earlier drift/diffusion labels refer to finite-window proxies requiring
extra local assumptions, not direct drift/diffusion identification.

\paragraph{Layer pairs and observation granularity.}
\label{app:gpt2-layer-pairs}
Motivated by the layer-increment angle measurements of
\citet{liu2026inversedepth}, we retain both increment magnitudes and
all layer-pair inner products.
On the same saved trajectories, form the uncentered increment Gram
$G_{\ell m}=\langle\Delta\bm h_\ell,\Delta\bm h_m\rangle/N_\flat$
for each token. Then $Q=\operatorname{tr}G$ and
$C=\sum_{\ell\ne m}G_{\ell m}$.
At update 15,000, the mean adjacent-increment cosine is similar for
twelve and twenty-four stages, $0.1764$ and $0.1838$, whereas
$C/Q$ is $0.2363$ and $0.6114$.
Adjacent pairs contribute $2.63068$ and $2.84889$ to $C$;
more distant pairs contribute $-0.77605$ and $1.06942$.
Local alignment alone therefore does not determine the accumulated
cross-depth contribution.

Combining consecutive observed increments into six equal depth windows
gives the following accounting at the same checkpoints:
\begin{center}
\begin{tabular}{rrrrr}
\toprule
Stages & Stages per window & $Q_{\rm window}$ & $C_{\rm window}$ & $D$\\
\midrule
12 & 2 & 9.76888 & $-0.06589$ & 9.70300\\
24 & 4 & 10.51574 & $-0.18851$ & 10.32723\\
\bottomrule
\end{tabular}
\end{center}
Here $D=\|\bm h_L-\bm h_0\|^2/N_\flat=Q+C$ is unchanged.
Within-window cross terms move into $Q_{\rm window}$, leaving
$C_{\rm window}$ negative in both cases.
This is a rebinning of fixed trajectories, not a sliding intervention
on the network. It illustrates why the observation granularity must
accompany $Q/C$ comparisons. Figure~\ref{fig:gpt2-layer-pairs} shows the
full lag contributions and their cumulative sums, alongside the change
in $C/D$ under these observation windows.

\begin{figure}[t]
\centering
\color{black}
\includegraphics{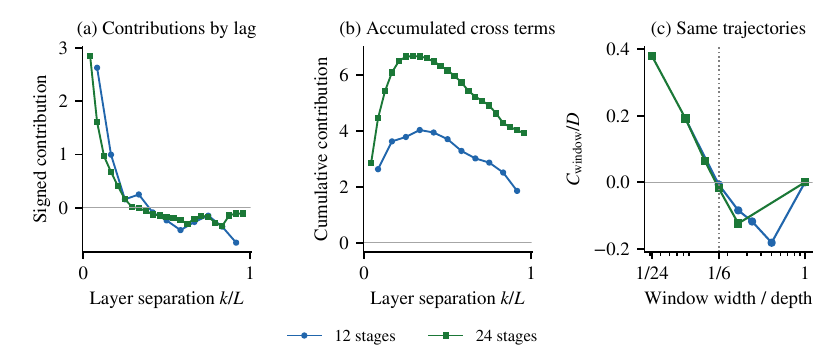}
\caption{Cross-depth contributions at update 15,000 in the AdamW models.
Left: the signed lag contribution $2\sum_\ell G_{\ell,\ell+k}$,
averaged over the four probe texts and their tokens; these are
magnitude-weighted inner products, not average cosines.
Middle: its cumulative sum, ending at $C$.
Right: regrouping the same increments into equal observation windows
changes $C_{\rm window}/D$ while keeping squared residual displacement $D$ fixed.
The dotted line marks six windows, each spanning $1/6$ of the depth.}
\label{fig:gpt2-layer-pairs}
\end{figure}

\subsection{A complete run with NS5 Muon on every matrix}
\label{app:gpt2-ns5}

This experiment uses a different GPT-2-style architecture, adapted from
the \texttt{modded-nanogpt} record of 2024-10-10
\citep{jordan2024moddednanogpt}
(commit \texttt{bc3a0c2d640d0d73dedaef87eae26148d2e32afb}).
The model has 123,532,032 parameters: twelve blocks, width 768,
twelve 64-dimensional heads, and MLP width 3072.
It uses GELU, rotary positional embeddings, non-affine RMSNorm with
stabilizer $10^{-6}$, no biases or dropout, and a tied
$50257\times768$ embedding/readout.
Each attention residual branch is multiplied by $1/\sqrt{24}$;
the MLP residual branch has multiplier one.
Every matrix with fan-in $d$ is independently Gaussian-initialized
with standard deviation $1/\sqrt{3d}$, except that the tied endpoint
matrix is initialized only once.

All 49 parameter matrices use Muon \citep{jordan2024muon},
with the five-step Newton--Schulz polynomial implementation from that record.
If $D_t$ is the current gradient,
the momentum and Nesterov direction are
$M_t=0.95M_{t-1}+D_t$ and $G_t=D_t+0.95M_t$, with zero initial
momentum. Normalize $X_0=G_t/(\|G_t\|_{\rm F}+10^{-7})$ and apply
five iterations
\begin{equation}
X_{j+1}=3.4445X_j-4.7750X_jX_j^{\mathsf T}X_j
 +2.0315(X_jX_j^{\mathsf T})^2X_j.
\end{equation}
The Frobenius norm is computed before conversion to bfloat16;
the iteration uses bfloat16. Tall matrices are transposed during the
iteration and restored afterward. For an $a\times b$ matrix,
the parameter increment is $-\eta_t\sqrt{\max(a,b)}X_5$.
The combined Q/K/V matrix is split into its three $768\times768$
parts before normalization and iteration; each uses multiplier
$\sqrt{768}$. The tied endpoint receives its combined input/readout
gradient once, with multiplier $\sqrt{50257}$.
This NS5 map is distinct from both exact spectral clipping and the
Taylor polynomial used in the logic experiment.

Base rates are 0.00036 for internal matrices and 0.00018 for the
tied endpoint. For $t$ completed updates, the next update multiplies
both rates by $\min\{1,(6200-t)/1800\}$, for $t=0,\ldots,6199$.
There is no warmup, weight decay, gradient clipping, or Adam group.
The endpoint rate was selected by validation loss in a prior
713-update, context-512 pilot and then fixed; Gaussianity plots did
not enter selection. An additional seed was checked in that pilot,
but the full run has only seed 1337.

Training uses 6,200 updates at 524,288 tokens per update, context
1024, and sixteen microbatches of 32 sequences. Parameters are float32;
training autocasting is bfloat16. The run uses PyTorch 2.8.0/CUDA 12.8
on one H100 80GB, with \texttt{torch.compile}.
Data use the GPT-2-tokenized FineWeb-Edu release
\citep{lozhkov2024finewebedu}
\href{https://huggingface.co/datasets/kjj0/finewebedu10B-gpt2}{\texttt{kjj0/finewebedu10B-gpt2}},
pinned to revision
\texttt{440d9d739f970c5d80492e12e81d63d60622ca30}.
Thirty-three ordered training shards each contain 100 million tokens;
a separate shard supplies validation. The cursor crosses shard
boundaries without cycling. Validation uses a fixed prefix of
262,144 tokens, excluded from training.
The final loss 3.083088636 is measured after all 6,200 updates.
This corpus preparation and evaluation set are not assumed identical
to those of the historical AdamW runs.

\paragraph{Distribution measurements.}
Before training, the measurement updates are fixed as
\[
0,1,5,10,25,50,100,250,500,750,1000,1500,2000,3000,4000,5000,6200.
\]
At each time, eight validation sequences start at offsets
$0,8192,\ldots,57344$. We record positions 255 and 1023 at all
13 residual boundaries. Backward variables are the actual derivatives
of token-1023 pre-softmax logits 13 and 262 with respect to each
recorded residual, without a loss derivative or random projection.
They are saved in raw Jacobian units; within-vector standardization
removes any scalar gradient convention from the shape diagnostic.
Measurement uses float32 without autocasting.
The last-boundary Jacobian is checked against the analytic derivative
of RMSNorm followed by the tied linear readout; its maximum error at
the final time is $3.73\times10^{-9}$.

Each time has 208 nondegenerate forward slices and 200 backward
slices per logit; the final-boundary derivative at the earlier token
is structurally zero. For each vector separately, we report its
uncentered RMS, centered standardized third moment, and fourth moment
minus three. Fixed token-1023 pairs are $(h^1,h^6)$,
$(h^6,h^{12})$, and $(h^6,\partial f_{13}/\partial h^6)$,
using the same residual coordinate in both components for all eight
texts. Gaussian 95\% ellipses use the measured covariance and are
descriptive references, not confidence regions.
At the terminal time, the fractions outside these ellipses span
3.906--6.641\%; the raw final-residual mean RMS is 22.152,
compared with 0.724 at initialization.
The maximum backward excess kurtosis of 0.9709 occurs at update 5000,
sample 2, boundary 6, token 1023, logit 13, and is retained.
Figure~\ref{fig:ns5-diagnostics} in Appendix~\ref{app:gaussian-numerics}
uses the first predefined text,
without selection by Gaussian appearance.
All predefined times and slices are retained in the measurement archive.
The numerical supplement provides the complete time-course summaries
and the illustrative raw vectors.

\paragraph{Internal MLP distributions.}
\label{app:gpt2-internal}
Additional measurements in the same seed-1337 model use the initial
and update-4500 checkpoints, with the same eight texts and two token
positions. Forward hooks record the pre-GELU up-projection, GELU
output, MLP down-projection increment, and residual after each complete
block, giving 192 separate slices per variable and time.
Hidden vectors have 3072 coordinates; increments and residuals have 768.
Measurements use CPU float32 without autocasting. Each vector is centered
and standardized separately, without pooling coordinates across texts,
positions, or blocks.

The first block develops strongly non-Gaussian preactivations:
all sixteen slices at update 4500 have skewness 3.38--7.54 and
excess kurtosis 29.16--78.75 (Table~\ref{tab:ns5-internal}).
This departure therefore precedes GELU.
It varies substantially across depth: the median pre-GELU excess
kurtosis over all 192 slices is 0.5121, with smaller deviations in
middle blocks (Fig.~\ref{fig:ns5-internal}).
Approximately Gaussian residual shapes thus coexist with strongly
non-Gaussian hidden variables in the same model.
Internal measurements stop at update 4500, whereas the residual and
Jacobian measurements above continue through update 6200.

\begin{table}[!ht]
\centering
\color{black}
\begin{tabular}{lrrrr}
\toprule
& \multicolumn{2}{c}{Initialization} & \multicolumn{2}{c}{Update 4500} \\
First-block variable & Skewness & Excess kurt. & Skewness & Excess kurt. \\
\midrule
Pre-GELU & $-0.0214$ & $0.0256$ & $5.5374$ & $54.8062$ \\
Post-GELU & $1.6722$ & $3.0864$ & $10.9190$ & $138.0886$ \\
MLP increment & $0.0006$ & $0.0035$ & $0.0572$ & $-0.0393$ \\
Post-block residual & $0.0081$ & $0.0132$ & $0.0374$ & $-0.1160$ \\
\bottomrule
\end{tabular}
\caption{Internal-variable contrast in the all-NS5 model. Entries are
medians over sixteen separately measured first-block slices, not
uncertainty estimates; coordinates and slices are not assumed independent.}
\label{tab:ns5-internal}
\end{table}

\begin{figure}[t]
\centering
\color{black}
\includegraphics{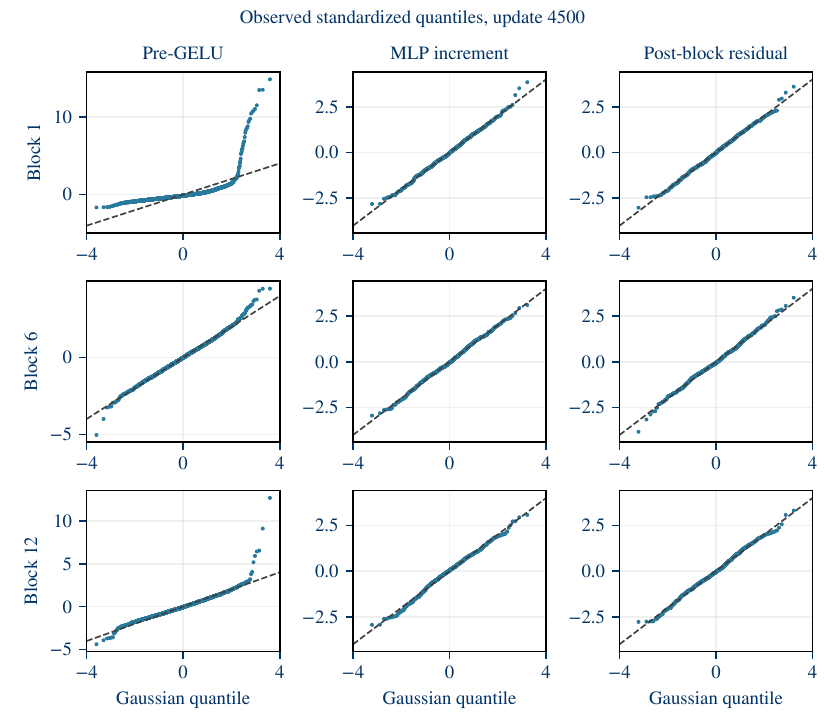}
\caption{Internal distributions at update 4500 for the first fixed text,
token 1023, and blocks 1, 6, and 12. Each panel standardizes one vector
separately and retains its full tails; vertical ranges differ across
panels. Preactivations have 3072 coordinates, increments and residuals 768.
Dashed lines give the standard Gaussian reference.}
\label{fig:ns5-internal}
\end{figure}

\subsection{Exploratory nonuniform depth interventions}
\label{app:gpt2-adaptive}

These historical AdamW runs keep the twelve- or twenty-four-stage
parameter budget and multiply both residual branches, including their
down-projection biases, by $c_\ell=L\Delta\tau_\ell$.
The original learning-rate and decay schedule is retained;
the corresponding block Adam stabilizer is $c_\ell10^{-8}$,
while endpoint stabilizers remain $10^{-8}$.
Half Copy (HC) changes initialization only: up-projections are copied
within a group and down-projections combine a shared and independent
Gaussian with correlation $\rho$, preserving marginal variances.
Attention groups consist of complete heads, with matching Q/K/V rows
and output columns. The parameters remain untied during training.

\begin{table}[!ht]
\centering
\color{black}
\setlength{\tabcolsep}{4pt}
\begin{tabular}{llrr}
\toprule
Stages & Intervention & Validation CE & HellaSwag\\
\midrule
12 & Fitted tail, HC & 3.0878 & 0.2996\\
24 & Two ends, HC & 3.0611 & 0.3089\\
12 & Mild tail, HC & 3.0741 & 0.3075\\
24 & Soft head, HC & 3.1034 & 0.3037\\
24 & Strong head, HC & 3.0981 & 0.3011\\
\bottomrule
\end{tabular}
\caption{All five completed adaptive candidates, seed 1337, evaluated
at the same endpoint as the uniform baselines in
Table~\ref{tab:gpt2-shape}. Favorable and unfavorable outcomes are shown.}
\label{tab:gpt2-adaptive}
\end{table}

For the twenty-four-stage two-end grid, groups of 8, 12, and 4
stages use $c=(0.75,1.25,0.75)$ and pair HC at both ends with
$\rho=1/3$. Soft-head and strong-head variants use
$c=(0.825,1.2,0.75)$ and $(0.675,1.30,0.75)$,
respectively; their head correlations are $7/33$ and $13/27$,
with tail correlation $1/3$.
The twelve-stage mild-tail grid has groups of 3, 7, and 2 stages,
$c=(1,15/14,0.75)$, and pair HC only in its last two stages.
The fitted twelve-stage grid ranges from $c_1=1.4568$ to
$c_{12}=0.3505$, with $\rho=1/3$ and integer HC groups in the
last five stages. Full-precision mesh and group-count data accompany
the supplementary protocol.
The no-HC twenty-four-stage control stopped at 1500 updates under its
preset early-loss rule; the twelve-stage two-end run was paused at 5466.
These incomplete runs supply no endpoint.
The completed single-seed comparisons change full recipes and show
no consistent adaptive-depth advantage.
\par
\endgroup

\endgroup
\end{document}